\documentclass[10pt,journal,compsoc]{IEEEtran}
\ifCLASSOPTIONcompsoc
    \usepackage[nocompress]{cite}
\else
    \usepackage{cite}
\fi
\ifCLASSINFOpdf
\else
\fi
\usepackage{subfiles}
\usepackage{graphicx}
\usepackage{amsmath}
\usepackage{amssymb}
\usepackage{multirow}
\usepackage{makecell}
\usepackage{color}
\usepackage{bm}
\usepackage[pagebackref,breaklinks,colorlinks]{hyperref}
\usepackage{marvosym}
\usepackage{xcolor}
\usepackage{booktabs}

\usepackage[capitalize]{cleveref}
\crefname{section}{Sec.}{Secs.}
\Crefname{section}{Section}{Sections}
\Crefname{table}{Tab.}{Tabs.}
\crefname{table}{Tab.}{Tabs.}

\newcommand{\NA}{N/A}

\newcommand{\xmark}{$\times$}
\newcommand{\xm}{\xmark}
\newcommand{\cm}{\checkmark}

\newcommand{\ETAL}{{\emph{et al.}}}
\newcommand{\EG}{{\emph{e.g.}}}
\newcommand{\IE}{{\emph{i.e.}}}
\newcommand{\TODO}[1]{\colorbox{yellow}{TODO}}
\newcommand{\C}[1]{{\textcolor[HTML]{00A7EE}{#1}}}

\newcommand{\FIG}[1]{\cref{#1}}
\newcommand{\TABLE}[1]{\cref{#1}}
\newcommand{\EQUA}[1]{Eq.~\ref{#1}}
\newcommand{\SEC}[1]{\cref{#1}}

\newcommand{\MODEL}{V$^{2}$-Net}
\newcommand{\MODELSHORT}{V}
\newcommand{\MODELCITE}{\cite{wong2022view}}
\newcommand{\EMODEL}{E-V$^{2}$-Net}
\newcommand{\EMODELSHORT}{EV}
\newcommand{\LINEARNOTE}{with linear S2}

\newcommand{\SOCIALLSTM}{Social-LSTM}
\newcommand{\SOCIALLSTMCITE}{\cite{alahi2016social}}
\newcommand{\SOCIALGAN}{S-GAN}
\newcommand{\SOCIALGANCITE}{\cite{gupta2018social}}
\newcommand{\SOPHIE}{SoPhie}
\newcommand{\SOPHIECITE}{\cite{sadeghian2019sophie}}
\newcommand{\BIGAT}{S-BiGAT}
\newcommand{\BIGATCITE}{\cite{kosaraju2019social}}
\newcommand{\SIMAUG}{SimAug}
\newcommand{\SIMAUGCITE}{\cite{liang2020simaug}}
\newcommand{\PECNET}{PECNet}
\newcommand{\PECNETCITE}{\cite{mangalam2020not}}
\newcommand{\SRLSTM}{SR-LSTM}
\newcommand{\SRLSTMCITE}{\cite{zhang2019sr}}
\newcommand{\SRPAMI}{E-SR-LSTM}
\newcommand{\SRPAMICITE}{\cite{zhang2020social}}
\newcommand{\GARDEN}{Multiverse}
\newcommand{\GARDENCITE}{\cite{liang2020garden}}
\newcommand{\PEEKING}{Next}
\newcommand{\PEEKINGCITE}{\cite{liang2019peeking}}
\newcommand{\STAR}{STAR}
\newcommand{\STARCITE}{\cite{yu2020spatio}}
\newcommand{\TP}{TF}
\newcommand{\TPCITE}{\cite{giuliari2020transformer}}
\newcommand{\TPNMS}{TPNMS}
\newcommand{\TPNMSCITE}{\cite{liang2020temporal}}
\newcommand{\TRAJECTRONPP}{Trajectron++}
\newcommand{\TRAJECTRONPPCITE}{\cite{salzmann2020trajectron}}
\newcommand{\INTROVERT}{Introvert}
\newcommand{\INTROVERTCITE}{\cite{shafiee2021introvert}}
\newcommand{\LBEBM}{LB-EBM}
\newcommand{\LBEBMCITE}{\cite{pang2021trajectory}}
\newcommand{\YNET}{$\textsf{Y}$-net}
\newcommand{\YNETCITE}{\cite{mangalam2020s}}
\newcommand{\SPECTGNN}{SpecTGNN}
\newcommand{\SPECTGNNCITE}{\cite{cao2021spectral}}

\newcommand{\AGENTFORMER}{Agentformer}
\newcommand{\AGENTFORMERCITE}{\cite{yuan2021agentformer}}
\newcommand{\MANTRA}{MANTRA}
\newcommand{\MANTRACITE}{\cite{marchetti2020mantra}}
\newcommand{\DAGNET}{DAG-Net}
\newcommand{\DAGNETCITE}{\cite{monti2021dag}}
\newcommand{\STGCNN}{STGCNN}
\newcommand{\STGCNNCITE}{\cite{mohamed2020social}}
\newcommand{\STCNET}{STC-Net}
\newcommand{\STCNETCITE}{\cite{li2021spatial}}
\newcommand{\MSN}{MSN}
\newcommand{\MSNCITE}{\cite{wong2021msn}}
\newcommand{\MID}{MID}
\newcommand{\MIDCITE}{\cite{gu2022stochastic}}
\newcommand{\SHENET}{SHENet}
\newcommand{\SHENETCITE}{\cite{meng2022forecasting}}
\newcommand{\SEEM}{SEEM}
\newcommand{\SEEMCITE}{\cite{wang2022seem}}
\newcommand{\CSCNET}{CSCNet}
\newcommand{\CSCNETCITE}{\cite{xia2022cscnet}}
\newcommand{\SSL}{Social-SSL}
\newcommand{\SSLCITE}{\cite{tsao2022social}}
\newcommand{\EQMOTION}{EqMotion}
\newcommand{\EQMOTIONCITE}{\cite{xu2023eqmotion}}

\begin{document}
%
\title{
    Another Vertical View: A Hierarchical Network for Heterogeneous Trajectory Prediction via Spectrums
}
%
%
%
%

\author{Beihao Xia*,
        Conghao Wong*,
        Duanquan Xu,
        Qinmu Peng,
        and~Xinge You~(\Letter),~\IEEEmembership{Senior Member,~IEEE}
\thanks{
    * Equal contribution.
    Codes at \url{https://github.com/cocoon2wong/E-Vertical}.
}
\thanks{
    The authors are with Huazhong University of Science and Technology, Wuhan, Hubei, P.R.China.
    Email: xbh\_hust@hust.edu.cn, conghaowong@icloud.com, \{pengqinmu, xudq, youxg\}@hust.edu.cn\\
}
}

%
%

\markboth{Journal of \LaTeX\ Class Files,~Vol.~14, No.~8, August~2015}%
{Shell \MakeLowercase{\textit{et al.}}: Bare Demo of IEEEtran.cls for Computer Society Journals}
%



\IEEEtitleabstractindextext{%

\begin{abstract}

    With the fast development of AI-related techniques, the applications of trajectory prediction are no longer limited to easier scenes and trajectories.
    More and more trajectories with different forms, such as coordinates, bounding boxes, and even high-dimensional human skeletons, need to be analyzed and forecasted.
    Among these heterogeneous trajectories, interactions between different elements within a frame of trajectory, which we call ``Dimension-wise Interactions'', would be more complex and challenging.
    However, most previous approaches focus mainly on a specific form of trajectories, and potential dimension-wise interactions are less concerned.
    In this work, we expand the trajectory prediction task by introducing the trajectory dimensionality $M$, thus extending its application scenarios to heterogeneous trajectories.
    We first introduce the Haar transform as an alternative to Fourier transform to better capture the time-frequency properties of each trajectory-dimension.
    Then, we adopt the bilinear structure to model and fuse two factors simultaneously, including the time-frequency response and the dimension-wise interaction, to forecast heterogeneous trajectories via trajectory spectrums hierarchically in a generic way.
    Experiments show that the proposed model outperforms most state-of-the-art methods on ETH-UCY, SDD, nuScenes, and Human3.6M with heterogeneous trajectories, including 2D coordinates, 2D/3D bounding boxes, and 3D human skeletons.

\end{abstract}

}

\maketitle

\IEEEdisplaynontitleabstractindextext

%
\IEEEpeerreviewmaketitle

\section{Introduction}

\IEEEPARstart{T}{}{RAJECTORY} prediction aims at inferring agents' possible future trajectories considering potential influencing factors.
It is an essential but challenging task, which can be widely applied to behavior analysis \cite{chai2019multipath}, robot navigation and planning \cite{chen2022scept}, autonomous driving \cite{lee2017desire}, tracking \cite{pellegrini2009youll}, detection \cite{fernando2018soft}, and many other tasks \cite{morris2011trajectory,xie2017learning}.
According to previous works \cite{zheng2021unlimited,ma2019trafficpredict}, trajectory prediction mainly involves homogeneous and heterogeneous trajectory prediction.
Different from homogeneous trajectory prediction, heterogeneous trajectory prediction aims to handle agents with different types and preferences simultaneously.
With the gradual enrichment of agents involved in this task (\EG, pedestrians, bikers, and carts), more researchers start exploring how to represent the interactions among heterogeneous agents \cite{ma2019trafficpredict,chandra2019traphic,zheng2021unlimited}, which is a hot and exciting problem.

\begin{figure}
    \centering
    \includegraphics[width=0.98\linewidth]{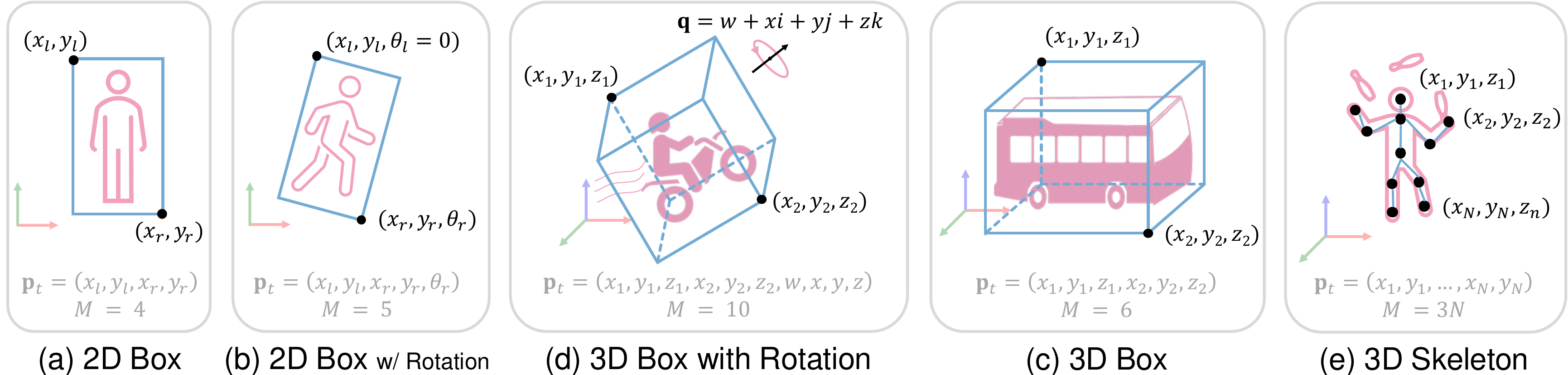}
    \caption{
        Examples of several trajectory forms.
        Different forms of trajectories may exist in complex scenarios regardless of agent categories.
        Trajectories are no longer limited to 2D coordinate series.
    }
    \label{fig_intro_ht}
\end{figure}

\textbf{Forms of Trajectories.}
Unfortunately, although these methods have achieved some success in complex scenarios, most of them mainly focus on heterogeneous agents but pay less attention to the heterogeneity of trajectories caused by differences in trajectory representations.
In detail, most researchers in predicting trajectories from videos are working on forecasting two-dimensional (2D) trajectories rather than trajectories with different structures.
Recently, more datasets with full 3D or high-dimensional annotations have become accessible \cite{caesar2019nuscenes}.
Accordingly, various trajectory forms could arise in different scenarios and applications.
As shown in \FIG{fig_intro_ht}, the trajectory could be formed with either coordinates or bounding boxes, even skeletons.
However, most methods could only handle one of these trajectory forms, especially mostly 2D trajectories.
Thus, the main focus of this manuscript is to find a ``uniform'' trajectory prediction structure, thus making it available to forecast trajectories with different forms according to different application requirements.
Referring to previous ``heterogeneous agents'', this manuscript defines trajectories with different representation forms as \textbf{``Heterogeneous Trajectories''}.

\begin{figure*}
    \centering
    \includegraphics[width=1.0\linewidth]{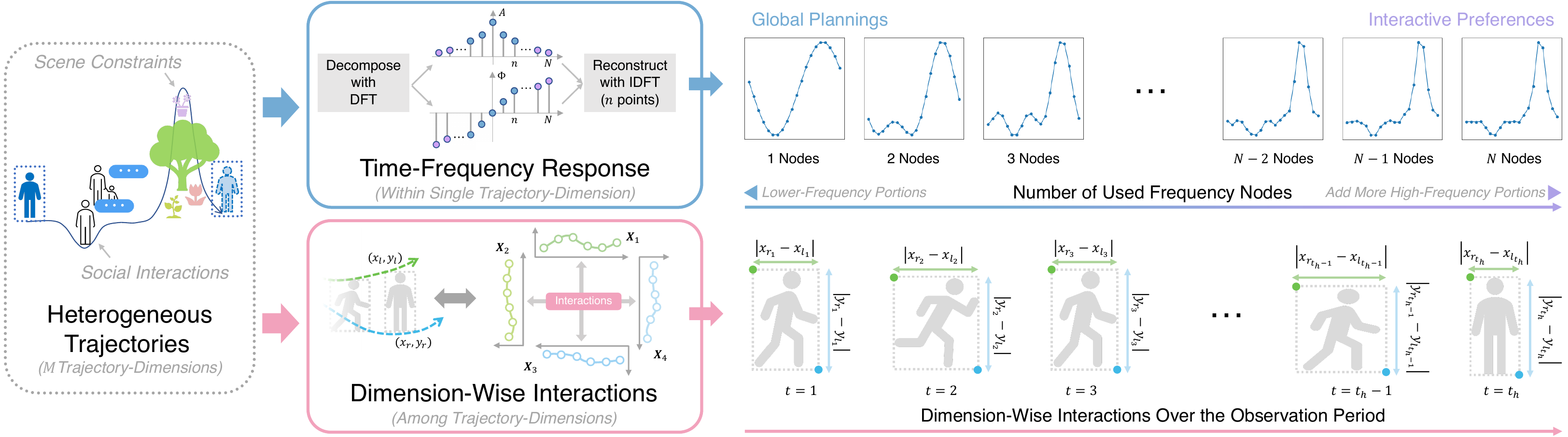}
    \caption{
        Illustrations of challenges in heterogeneous trajectory prediction, \IE, ``Time-Frequency Response'' and ``Dimension-wise Interaction''.
    }
    \label{fig_intro_hp}
\end{figure*}

\textbf{Heterogeneous Trajectories and Challenges.}
In this manuscript, we denote the length of a frame of recorded trajectory vector at some moment as $M$ and call the corresponding trajectory the \emph{$M$-dimensional trajectory}.
The heterogeneous trajectories could be treated as a set of $M$-dimensional trajectories, where $M \in \{M_1, M_2, ...\}$.
We partially list heterogeneous trajectories with different dimensionalities $M$ in 2D and 3D scenes in \FIG{fig_intro_ht}.
For example, various bounding boxes can be treated as a trajectory frame.
Even the high-dimensional human skeleton could also be considered as a frame of trajectory.
With the development of various sensing technologies, trajectories with more dimensions ($M$) may also be required for prediction.
Thus, compared to homogeneous trajectories, the most significant difference in heterogeneous trajectory prediction is that it considers different forms of trajectories when making predictions, especially those trajectory forms with higher dimensionalities.
Accordingly, as shown in \FIG{fig_intro_hp}, two main challenges have arisen corresponding to the higher dimensionality, including the description of a single trajectory-dimension and the interactions among dimensionalities.

\textbf{Challenge A. Modeling of a Single Trajectory-Dimension.}
The $M$-dimensional trajectory has $M$ trajectory-dimensions.
Each dimension is a time sequence.
Thus, modeling and analyzing each dimension significantly matters when forecasting heterogeneous trajectories.
However, most current approaches may lack the overall analyses of all factors at different temporal scales.
Most agents, such as pedestrians, always simultaneously plan their future activities at different ``levels''.
For example, they may first plan their coarse motion trends and then make fine decisions about interactive variations.
Although some methods employ networks with attention mechanisms \cite{girdhar2017attentional} (like Graph Attentive Networks \cite{huang2019stgat} and Transformers \cite{yu2020spatio,yuan2021agentformer}) as backbones to model agents' status, they may have difficulty directly representing agents' detailed motions at different temporal scales.
In addition, capturing sudden changes in trajectories is a problem that has not been well addressed.
Especially for heterogeneous trajectories, in which a single dimension could also change dramatically over time, not to mention all $M$ dimensions, modeling and encoding these dimensions has become another challenge.

The classical theories of time-frequency analysis provide us with ideas to address this challenge.
The Fourier transform (FT), and its variations have significantly succeeded in signal processing and computer vision communities \cite{zhu2007real,kaur2021fractional}.
They decompose signals into a series of sinusoids on different frequency portions to reflect the frequency response of different frequency scales that may be difficult to obtain in original signals directly.
However, the Fourier transform only knows which frequency components the signal contains and cannot know the frequency information of the signal at different times from the frequency domain, which is unsuitable for analyzing a signal whose frequency changes over time.
Especially in heterogeneous trajectory prediction, agents' diverse motion patterns, differentiated behavior preferences, and uncertain future decisions shift trajectories irregularly over time, making Fourier transform challenging to handle.
Wavelet Transforms (WTs) have better capabilities in terms of multi-resolution and multi-scale characteristics for adaptive time-frequency signal analyses and achieve great success in numerous tasks \cite{zhang2002edge,tang2003skeletonization,cheng2006real}, prompting it to be selected as an alternative to the Fourier transform to deal with rapidly changing trajectories.
Similar to the Fourier transform, signals can be represented by a series of wavelets in the wavelet transform, where wavelets can be localized in both time and frequency.
Therefore, we can choose different transformations to cope with diverse heterogeneous trajectory prediction scenarios.
They provide a ``vertical'' view for processing and analyzing sequences, thus presenting elusive features in original signals.

Interestingly, Becker \ETAL\cite{becker2018red} find that the latest two observed steps contribute to the predicted trajectory for a surprising 88.3\%.
Similarly, Monti \ETAL\cite{monti2022many} also indicate that subsequent states contribute more when forecasting.
The above study shows that trajectory prediction can also be achieved with a few trajectory points containing most of the trajectory's energy.
Recently, some works have divided trajectory prediction into a two-stage pipeline by considering the contribution of each future step, which we call the ``hierarchical prediction'' strategy.
For example, \cite{mangalam2020not,mangalam2020s,tran2021goal,wong2021msn} have been proposed to predict agents' keypoints, waypoints, or destinations first and then interpolate the complete trajectories under the chosen conditions.
Inspired by these approaches, a natural thought is to hierarchically predict agents' trajectories at different \emph{frequency scales}, including \emph{Global Plannings} and \emph{Interactive Preferences}.
Correspondingly, the low-frequency portions in trajectories could reflect agents' motion trends, and the high-frequency portions as agents' detailed interactions, shown in \FIG{fig_intro_hp}.
Forecasting trajectories hierarchically via spectrums could better describe multi-scale characteristics due to the flexibility of splitting based on different frequency components without specifying waypoints and destinations.
At the same time, when the trajectory is transferred from the time domain to the frequency domain, the energy contained in the trajectory will be redistributed.
Therefore, a suitable keypoints selection strategy is crucial.
Too few keypoints will cause excessive energy loss, resulting in large deviations in the reconstructed trajectory; on the contrary, too many keypoints cannot reflect the superiority of hierarchical prediction.

One natural thought is that transforms could be used before the model implementing to learn agents' behaviors at different scales via spectrums hierarchically.
Considering the properties of transforms, to model a single trajectory-dimension, we employ Discrete Fourier Transform and Discrete Haar Transform to get the corresponding time-frequency joint representation, \IE, trajectory spectrums, to capture agents' detailed motion preferences at different frequency scales, thus better modeling trajectory-dimensions.

\textbf{Challenge B. Interactions among Dimensionalities.}
Another important change is that heterogeneous trajectories may have higher dimensionality compared to homogeneous trajectories.
Thus, ``relations'' between trajectory-dimensions have become crucial when forecasting.
It means that when predicting the trajectory of the next moment in one dimension, other dimensions and their relationships need to be considered simultaneously.
These relationships are quite similar to social interactions, whereas the interaction participators have become different dimensions of the heterogeneous trajectory rather than several neighbor agents.
Since these dimensional interactions happen ``vertically'' to both the time and space axes, in this manuscript, we will investigate the possible effects of these interactions within heterogeneous trajectories by introducing \emph{``another vertical view''} in trajectory prediction.
Similar to social interaction, we call them the \textbf{``Dimension-wise Interaction''}.

In statistics, an interaction may arise when considering the relationship among three or more variables and describes a situation in which the effect of one causal variable on an outcome depends on the state of a second causal variable.
The social interaction considered in trajectory prediction is a special case of the above interaction.
Similarly, we define dimension-wise interaction as the interactions among two or more trajectory-dimensions within the same agent's trajectory when forecasting their future trajectories.
The dimension-wise interaction also happens among at least three variables, but these variables all belong to the same agent's trajectory (at different time steps).
Denote the observed $M$-dimensional position (can be coordinates, 2D or 3D bounding boxes, or even 3D skeletons) of an agent at some moment $t$ as $\mathbf{p}_t = \left(p^1_t, ..., p^m_t, ..., p^M_t\right)^\top \in \mathbb{R}^M$, the dimension-wise interaction can be represented as the following term $I$ when forecasting the next time step:
\begin{equation}
    \label{eq_interaction}
    \hat{p}^m_{t+1} = \mathrm{Net}\left(
        p^m_t,
        I \left(p^1_t, p^2_t, ..., p^m_t, ..., p^M_t\right)
    \right).
\end{equation}

As the dimensionality of the heterogeneous trajectories rises, these inter-dimensional interactions will become non-negligibly complex (at least $\mathcal{O}(M^2)$) when forecasting.
For example, there will be at least $N(M) = M(M-1)/2$ paris of interactions even if only considering them as undirected.
In addition, trajectory-dimensions on adjacent moments are often highly correlated, like $\hat{p}^m_{t+1}$ and $p^m_t$ in \EQUA{eq_interaction}.
Nevertheless, the time-frequency properties of each trajectory-dimension still need to be considered when representing these interactions.
Therefore, there are also challenges in finding an effective modeling way to fit this complex interaction term $I$.
Thus, how to model the above dimension-wise interaction term $I \left(p^1_t, p^2_t, ..., p^m_t, ..., p^M_t\right)$ in heterogeneous trajectories has become our other concern.

In fact, although these interactions may be established due to the use of two or more fully connected layers when embedding trajectories, which is the first operation in most current approaches, these established ones are not representative enough (
    discussed in \SEC{sec_qual}
).
It could be difficult to model these dimension-wise interactions directly because of either the higher computation complexity or the frequency response characteristics of each trajectory-dimension.
Modeling these interactions from a different perspective relative to the original time sequences has become a better option.
As discussed above, transforms and trajectory spectrums can help us locate the ``buried'' information.
Like a single trajectory-dimension, we can also model dimension-wise interactions via trajectory spectrums to better capture the time-frequency characteristics within these interactions.
Thus, our focus has turned to how to simulate dimension-wise interaction $I$ via spectrums while simultaneously considering how it varies over time or frequency.

Talking about interactions, a natural thought is to use graphs to model the relations between interaction participators.
Among dimension-wise interactions, each participator is a single trajectory-dimension, which means that we need to calculate the similarity of these participators as edges of the graph structure. 
However, it is challenging to model such a complex relation between two trajectory-dimensions or the corresponding spectrums within a static graph since both of them may change over time or frequency and own specific time-frequency joint representation, not to mention how to gather information on all adjacent edges into a node.
Thus, we need to find a powerful approach to simultaneously model ``time-frequency response'' and ``dimension-wise interactions'' two factors with trajectory spectrums.

Fortunately, bilinear structure \cite{tenenbaum2000separating,lin2017bilinear} have been proposed to model ``two-factor'' characteristics jointly like ``style'' and ``content'' in images, and achieved great success in many computer vision tasks \cite{lin2017bilinear,xu2021multimodal,girdhar2017attentional,guo2021bilinear}.
These two factors may be connected in some undirect forms, which could be difficult to represent directly.
The bilinear structure combines these two factors by utilizing outer-product and pooling operations.
During this process, two factors will be simultaneously considered, and their relations will be constructed through the outer product matrix when classifying images, in which element-wise similarity computation will be conducted to measure how each vector component contributes adaptively.
The bilinear form also simplifies gradient computations to ensure both factors are equally optimized while speeding up training \cite{lin2017bilinear}.

Similar to these two image factors ``style'' and ``content'', our considered frequency response and the dimension-wise interaction also represent heterogeneous trajectories from two different perspectives.
In detail, the frequency characters describe how each trajectory-dimension forwards and changes over time, while the dimension-wise interactions describe how other trajectory-dimensions affect or limit the change of a specific trajectory-dimension.
The outer-product operation in the bilinear structure also provides a new implementation to view and simulate interactions among these trajectory dimensions.
In this process, the dimension-wise interactions will be established as the projections for time-frequency representations of different dimensions onto the others, thus representing interactions while efficiently integrating time-frequency properties.

Thus, in this manuscript, we will take advantage of both transforms and bilinear structures to encode and forecast heterogeneous trajectories \emph{from another view}, \IE, modeling each trajectory-dimension as well as the dimension-wise interaction with trajectory spectrums, thus modeling the interactions among trajectory-dimensions while simultaneously considering their time-frequency response characters, achieving the refined extraction of agents' behavior representations while maintaining inter-dimensional interactions within heterogeneous trajectories.

\textbf{Contributions.}
This manuscript is an extension of our previous conference paper \MODELCITE.
The former proposed \MODEL~predicts trajectories in a ``vertical'' view, \IE, modeling and forecasting trajectories via the Fourier spectrums instead of the original time series hierarchically.
In this work, the proposed \EMODEL~extends the existing \MODEL~\emph{from another vertical view} by focusing on dimension-wise interactions in the heterogeneous trajectories.
Specifically, this manuscript enhances \MODEL~from the following ways.
On the one hand, we define and verify the existence of dimension-wise interactions in heterogeneous trajectories, which extends the structure and network settings of the original network to be applied to forecast heterogeneous trajectories.
On the other hand, \EMODEL~introduces Wavelet transform \cite{haar1910theorie} with time-frequency localization properties as an alternative to Fourier transform to obtain trajectory spectrums to cope with diverse scenarios (such as rapidly changing trajectories).
Furthermore, we adopt an extra bilinear structure \cite{tenenbaum2000separating} to model and fuse the above ``dimension-wise interactions'' and ``frequency response'' two factors, thus extracting refined behavior representations while maintaining interactions within heterogeneous trajectories.
Experiments show that the proposed \EMODEL~outperforms most state-of-the-art methods on the ETH-UCY benchmark, Stanford Drone Dataset, nuScenes, Human3.6M with heterogeneous trajectories, including 2D coordinates, 2D and 3D bounding boxes, and 3D skeletons.
In summary, the contributions of this manuscript are listed as follows:

$\bullet$ We define the ``Heterogeneous Trajectories'' to extend the current trajectory prediction task and propose a generic prediction method \EMODEL~to handle trajectories with different representation forms.

$\bullet$ This work further investigates trajectories with different spectrums to model the frequency responses of trajectories at different frequency scales simultaneously to obtain richer multi-scale characteristics.

$\bullet$ Potential interactions between different elements within heterogeneous trajectories, which we call the ``Dimension-wise Interactions'', have been focused specifically, and a bilinear structure is adopted to model and fuse ``frequency response'' and ``dimension-wise interaction'' factors when forecasting heterogeneous trajectories.

\section{Related Work}

Trajectory prediction has received increasing attention recently.
Alahi \ETAL~\cite{alahi2016social} treat this task as sequence generation and employ LSTMs to model and predict pedestrians' positions in the next time step recurrently.
Researchers like \cite{giuliari2020transformer,yu2020spatio,yuan2021agentformer,monti2022many} have also designed different Transformers to obtain better trajectory representations.
In addition, several factors, such as social/scene interactions \cite{mohamed2020social,zhang2019sr,tsao2022social}, agents' motion preferences \cite{chai2019multipath,wong2021msn}, the goal distributions \cite{choi2019drogon,girase2021loki,rhinehart2019precog} and stochastic trajectory prediction \cite{gupta2018social,sadeghian2019sophie,kosaraju2019social,salzmann2020trajectron}, have been widely investigated.
Notably, trajectory prediction discussed in this manuscript focuses more on the scenarios such as walking pedestrians and city streets, and less on fast-changing scenarios like highway vehicles, which means the safety concerns in autonomous-driving-related tasks may be less considered.
In contrast, we pay more attention to the diversity of agents' activities.

~\\\textbf{Heterogeneous Trajectory Prediction.}
According to the classification methods of previous works like \cite{zheng2021unlimited,ma2019trafficpredict}, trajectory prediction mainly involves homogeneous and heterogeneous trajectory prediction in real scenarios.
Homogeneous trajectory prediction predicts future trajectories under the same category (\EG, only pedestrians).
On the contrary, heterogeneous predicts future trajectories of the \emph{Heterogeneous Agents} under different categories (\EG, pedestrians, cars, and bikes).
The heterogeneous trajectory prediction considered in this manuscript further expands this definition to include trajectories with different forms, which we call the \emph{Heterogeneous Trajectories}.
Yagi \ETAL \cite{yagi2018future} present an approach to forecast agents' 2D bounding box positions in the first-person videos by taking into account agents' post-locations, poses, and ego-motions.
Quan \ETAL \cite{quan2021holistic} propose the Holistic LSTM to predict 2D bounding boxes along with the help of depth estimation and the optical flow.
Saadatnejad \ETAL \cite{saadatnejad2022pedestrian} design an LSTM-based method to predict 3D bounding boxes by encoding positions and velocities together.
Xu \ETAL \cite{xu2023eqmotion} also propose a graph network to forecast human skeletons.
Although these methods have achieved great performance enhancement for trajectories with different forms, they often require additional model inputs, like skeletons and optical flow information, making them difficult to adapt their models to different trajectory forms and scenarios.
Few methods can handle heterogeneous trajectories with different forms, let alone interactions and relations within heterogeneous trajectories at the same time.
Combining these strengths of existing methods, a natural thought is to design a ``generic'' prediction network to cope different forms of heterogeneous trajectories.

~\\\textbf{Applications of Transforms.}
As powerful tools, the Fourier transform (FT) and its variations have been extensively studied and achieved impressive success in various image processing and computer vision tasks.
Cheng \ETAL~\cite{cheng2021amenet} present a Fast Fourier Transform-based algorithm, which brings high computational efficiency and reliability for the multichannel interpolation in image super-resolution.
Kaur \ETAL~\cite{kaur2021fractional} propose a Fractional Fourier Transform based Riesz fractional derivative approach for edge detection and apply it to enhance images.
Komatsu \ETAL~\cite{komatsu2017mean} construct the 3-D mean-separation-type short-time DFT and apply it to denoise moving images.
It is worth noting that FTs have also been widely applied in handling time-series forecasting problems.
Mao \ETAL~\cite{mao2020history,mao2019learning} employ DCT to help predict human skeleton graphs in motion forecasting.
Cao \ETAL~\cite{cao2020spectral,cao2021spectral} propose a spectral-temporal graph to model interactions (not trajectories) when forecasting trajectories.

As an improvement to FTs, Wavelet Transforms (WTs) have better capabilities in terms of multi-resolution and multi-scale characteristics to achieve the adaptive time-frequency signal analysis.
Similar to FTs, WTs have also succeeded in numerous tasks.
Zhang \ETAL~\cite{zhang2002edge} present a Dyadic Wavelet Transform-based edge detection method by scale multiplication.
Tang and You \cite{tang2003skeletonization} construct a novel Wavelet function to extract the skeleton of ribbon-like shapes to detect edges.
Cheng \ETAL~\cite{cheng2006real} provide a Discrete Wavelet Transform-based approach for detecting, tracking, and identifying multiple moving objects.
WTs are also well-applied for time-series analysis.
For example, Yang \ETAL~\cite{yang2018time} propose a time-series prediction model based on fuzzy cognitive maps and the Wavelet Transform.

Considering that a trajectory can also be treated as a time series, the successful use of these powerful transforms motivates us to model \emph{agents' trajectories} with trajectory spectrums, therefore trying to obtain better representations.
Unfortunately, there seem to be no methods that directly use either the Fourier transform or its variations to analyze and forecast agents' trajectories in the field of trajectory prediction.
This manuscript attempts to model and forecast agents' trajectories by adopting different transforms (DFT and Haar wavelet) in the spectrum view for the first time.

~\\\textbf{Hierarchical Prediction and Bilinear Structures.}
More and more researchers \cite{choi2019drogon,girase2021loki,rhinehart2019precog} have treated trajectory prediction as a ``two-stage'' problem and predict trajectories \emph{hierarchically}.
Mangalam \ETAL \cite{mangalam2020not} infer trajectory end-points to assist in long-range multi-modal trajectory prediction.
Tran \ETAL \cite{tran2021goal} attempt to obtain multi-modal goal proposals from the additional goal channel to generate multiple predictions.
Wong \ETAL \cite{wong2021msn} use a set of generators to generate multiple destination proposals, and then interpolate to forecast multiple trajectories.
Others like \cite{mangalam2020s} also introduce several ``waypoints'' to help better predict agents' potential future intentions rather than the only destination points.

These methods have made an initial decomposition of the trajectory and the trajectory prediction task, where the features to be focused on in each of the two prediction stages can be distinguished, thus improving the efficiency of the prediction network while reducing the difficulty of network training.
The transforms presented above could decompose trajectories according to different frequency components, similar to the hierarchical prediction methods introduced here.
They may further improve the efficiency of the prediction network if the hierarchical prediction is implemented in the frequency domain.
Specifically, considering the Fourier transform and its variations could decompose time series into different frequency portions, a natural thought is to predict agents' future trajectories hierarchically on different \emph{frequency scales} rather than the ``end-point'' or ``way-points''.

Along with the hierarchical predictions, the heterogeneous trajectories mentioned above need to be considered simultaneously.
In order to combine two different factors, \IE, the frequency response and the dimension-wise interaction within heterogeneous trajectories, a natural idea is to use a bilinear structure to fuse these two factors.
Bilinear structures \cite{tenenbaum2000separating} are used to model two-factor variations such as ``style'' and ``content'' in images, and they have been widely applied in various tasks.
Lin \ETAL \cite{lin2017bilinear} utilize the bilinear structure in fine-grained visual recognition to obtain discriminative features and have reached better classification accuracy.
Xu \ETAL \cite{xu2021multimodal} design a second-order bilinear pooling module to effectively aggregate targets' deep and shallow features for better tracking.
Guo \ETAL \cite{guo2021bilinear} propose bilinear graph networks to model the context of joint embeddings of words and objects in visual question answering to model relationships between words.

However, the combination of frequency responses in trajectories and the potential dimension-wise interactions through bilinear structures has rarely appeared in the field of trajectory prediction, which is the main focus of this work.

\section{Method}

\begin{figure}[tbp]
    \centering
    \includegraphics[width=1.0\linewidth]{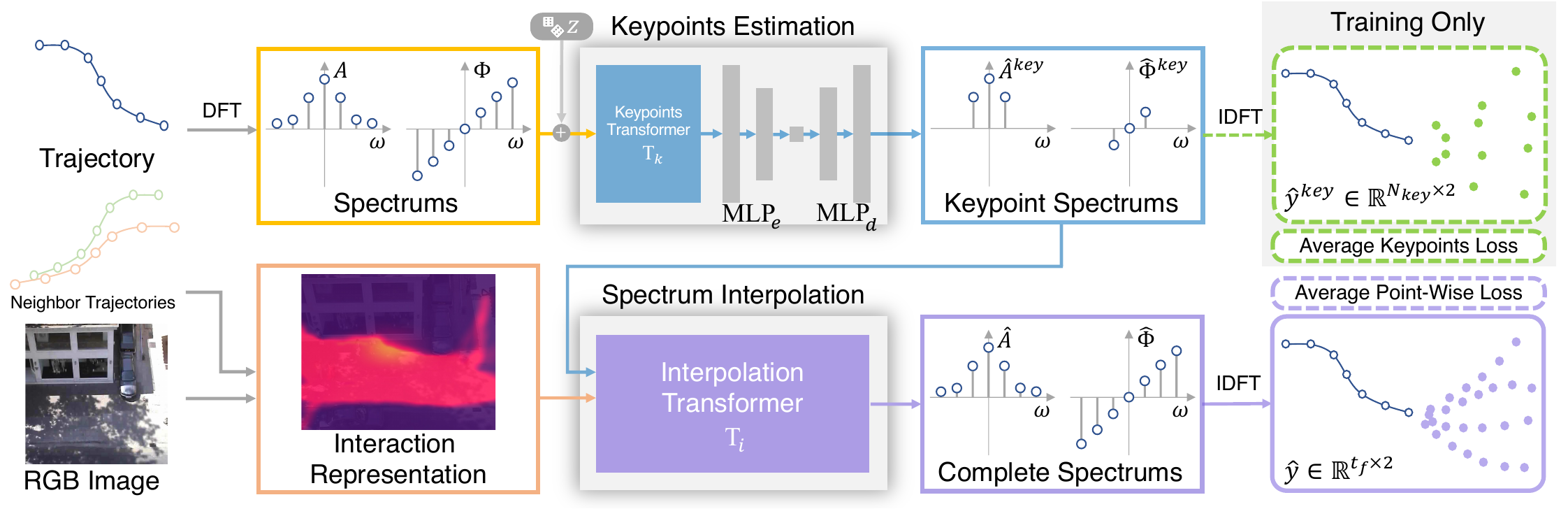}
    \caption{
        \MODEL~Overview.
        It has keypoints estimation and spectrum interpolation two sub-networks.
        It forecasts 2D coordinate trajectories ``from-coarse-to-fine'' hierarchically via Fourier spectrums.
    }
    \label{fig_overview}
\end{figure}


\subsection{\MODEL~($M$ Dimensions and Transform $\mathcal{T}$)}

As shown in \FIG{fig_overview}, \MODEL~has two main sub-networks, keypoints estimation sub-network and spectrum interpolation sub-network.
In our conference paper \MODELCITE, \MODEL~is used to forecast 2D coordinate trajectories via Fourier spectrums.
Here, we expand dimension of forecasted trajectories into $M$ and make it compatible with a generic transform $\mathcal{T}$.

\textbf{Problem Formulation.}
Let $\mathbf{p}^i_t = ({r_1}^i_t, {r_2}^i_t, ..., {r_M}^i_t)^\top \in \mathbb{R}^M$ denote the position of $i$-th agent at step $t$.
We use the time sequences of the above $M$-dimensional vectors to represent one form of heterogeneous trajectories.
Formally, we denote agents' $N$-point $M$-dimensional trajectory as
\begin{equation}
\begin{aligned}
    \label{eq_MD_traj}
    \mathbf{X}^i (1 \leq t \leq N, M)
    &= \left(\mathbf{p}_1^i, \mathbf{p}_2^i, ..., \mathbf{p}_N^i\right)^{\top} \\
    &= \left(
        \mathbf{X}_1^i, \mathbf{X}_2^i, ..., \mathbf{X}_M^i
    \right) \in \mathbb{R}^{N \times M}.
\end{aligned}
\end{equation}
Thus, the observed trajectory of agent $i$ during $t_h$ observation moments is denoted as $\mathbf{X}^i = \mathbf{X}^i(1\leq t \leq t_h, M) \in \mathbb{R}^{t_h \times M}$.
Given a video clip $\{\mathbf{I}\}$ that contains $N_a$ agents' observed trajectories $\{\mathbf{X}^i\}_{i=1}^{N_a}$, trajectory prediction focused in this manuscript aims to forecast their possible future positions $\{\hat{\mathbf{Y}}^i\}_{i=1}^{N_a}$ during the future period $(t_h+1 \leq t \leq t_h + t_f)$.
Here, $\hat{\mathbf{Y}}^i = \{\hat{\mathbf{y}}_k\}_{k=1}^K$ are multiple possible predictions for agent $i$, where $\hat{\mathbf{y}}_k = \hat{\mathbf{X}}^i(t_h + 1 \leq t \leq t_h + t_f, M) \in \mathbb{R}^{t_f \times M}$ denotes one of the expected model prediction.
Formally, we aim at building and optimizing a network $\mbox{Net} (\cdot)$, such that $\hat{\mathbf{Y}}^i = \mbox{Net}\left(\mathbf{X}^i, \{\mathbf{X}^j\}_{j \neq i}, \{\mathbf{I}\}\right)$.
For a clearer representation, agent superscripts ($i$) will be omitted if there are no conflicts.

\begin{table}[bp]
    \scriptsize
    \centering
    \caption{
        Architecture details of the enhanced \MODEL~and \EMODEL.
    }
    \label{tab_method_v}
    \begin{tabular}{c|c|c}
        \toprule
        Layer & Architecture & Out Shape \\

        \midrule
        Tran.
        & \makecell[l]{
            $\mathbf{X} \to \mathcal{T}(\mathbf{X}) = \mathbf{\mathcal{S}}$
        }
        & $(\mathcal{N}_h, \mathcal{M})$ 
        \\
        
        \hline
        MLP$_t$
        & \makecell[l]{
            $\mathbf{\mathcal{S}} \to \mbox{fc}(64, \mbox{ReLU})$ 
            $\to \mbox{fc}(64, \tanh) \to \mathbf{f}_t$ 
        }
        & $(\mathcal{N}_h, 64)$
        \\

        \hline
        MLP$_i$
        & \makecell[l]{
            $\mathbf{z} \to \mbox{fc}(64, \mbox{ReLU})$ 
            $\to \mbox{fc}(64, \tanh) \to \mathbf{f}_i$ 
        }
        & $(\mathcal{N}_h, 64)$
        \\

        \hline
        T$_k$ 
        & \makecell[l]{
            $[\mathbf{f}_t, \mathbf{f}_i] \to$ TEncoder ($128$) $\to \mathbf{f}_k'$;\\
            $\mathbf{f}_k', \mathbf{\mathcal{S}} \to$ 
            TDecoder ($128$) $\to \mathbf{f}_k''$\\
        }
        & $(\mathcal{N}_h, 128)$
        \\

        \hline
        MLP$_e$
        & \makecell[l]{
            $\mathbf{f}_k'' \to $ fc($128, \tanh$) $\to$ fc($128$) $\to \mathbf{f}$
        } 
        & $(\mathcal{N}_h, 128)$
        \\

        \hline
        MLP$_{d}$ 
        & \makecell[l]{
            $\mathbf{f} \to \mbox{fc}(128, \mbox{ReLU}) \to \mbox{fc}(128, \mbox{ReLU})$ \\
            \hphantom{$\mathbf{f}$} $\to \mbox{fc}(MN_{key})$ \\
            \hphantom{$\mathbf{f}$} $\to \mbox{Reshape}~(N_{key} , M) \to \hat{\mathbf{\mathcal{S}}}^{key}$
        } 
        & $(\mathcal{N}_{key}, \mathcal{M})$
        \\

        \hline
        MLP$_c$
        & \makecell[l]{
            $\mathbf{C}$ $\to$ MaxPool$(5 \times 5) \to$ Flatten \\
            \hphantom{$\mathbf{C}$} $\to$ $\mbox{fc}(64N_{key}, \tanh)$  \\
            \hphantom{$\mathbf{C}$} $\to$ Reshape($N_{key}, 64$) $\to \mathbf{f}_c$
        }
        & $(N_{key}, 64)$
        \\

        \hline
        Interp.
        & \makecell[l]{
            $\mathbf{\mathcal{S}}, \hat{\mathbf{\mathcal{S}}}^{key}$
            $\to \hat{\mathbf{\mathcal{S}}}^{key}_{l}$
        }
        & $(\mathcal{N}_p, \mathcal{M})$
        \\

        \hline
        T$_i$
        & \makecell[l]{
            $[\mathbf{f}_t^{key}, \mathbf{f}_c] \to$ TEncoder ($128$) $\to \mathbf{f}_i'$;\\
            $\mathbf{f}_i', \hat{\mathbf{\mathcal{S}}}^{key}_{l} \to$ 
            TDecoder ($\mathcal{M}$) $\to \hat{\mathbf{\mathcal{S}}}$
        }
        & $(\mathcal{N}_p, \mathcal{M})$
        \\

        \hline
        Inv-Tran.
        & \makecell[l]{
            $\hat{\mathbf{\mathcal{S}}} \to \mathcal{T}^{-1}(\hat{\mathbf{\mathcal{S}}})[t_h:] \to \hat{\mathbf{Y}}$
        }
        & $(t_f, M)$ 
        \\

        \midrule
        \makecell{MLP$_{bi}$ \\ (\EMODEL)}
        & \makecell[l]{
            $\mathbf{f}_\alpha \otimes \mathbf{f}_\alpha$ $\to \mbox{MaxPool}(2\times 2) \to \mbox{Flatten}$\\
            \hphantom{$\mathbf{f}_t \otimes \mathbf{f}_t$} $\to \mbox{fc}(64, \tanh) \to \mathbf{f}_R$
        }
        & $(\mathcal{N}_h, 64)$
        \\
        
        \bottomrule
    \end{tabular}
\end{table}

\textbf{Keypoints Estimation Sub-network.}
The coarse-level keypoints estimation sub-network aims to forecast agents' keypoint spectrums with a lower spatio-temporal resolution.
Similar to the original \MODEL, we first apply transform $\mathcal{T}$ on \emph{each dimension} of trajectory $\mathbf{X}$ (we call the ``trajectory-dimension'') to obtain the corresponding $\mathcal{N}$-point $\mathcal{M}$-dimensional spectrum $\mathbf{\mathcal{S}} \in \mathbb{R}^{\mathcal{N} \times \mathcal{M}}$, \IE,
\begin{equation}
    \label{eq_MD_spectrum}
        \mathbf{\mathcal{S}}
            = \mathcal{T}(\mathbf{X})
            = \left(\mathcal{T}(\mathbf{X}_1), \mathcal{T}(\mathbf{X}_2), ..., \mathcal{T}(\mathbf{X}_M)\right).
\end{equation}
Here, $\mathcal{N}$ and $\mathcal{M}$ are the number of points and dimensions of the transformed trajectory spectrum $\mathbf{\mathcal{S}}$, which can be represented by mapping functions $f_{(\mathcal{T}, n)}$ and $f_{(\mathcal{T}, m)}$:
\begin{equation}
    \mathcal{N} = f_{(\mathcal{T}, n)}(N, M), \quad
    \mathcal{M} = f_{(\mathcal{T}, m)}(N, M).
\end{equation}

In our experiments, we use two different transforms, including Haar wavelet and DFT, to further verify the impact of different transforms on the trajectory prediction performance.
The above mapping functions are determined by different transforms themselves.
Note that we regard all values in this spectrum matrix as real numbers for the convenience of the calculation in neural networks.
If the transform yields a set of complex values, we will split them into two sets of real numbers.
Take DFT as an example.
We will split the spectrum into two parts, like amplitudes and phases or real and imaginary.
Thus, we have $\mathcal{N} = N$ and $\mathcal{M} = 2M$ for Fourier specturms.
In addition, we have $\mathcal{N} = N/2$ and $\mathcal{M} = 2M$ for Haar transform.
Sepcifically, we denote $\mathcal{N}_h = f_{(\mathcal{T}, n)}(t_h, M)$, and $\mathcal{N}_f = f_{(\mathcal{T}, n)}(t_f, M)$.


Then, we employ an embedding MLP (the MLP$_t$ in \TABLE{tab_method_v}) to embed agents' observed trajectory spectrums into the high-dimensional $\mathbf{f}_t$.
We sample and embed the noise vector $\mathbf{z} \sim \mathcal{N}(\mathbf{0}, \mathbf{I})$ and concatenate the corresponding random representation $\mathbf{f}_i$ to the $\mathbf{f}_t$ to generate stochastic predictions.
The encoder for these noise vectors (MLP$_i$) has the same structure as the MLP$_t$, but their weights are not shared.
We combine the above representations to obtain the embedded spectrum representation $\mathbf{f}_e \in \mathbb{R}^{\mathcal{N}_h \times 128}$.
Formally,
\begin{equation}
    \label{eq_2}
    \mathbf{f}_e = \left[\mathbf{f}_t, \mathbf{f}_i\right] = \left[
        \mbox{MLP}_t(\mathbf{\mathcal{S}}),
        \mbox{MLP}_i(\mathbf{z})
    \right].
\end{equation}

Then, we use a Transformer\cite{vaswani2017attention} T$_{k}$ to encode agents' behaviors.
The embedded vectors $\mathbf{f}_e$ will be fed into the Transformer Encoder to calculate self-attention features.
Transformer T$_{k}$ is used as a feature extractor and does not contain the output layer.
Spectrums $\mathbf{\mathcal{S}}$ will query these attention features in the Transformer Decoder to compute the final behavior representations.
We employ another encoder MLP (MLP$_e$) to aggregate features at different frequency nodes \cite{wong2021msn}, thus inferring the behavior feature $\mathbf{f}$.
Formally,
\begin{equation}
    \label{eq_3}
    \begin{aligned}
        \mathbf{f} & = \mbox{MLP}_e \left( \mbox{T}_{k}(\mathbf{f}_e, \mathbf{\mathcal{S}}) \right) \in \mathbb{R}^{\mathcal{N}_h \times 128}.
    \end{aligned}
\end{equation}

Given the number of keypoints $N_{key}$, it finally utilizes a decoder MLP (MLP$_{d}$) to predict the $\mathcal{N}_{key}$-point \textbf{keypoint spectrums} $\hat{\mathbf{\mathcal{S}}}^{key} \in \mathbb{R}^{\mathcal{N}_{key} \times \mathcal{M}}$ ($\mathcal{N}_{key} = f_{(\mathcal{T}, n)}(N_{key}, M)$):
\begin{equation}
    \label{eq_keypointsprediction}
    \hat{\mathbf{\mathcal{S}}}^{key} = \mbox{MLP}_{d}(\mathbf{f}),
\end{equation}
Note that we still call $N_{key}$ \textbf{the number of spectrum keypoints} to avoid confusion.
Then, the Inverse Transform $\mathcal{T}^{-1}$ will be applied to obtain the corresponding spatial keypoints $\hat{\mathbf{y}}^{key} \in \mathbb{R}^{N_{key} \times M}$, where
\begin{equation}
    \hat{\mathbf{y}}^{key} = \mbox{IDFT}\left(\hat{\mathbf{\mathcal{S}}}^{key}\right).
\end{equation}

When training the sub-network, the groundtruth key spatial observations $\mathbf{y}^{key}$ will be used as the supervision to make it learn to predict the keypoint spectrums $\hat{\mathbf{\mathcal{S}}}^{key}$.
It is obtained by sampling the groundtruth $\mathbf{y}$ on $N_{key}$ key temporal moments.
Formally,
\begin{equation}
    \mathbf{y}^{key} = \left(
        \mathbf{p}_{t_1^{key}},
        \mathbf{p}_{t_2^{key}},
        ...,
        \mathbf{p}_{t_{N_{key}}^{key}}
    \right)^\top \in \mathbb{R}^{N_{key} \times M}.
\end{equation}
In our experiments, these key temporal moments will be sampled at a uniform interval from the prediction period.
The network variables will be tuned by minimizing the average Euclidean distance between $\mathbf{y}^{key}$ and the predicted $\hat{\mathbf{y}}^{key}$, \IE, the \textbf{Average Keypoints Loss} ($\mathcal{L}_{\mbox{AKL}}$):
\begin{equation}
    \label{eq_loss_1}
    \mathcal{L}_{\mbox{AKL}} = \Vert \hat{\mathbf{y}}^{key} - \mathbf{y}^{key} \Vert_2.
\end{equation}

\begin{figure*}[bt]
    \centering
    \includegraphics[width=1.0\linewidth]{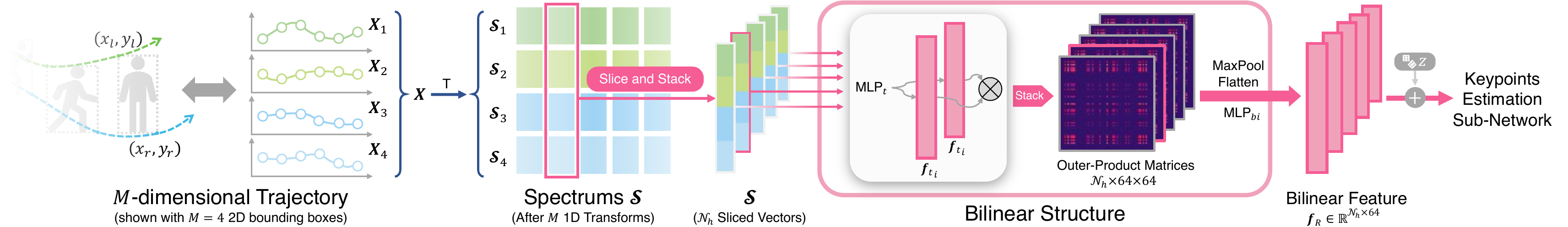}
    \caption{
        Bilinear Structure in \EMODEL.
        It takes spectrums $\mathbf{\mathcal{S}} \in \mathbb{R}^{\mathcal{N}_h \times \mathcal{M}}$ as the input, and finally outputs the refined spectrum features $\mathbf{f}_R \in \mathbb{R}^{\mathcal{N}_h \times 64}$.
    }
    \label{fig_method_bilinear}
\end{figure*}


\textbf{Spectrum Interpolation Sub-network.}
The fine-level spectrum interpolation sub-network reconstructs the complete trajectory spectrums from the formerly predicted keypoint spectrums with a higher spatio-temporal resolution.
It takes the predicted $\mathcal{N}_{key}$-point keypoint spectrums $\hat{\mathbf{\mathcal{S}}}^{key}$ as one of the primary inputs, thus learning possible interactive behaviors from spectrum biases.
Similar to \EQUA{eq_2}, we have:
\begin{equation}
    \label{eq_keydft}
    \mathbf{f}^{key}_t = \mbox{MLP}_t\left(\hat{\mathbf{\mathcal{S}}}^{key}\right) \in \mathbb{R}^{\mathcal{N}_{key} \times 64}.
\end{equation}
Note that MLP$_t$ in \EQUA{eq_2} and \EQUA{eq_keydft} do not share weights.
Then, we use a context MLP (MLP$_c$) to encode the interaction representation $\mathbf{C}$ \cite{xia2022cscnet} (which encodes social interactions and scene constraints together in an energy map way) into the context feature
$
    \mathbf{f}_c = \mbox{MLP}_c(\mathbf{C}).
$

Then, this sub-network uses a similar Transformer (called the Interpolation Transformer, T$_i$) to forecast the \textbf{complete spectrums} $\hat{\mathbf{\mathcal{S}}} \in \mathbb{R}^{\mathcal{N}_p \times \mathcal{M}}$, which is supposed to be the trajectory spectrums of the entire observed trajectories and forecasted trajectories on $t_h + t_f$ time steps, \IE, $\mathcal{N}_p = f_{(\mathcal{T}, n)}(t_h + t_f, M)$.
In particular, to adapt this sub-network to different keypoint locations, we linearly interpolate keypoint spectrums to let them have the same number of frequency portions as the final predicted ones.
This linear interpolation layer is denoted as ``Interp'' in \TABLE{tab_method_v}.
We use the interpolated spectrum $\hat{\mathbf{\mathcal{S}}}^{key}_l$ as the actual transformer input.
The $\mathbf{f}_e^{key} = [\mathbf{f}^{key}_t, \mathbf{f}_c]$ will be fed into the Transformer Encoder to calculate self-attention.
After that, spectrums $\hat{\mathbf{\mathcal{S}}}^{key}_l$ will query self-attention features in the Transformer Decoder to infer the predicted spectrums.
Formally,
\begin{equation}
    \label{eq_6}
    \hat{\mathbf{\mathcal{S}}} = \mbox{T}_i\left(\mathbf{f}_e^{key}, \hat{\mathbf{\mathcal{S}}}^{key}_l\right) \in \mathbb{R}^{\mathcal{N}_p \times \mathcal{M}}.
\end{equation}
Then, the inverse transform $\mathcal{T}^{-1}$ will be applied to obtain the final reconstructed trajectory $\hat{\mathbf{y}}_o$ for all $1 \leq t \leq t_h + t_f$ time steps.
By slicing this whole reconstructed trajectories, we have the $M$-dimensional predictions $\hat{\mathbf{y}} \in \mathbb{R}^{t_f \times M}$:
\begin{equation}
    \hat{\mathbf{y}}_o = \mathcal{T}^{-1} \left(\hat{\mathbf{\mathcal{S}}}\right),\quad \hat{\mathbf{y}} = \hat{\mathbf{y}}_o [t_h:],
\end{equation}
where the $[t_h:]$ indicates the slicing operation on tensors.

This sub-network learns to interpolate the ``key'' trajectory spectrums into the complete trajectory spectrums, thus reflecting agents' fine interactive details by predicting the remaining spectrum portions.
Similar to the keypoints estimation sub-network, its trainable variables will be tuned through the \textbf{Average Point-wise Loss} ($\mathcal{L}_{\mbox{APL}}$):
\begin{equation}
    \label{eq_loss_2}
    \mathcal{L}_{\mbox{APL}} = \Vert \hat{\mathbf{y}} - \mathbf{y} \Vert_2.
\end{equation}


\subsection{\EMODEL}

\MODEL~can not consider the interactions within heterogeneous trajectories, which we call ``dimension-wise interactions''.
The \EMODEL~is proposed to address this limitation.

\textbf{Dimension-wise Interaction and Its Graph View.}
Heterogeneous trajectories are made up of multiple $M$-dimensional trajectories.
Each $M$-dimensional trajectory consists of $M$ trajectory-dimensions.
However, most current approaches have not explicitly focused on the relationships between different trajectory dimensions.
The dimension-wise interactions are actually interactions between different trajectory-dimensions, and each interaction participator is a single trajectory-dimension.
Given an undirected graph $\mathbf{G}(t) = (\mathbf{V}(t), \mathbf{E}(t))$ with time variable $t$, where $\mathbf{V}(t) = \left\{\mathbf{g}_{m}(t)\right\}_{m=1}^{M}$ is the set of vertices that contains all $M$-dimensional position information of an agent at time $t$, and $\mathbf{E}(t)$ is the set of edges, which represents the connections between these vertices at time $t$, to establish these connections, we actually need to learn the adjacency matrix
\begin{equation}
    \mathbf{A}(t) = \begin{pmatrix}
        \mathbf{W}_{1, 1}(t) & \cdots & \mathbf{W}_{1, M}(t) \\
        \vdots & \ddots & \vdots \\
        \mathbf{W}_{M, 1}(t) & \cdots & \mathbf{W}_{M, M}(t) \\
    \end{pmatrix},
\end{equation}
so that the effect of dimension-wise interactions onto the future trajectories can be gathered into the vertex of each single trajectory-dimension into the interactive $\mathbf{g}'_m (t)$, \IE,
\begin{equation}
    \label{eq_method_graphConv}
    \mathbf{g}'_m (t) = \sigma\left( \sum_{n=1}^{M} \mathbf{W}_{m, n}(t) e(\mathbf{g}_m (t), \mathbf{g}_n (t)) \right).
\end{equation}
Here, $e(\cdot)$ represents the calculation on edges, and $\sigma(\cdot)$ denotes the non-linear gather operation.
Thus, we need to determine each $\mathbf{g}_m$, $e$, $\sigma$, and $A(t)$ simultaneously to model dimension-wise interactions, which is obversely challenging since each of them may vary over time $t$.

\textbf{Vanilla Bilinear Structures.}
A natural thought is to model these dimension-wise interactions in the frequency domain so that the time-frequency properties of each trajectory-dimension could be better represented.
We introduce the bilinear structure to model dimension-wise interactions as well as fuse ``dimension-wise interactions'' and ``time-frequency response'' factors simultaneously.
Following \cite{lin2017bilinear}, a bilinear structure consists of a quadruple
\begin{equation}
    \mathbf{\mathcal{B}} = (f_A, f_B, \mathcal{P}, \mathcal{E}).
\end{equation}
Here, $f_A$ and $f_B$ are feature functions, $\mathcal{P}$ is the max pooling function, and $\mathcal{E}$ represents an encoding function.

In the two-factor tasks, denote the set of features of one factor as $\mathbf{\mathcal{S}}_1$, and the other as $\mathbf{\mathcal{S}}_2$.
A feature function is a mapping $f: \mathbf{\mathcal{S}}_1 \times \mathbf{\mathcal{S}}_2 \to \mathbb{R}^{K \times D}$.
It takes the two-factor representations $\mathbf{f}_{s1} \in \mathbf{\mathcal{S}}_1$ and $\mathbf{f}_{s2} \in \mathbf{\mathcal{S}}_2$ as the input and finally outputs the corresponding bilinear features $\mathbf{f}_b \in \mathbb{R}^{K \times D}$.
Features for these two factors are combined at each location using the matrix outer product.
Formally,
\begin{equation}
    \label{eq_outerproduct}
    \mathbf{f}_b = \mbox{bilinear}(\mathbf{f}_{s1}, \mathbf{f}_{s2}, f_A, f_B) = f_A(\mathbf{f}_{s1}, \mathbf{f}_{s2}) \otimes f_B(\mathbf{f}_{s1}, \mathbf{f}_{s2}).
\end{equation}
The final encoded bilinear feature $\mathbf{f}_R$ is obtained by applying the max pooling and encoding operations, \IE,
\begin{equation}
    \label{eq_bilinear}
    \mathbf{f}_R = \mathcal{E} \left( \mathcal{P} \left( \mathbf{f}_b \right)\right).
\end{equation}

\textbf{Bilinear Structure in \EMODEL.}
Now we describe how to apply the above vanilla bilinear structure in \EMODEL.
\FIG{fig_method_bilinear} shows its overall structure.
In \EMODEL, both $\mathbf{\mathcal{S}}_1$ and $\mathbf{\mathcal{S}}_2$ are the spectral representation of the $M$-dimensional trajectory $\mathbf{X}$, \IE, we have $\mathbf{f}_{s1} = \mathbf{f}_{s2} = \mathcal{S}$, and $f_A = f_B = \mbox{MLP}_t$.
Thus, according to \EQUA{eq_outerproduct}, we have the \textbf{Outer-Product Matrix}
\begin{align}
    \mathbf{R} &= \mathbf{f}_b = \mathbf{f}_\alpha \otimes \mathbf{f}_\alpha \in \mathbb{R}^{\mathcal{N}_h \times 64 \times 64}, \\
    \mbox{where~~}\mathbf{f}_\alpha &= \left({\mathbf{f}_{\alpha, 1}}, {\mathbf{f}_{\alpha, 2}}, ..., {\mathbf{f}_{\alpha, \mathcal{N}_h}}\right)^{\top} = \mbox{MLP}_t(\mathbf{\mathcal{S}}), \\
    \mbox{and~~}\mathbf{R}[i] &= {\mathbf{f}_{\alpha, i}} {{\mathbf{f}_{\alpha, i}}}^{\top} \in \mathbb{R}^{64 \times 64},~~1\leq i \leq \mathcal{N}_h.
\end{align}
Here, $\left\{{\mathbf{f}_{\alpha, i}} \right\}_{i=1}^{\mathcal{N}_h}$ are obtained by slicing the spectrum $\mathbf{\mathcal{S}}$ vertically to the frequency axis and feeding them to MLP$_t$.
During this process, the element-wise outer product ${\mathbf{f}_{\alpha, i}} {{\mathbf{f}_{\alpha, i}}}^{\top}$ on the batch input $\{{\mathbf{f}_{\alpha, i}}\}$ could also measure similarities of different dimensions, which plays similar roles to our expected $\mathbf{A}$ and $e$ simultaneously in \EQUA{eq_method_graphConv}.

Then, we only need to construct the $\sigma$ function to model relations between factors ``time-frequency response'' and ``dimension-wise interactions''.
According to \EQUA{eq_bilinear}, the max pooling and a new encoder MLP (named MLP$_{bi}$, structures in \TABLE{tab_method_v}) are applied on the original bilinear feature $\mathbf{f}_b$ (also named the outer-product matrix $\mathbf{R}$) to encode the final interactive bilinear features $\mathbf{f}_R$.
Formally,
\begin{equation}
    \label{eq_outerproduct_matric}
    \mathbf{f}_R = \mbox{MLP}_{bi} \left( \mbox{Flatten} \left( \mbox{MaxPool}(\mathbf{R}) \right) \right).
\end{equation}
It can be treated as a ``frequency-bilinear'' version of \EQUA{eq_method_graphConv}, which enables the network to express the time-frequency characteristics of each trajectory-dimension better while avoiding the complex computations in graph-like structures.

The \EMODEL~takes bilinear features $\mathbf{f}_R$ to replace the original $\mathbf{f}_t$ in the keypoints estimation sub-network (\EQUA{eq_2}) to achieve the goal of modeling dimension-wise interactions and time-frequency response simultaneously.
Other network computations are the same as the original \MODEL.



\section{Experiments}

\subsection{Settings}

We evaluate models on multiple datasets:
(a) \textbf{ETH-UCY (2D Coordinate).}
ETH\cite{pellegrini2009youll}-UCY\cite{lerner2007crowds} contains several bird-eye videos captured in pedestrian walking scenes, including eth, hotel, univ, zara1, and zara2.
The annotations are 2D coordinates in meters.
Researchers \cite{alahi2016social,gupta2018social} mostly train and evaluate their models with the ``leave-one-out'' strategy \cite{alahi2016social} on this dataset.
\footnote{
    Dataset files used to train and validate on ETH-UCY are the same as \cite{zhang2020social}.
    In particular, the sampling interval on the eth subset is 6 frames to match the actual ``3.2s-to-4.8s'' setting.
}
(b) \textbf{Stanford Drone Dataset (2D Coordinate).}
The Stanford Drone Dataset \cite{robicquet2016learning} (SDD) has 60 bird-view videos captured by drones.
More than 11,000 agents with different categories are annotated with 2D bounding boxes in pixels.
Notably, most existing methods only use the 2-dimensional center points to train and validate their trajectory prediction models rather than the 4-dimensional bounding box points.
We train our model with 60\% data (36 clips), validate and test with 20\% (12 clips) respectively \SIMAUGCITE\GARDENCITE.
\footnote{
    Dataset splits used to train and validate on SDD are the same as \cite{liang2020simaug}, which includes all categories of agent annotations.
}
(c) \textbf{Stanford Drone Dataset (2D Bounding Box).}
The original SDD is labeled with bounding box positions.
We treat SDD labeled with 4-dimensional 2D bounding boxes as a new dataset to validate the performance of 2D bounding box prediction.
Splits and settings are the same as SDD.
(d) \textbf{nuScenes (3D Bounding Box).}
The nuScenes Dataset \cite{caesar2019nuscenes} is a large-scale real-world dataset of 1000 driving scenes collected in the urban cities of Boston and Singapore.
This dataset is used to validate the performance of 3D bounding box prediction, different from previous approaches that predict 2D coordinates like \cite{yuan2021agentformer}.
We only use two 3D corner points when training and validating, \IE, forecasting a 6-dimensional trajectory.
Following \cite{saadatnejad2022pedestrian}, we use 550 scenes to train, 150 to validate, and the other 150 to test.
(e) \textbf{Human3.6M (3D Skeleton).}
The Human3.6M\cite{ionescu2013human3,ionescu2011latent} dataset is a large-scale dataset with 3.6 million 3D human poses, performed by 11 professional actors in 15 scenarios (such as discussion, smoking, taking photos, and talking on the phone).
Each person is annotated with 51 points (17 joints in 3D).
We use subjects $\{1, 6, 7, 8, 9\}$ to train, subjects $\{11\}$ to validate, and subjects $\{5\}$ to test\cite{xu2023eqmotion}.


\textbf{Metrics.}
(a) \textbf{ADE \& FDE \cite{alahi2016social,pellegrini2009youll}.}
ADE is the average point-wise Euclidean distance between each groundtruth and predicted point, and FDE is the Euclidean distance on the last prediction step.
For $K$ $M$-dimensional predictions $\{\hat{\mathbf{y}_k}\}_{k=1}^K$ to the same agent, where $\hat{\mathbf{y}_k}$ consists of $m$ 2D or 3D points $\hat{\mathbf{p}}^i_t$, we have
\begin{equation}
    \mbox{ADE}(\mathbf{y}, \{\hat{\mathbf{y}_k}\}_{k=1}^K) = \min_k \frac{1}{mt_f} \sum_{t = t_h + 1}^{t_h + t_f} \sum_{i=1}^m \Vert \mathbf{p}^i_t - \hat{\mathbf{p}_k}^i_t \Vert_2,
\end{equation}
\begin{equation}
    \label{eq_fde}
    \mbox{FDE}(\mathbf{y}, \{\hat{\mathbf{y}_k}\}_{k=1}^K) = \min_k \frac{1}{m} \sum_{i=1}^m \Vert \mathbf{p}^i_{t_h + t_f} - \hat{\mathbf{p}_k}^i_{t_h + t_f} \Vert_2.
\end{equation}
(b) \textbf{AIoU \& FIoU \cite{saadatnejad2022pedestrian}.}
To better estimate 2D and 3D bounding box prediction performance, we compare the volume of the predicted bounding box and the ground truth, then compute the Intersection volume Over Union between them similar to ADE and FDE as AIoU and FIoU.
(c) \textbf{MPJPE.}
For 3D skeleton cases, we use FDE in \EQUA{eq_fde} as the metric to measure model performance ($m=17$).
Note that in the field of motion prediction, this metric is also more commonly known as the Mean Per Joint Position Error (MPJPE).

\textbf{Prediction types.}
Our focused trajectory types include
2D coordinates (``co'', $M = 2$),
2D bounding box (``bb'', $M = 4$),
3D bounding box (``3Dbb'', $M = 6$), and
3D human skeleton (``ske'', $M = 51$).
We also provide ``\textbf{SEP}'' variations for ablation studies, which separately predict each element (\IE, a single point in 2D coordinates, a single 2D point in 2D bboxes, a single 3D point in 3D bboxes or skeletons) and then concatenating them to obtain final predictions.

\textbf{Implementation details.}
For different prediction types, \EMODEL s' structures are the same except for the number of input dimensions $M$.
We first introduce the common settings, then specify different datasets' prediction settings.
Notably, \EMODEL~and \MODEL~will be short as ``\textbf{\EMODELSHORT}''~and ``\textbf{\MODELSHORT}''~to suit space limitations in some paragraphs.

(a) \textbf{Common Settings.}
Following \cite{zhang2020social}, each trajectory is pre-processed by moving to the origin before the model implementation.
We train \MODEL~and \EMODEL~with Adam optimizer (learning rate $=0.0003$) on one NVIDIA GeForce GTX 1080Ti card.
Models are trained with the batch size bs $=2500$ for 800 epochs on ETH-UCY and 150 epochs on SDD, SDD (2D bounding box), and nuScenes.
We employ $L = 4$ layers of encoder-decoder structure with $H = 8$ attention heads in each Transformer sub-network.
The output unit in attention layers is set to 128.
Detailed Transformer structures are listed in the supplemental material.

(b) \textbf{Dataset-Wise Settings.}
\textbf{(i) For ETH-UCY and SDD}, we predict with $M = 2$ and type ``co''.
For SDD (2D bounding box), we set $M = 4$ and predict with type ``bb''.
On these datasets, we predict agents' trajectories in future $t_f = 12$ frames according to their $t_h = 8$ frames' observations.
The frame rate is set to 2.5 fps when sampling trajectories.
In addition, we set the number of spectrum keypoints $N_{key} = 3$, and $\{t^{key}_1, t^{key}_2, t^{key}_3\} = \{t_{h} + 4, t_{h} + 8, t_{h} + 12\}$ for DFT variations.
Similarly, we set $N_{key} = 4$ and $\{t^{key}_1, t^{key}_2, t^{key}_3\, t^{key}_4\} = \{t_{h} + 3, t_{h} + 6, t_{h} + 9, t_{h} + 12\}$ for Haar variations.
\textbf{(ii) For nuScenes (3D bounding box)}, we set $M = 6$ and predict with type ``3Dbb''.
We set $\{t_h, t_f\} = \{4, 4\}$ on this dataset when forecasting trajectories annotated with 3D bounding boxes with the frame rate of 2fps.
We set $N_{key} = 2$ and $\{t^{key}_1, t^{key}_2\} = \{t_{h} + 2, t_{h} + 4\}$ for all variations.
Note that we do not use agents' interactions representations $\mathbf{C}$ in this dataset due to the scale differences.
\textbf{(iii) For Human3.6M (3D Skeleton)}, we set $M = 51$ with type ``ske''.
We sample observations with the frequency of 25Hz (\IE, the sample interval is 40ms) and use $t_h = 10$ frames (400ms) of observations to predict $t_f = 10$ future frames (400ms).
We set $N_{key} = 4$, and $\{t^{key}_1, t^{key}_2, t^{key}_3, t^{key}_4\} = \{t_h + 1, t_h + 4, t_h + 7,t_h + 10\}$.
We extend the feature dimension from 128 to 512 for each layer in the network to expand the model capacity.
Models are set to predict one deterministic trajectory for each agent.
All social interaction and scene interaction modules are also disabled since there is only one subject in each scene.

\begin{table}[tbp]
    \centering
    \caption{
        Comparisons with the \emph{best-of-20} on ETH-UCY (\emph{2D coordinates}).
        Metrics are shown in the format of ``ADE/FDE'' in meters.
    }
    \fontsize{6}{7}\selectfont{
    \label{tab_ade_ethucy}
    \begin{tabular}{
        c|
        p{0.7cm}
        p{0.7cm}
        p{0.7cm}
        p{0.7cm}
        p{0.7cm}
        p{0.7cm}
    }
        \toprule
        Models & eth $\downarrow$ & hotel $\downarrow$ & univ $\downarrow$ & zara1 $\downarrow$ & zara2 $\downarrow$ & Avg. $\downarrow$ \\
    
        \midrule

        \BIGAT\BIGATCITE & 0.69/1.29 & 0.49/1.01 & 0.55/1.32 & 0.30/0.62 & 0.36/0.75 & 0.48/1.00 \\
        \TP\TPCITE & 0.61/1.12 & 0.18/0.30 & 0.35/0.65 & 0.22/0.38 & 0.17/0.32 & 0.31/0.55 \\
        \PECNET\PECNETCITE & 0.54/0.87 & 0.18/0.24 & 0.35/0.60 & 0.22/0.39 & 0.17/0.30 & 0.29/0.48 \\
        \STAR\STARCITE & 0.56/1.11 & 0.26/0.50 & 0.52/1.13 & 0.40/0.89 & 0.52/1.13 & 0.41/0.87 \\
        \TRAJECTRONPP\TRAJECTRONPPCITE & 0.43/0.86 & \C{0.12}/0.19 & \C{0.22}/0.43 & \C{0.17}/0.32 & \C{0.12}/0.25 & 0.20/0.39 \\
        \TPNMS\TPNMSCITE & 0.52/0.89 & 0.22/0.39 & 0.55/1.13 & 0.35/0.70 & 0.27/0.56 & 0.38/0.73 \\
        \INTROVERT\INTROVERTCITE & 0.42/0.70 & \C{0.11}/0.17 & \C{0.20/0.32} & \C{0.16}/0.27 & 0.16/0.25 & 0.21/0.34 \\
        \LBEBM\LBEBMCITE & 0.30/0.52 & 0.13/0.20 & 0.27/0.52 & 0.20/0.37 & 0.15/0.29 & 0.21/0.38 \\
        \AGENTFORMER\AGENTFORMERCITE & 0.26/0.39 & \C{0.11/0.14} & 0.26/0.46 & \C{0.15/0.23} & 0.14/\C{0.23} & \C{0.18/0.29} \\
        \YNET~(TTST)\YNETCITE & 0.28/\C{0.33} & \C{0.10/0.14} & 0.24/0.41 & \C{0.17}/0.27 & \C{0.13/0.22} & \C{0.18/0.27} \\
        \STCNET\STCNETCITE & 0.64/1.18 & 0.33/0.54 & 0.39/0.74 & 0.29/0.49 & 0.26/0.45 & 0.38/0.68 \\
        \SRPAMI\SRPAMICITE & 0.44/0.79 & 0.19/0.31 & 0.50/1.05 & 0.32/0.64 & 0.27/0.54 & 0.34/0.67 \\
        \CSCNET\CSCNETCITE & 0.51/1.05 & 0.22/0.42 & 0.36/0.81 & 0.31/0.68 & 0.47/1.02 & 0.37/0.79 \\
        \MID\MIDCITE & 0.39/0.66 & 0.13/0.22 & \C{0.22}/0.45 & \C{0.17}/0.30 & 0.13/0.27 & 0.21/0.38 \\
        \SHENET\SHENETCITE & 0.41/0.61 & 0.13/0.20 & 0.25/0.43 & 0.21/0.32 & 0.15/0.26 & 0.23/0.36 \\
        \MODEL\MODELCITE & \C{0.23}/0.37 & \C{0.10/0.16} & 0.24/0.43 & 0.19/0.30 & \C{0.14/0.24} & \C{0.18/0.30}\\
        \SSL\SSLCITE & 0.69/1.37 & 0.24/0.44 & 0.51/0.93 & 0.42/0.84 & 0.34/0.67 & 0.44/0.85\\
        \SEEM\SEEMCITE & 0.62/1.20 & 0.61/1.21 & 0.50/1.04 & 0.31/0.61 & 0.36/0.68 & 0.48/0.95 \\
        \EQMOTION\EQMOTIONCITE & 0.40/0.61 & \C{0.12}/0.18 & 0.23/0.43 & 0.18/0.32 & \C{0.13}/0.23 & 0.21/0.35 \\

        \midrule
        \EMODEL-DFT
        & \C{0.25}/0.38
        & \C{0.11/0.16}
        & 0.23/0.42
        & 0.19/0.30
        & \C{0.13/0.24}
        & \C{0.18/0.30}
        \\

        \EMODEL-Haar
        & \C{0.25}/0.41
        & \C{0.10/0.15}
        & 0.23/0.43
        & 0.18/0.31
        & \C{0.13/0.24}
        & \C{0.18/0.30}
        \\
        
        \bottomrule
    \end{tabular}}
\end{table}

\begin{table}[tbp]
    \centering
    \scriptsize
    \caption{
        Comparisons with the \emph{best-of-20} on SDD (\emph{2D coordinates}).
    }
    \label{tab_ade_sdd}
    \begin{tabular}{c|c|c|c}
        \toprule
        Models & ADE/FDE $\downarrow$ & Models & ADE/FDE $\downarrow$ \\

        \midrule
        \SOPHIE\SOPHIECITE      & 16.27/29.38   &
        \GARDEN\GARDENCITE      & 14.78/27.09   \\
        \MANTRA\MANTRACITE      & 8.96/17.76    &
        \SIMAUG\SIMAUGCITE      & 12.03/23.98   \\
        \PECNET\PECNETCITE      & 9.96/15.88    &
        \LBEBM\LBEBMCITE        & 8.87/15.61    \\
        \SPECTGNN\SPECTGNNCITE  & 8.21/12.41    &
        \YNET~(TTST)\YNETCITE   & 7.85/11.85    \\
        \CSCNET\CSCNETCITE      & 14.63/26.91   &
        \MID\MIDCITE            & 7.91/14.50    \\
        \SHENET\SHENETCITE      & 9.01/13.24    &
        \MODEL\MODELCITE        & 7.12/11.39    \\

        \midrule
        \EMODELSHORT-DFT (Ours) & \C{6.57/10.49} &
        \EMODELSHORT-Haar (Ours) & \C{6.83/11.01}\\

        \bottomrule
    \end{tabular}

\end{table}

\begin{table}[tp]
    \centering
    \scriptsize
    \caption{
        Comparisons (\emph{best-of-20}) of \emph{2D bounding boxes} on SDD.
    }
    \label{tab_ade_sdd_bb}
    \begin{tabular}{c|cccc}
        \toprule
        Models
        & ADE $\downarrow$ & FDE $\downarrow$
        & AIoU $\uparrow$ & FIoU $\uparrow$ \\

        \midrule
        Linear & 27.18 & 52.40 & 0.562 & 0.401 \\
        \MSN~(SEP)\MSNCITE & 7.87 & 12.98 & 0.691 & 0.543 \\
        \MODEL~(SEP)\MODELCITE & 7.10 & 10.92 & 0.698 & 0.578 \\
        \MODEL &  6.78 & 10.73 & 0.717 & 0.601 \\

        \midrule
        \EMODEL-DFT & \C{6.62} & \C{10.57} & \C{0.725} & \C{0.604} \\
        \EMODEL-Haar & \C{6.44} & \C{10.28} & \C{0.730} & \C{0.613} \\

        \bottomrule
    \end{tabular}

\end{table}

\begin{table}[tp]
    \centering
    \scriptsize
    \caption{
        Comparisons (forecasting $k=1$ deterministic trajectory for each agent with 2 seconds' prediction horizon) of \emph{3D bounding boxes} on nuScenes.
    }
    \label{tab_ade_nuScenes}
    \begin{tabular}{c|cccc}
        \toprule
        Models & ADE $\downarrow$ & FDE $\downarrow$ & AIoU $\uparrow$ & FIoU $\uparrow$ \\

        \midrule
        Zero-Vel \cite{martinez2017human} & 4.364 & 6.952 & 0.084 & 0.065 \\
        P-LSTM \cite{saadatnejad2022pedestrian} & 1.094 & 1.985 & 0.251 & 0.110 \\
        PV-LSTM \cite{saadatnejad2022pedestrian} & 0.994 & 1.811 & 0.305 & 0.158 \\

        \midrule
        \EMODEL-DFT ($k=1$) & 0.529 & \C{0.919} & 0.602 & 0.445 \\
        \EMODEL-Haar ($k=1$) & \C{0.525} & \C{0.918} & \C{0.608} & \C{0.450} \\

        \bottomrule
    \end{tabular}
\end{table}

\begin{table*}[tbp]
    \scriptsize
    \centering
    \caption{
        Comparisons of 3D skeleton prediction ($t_h = t_f = 10$) on several Human3.6M subsets and the average metrics over all 15 subsets.
        Metrics are shown in the format of MPJPE (FDE) at 80/160/320/400 ms in millimeters.
        All models only forecast one trajectory for each agent.
    }
    \label{tab_h36m}

    \begin{tabular}{c|cccccc}
        \toprule
        Models & Sitting Down $\downarrow$ & Phone Call $\downarrow$ & Directions $\downarrow$ & Smoking $\downarrow$ & Discussion $\downarrow$ & Average (15 actions) $\downarrow$ \\
        
        \midrule
        Traj-GCN~\cite{mao2019learning} & 16.1/31.1/61.5/75.5 & 10.2/21.0/42.5/52.3 & 9.0/19.9/43.4/53.7 & 7.9/16.2/31.9/38.9 & 12.5/27.4/58.5/71.7 & 26.1/52.3/63.5/63.5 \\
        DMGNN~\cite{li2020dynamic} & 15.0/32.9/77.1/93.0 & 12.5/25.8/48.1/58.3 & 13.1/24.6/64.7/81.9 & 9.0/17.6/32.1/40.3 & 17.3/34.8/61.0/69.8 & 33.6/65.9/79.7/79.7 \\
        MSR-GCN~\cite{dang2021msr} & 16.1/31.6/62.5/76.8 & 10.1/20.7/41.5/51.3 & 8.6/19.7/43.3/53.8 & 8.0/16.3/31.3/38.2 & 12.0/26.8/57.1/69.7 & 12.1/25.6/51.6/62.9 \\
        PGBIG~\cite{ma2022progressively} & 13.9/27.9/57.4/71.5 & 8.3/18.3/38.7/48.4 & 7.2/17.6/40.9/51.5 & 6.6/14.1/28.2/34.7 & 10.0/23.8/53.6/66.7 & 10.3/22.7/47.4/58.5 \\
        SPGSN~\cite{li2022skeleton} & 14.2/27.7/56.8/70.7 & 8.7/18.3/38.7/48.5 & 7.4/17.2/39.8/\C{50.3} & 6.7/13.8/28.0/\C{34.6} & 10.4/23.8/53.6/67.1 & 10.4/22.3/47.1/58.3 \\
        EqMotion~\cite{xu2023eqmotion} & 13.0/26.5/56.2/70.7 & 7.4/16.7/36.9/47.0 & \C{6.3/15.8/38.9/50.1} & \C{5.5/11.3/23.0/29.3} & \C{8.2/18.9/42.1/53.9} & 9.1/20.1/\C{43.7/55.0} \\
        
        \midrule
        \MODEL-Haar & \C{9.5/20.9/47.7/62.1} & \C{7.0/15.5/35.3/46.1} & \C{6.9}/16.2/39.7/52.9 & \C{5.6/12.3}/28.1/36.9 & \C{9.2}/21.3/50.0/65.2 & \C{8.7/19.7}/45.5/59.5 \\
        \EMODEL-Haar & \C{9.8/21.3/47.8/61.9} & \C{7.2/15.5/34.3/44.5} & 7.0/\C{16.1/38.4}/50.7 & 5.9/12.4/\C{27.4}/35.9 & 9.4/\C{21.2/49.0/63.6} & \C{9.0/19.6/44.4/57.8} \\
    
        \bottomrule
    \end{tabular}

\end{table*}

\subsection{Comparisons to State-of-the-Art Methods}

(a) \textbf{ETH-UCY (2D Coordinate).}
As shown in \TABLE{tab_ade_ethucy}, \EMODEL s perform at the same level as the state-of-the-art works like \AGENTFORMER~and \YNET.
Compared to the newly-published \SHENET, \EMODEL-DFT has improved 26.1\% and 22.2\% in ADE and FDE.
A recent work \EQMOTION~could predict multiple tasks, including 2D trajectories and human skeletons.
The proposed models outperform \EQMOTION~by 19.0\% and 20.0\% in ADE and FDE, respectively.
In short, the performances of \MODEL~and \EMODEL s are strongly comparable to these state-of-the-art methods on ETH-UCY.

(b) \textbf{SDD (2D Coordinate).}
As listed in \TABLE{tab_ade_sdd}, both \MODEL~and \EMODEL s outperform most state-of-the-art methods.
Compared with the current state-of-the-art method \YNET, \EMODEL-DFT achieves about 16.3\% and 11.5\% improvement in ADE and FDE, respectively.
Compared to the recent \SHENET, \EMODEL-DFT shows a significant improvement of 27.1\% in ADE and 20.8\% in FDE.
Although not performing as well as \EMODEL-DFT, \EMODEL-Haar also shows strong competitiveness.
Compared to \YNET, it still has a 13.0\% better ADE and a 7.0\% better FDE.

(c) \textbf{SDD (2D Bounding Box).}
Since few researchers have tested the 2D bounding box prediction performance on SDD, we use the ``SEP'' variations to verify their effectiveness in forecasting 2D bounding boxes.
As shown in \TABLE{tab_ade_sdd_bb}, both \MODEL~and \EMODEL~show better prediction performance on 2D bounding boxes.
Compared to \MSN~(SEP), \MODEL~(SEP) obtains a remarkable improvement of 9.8\% in ADE, 15.9\% in FDE, and a slight improvement of 0.07 in AIoU and 0.035 in FIoU, which demonstrates its effectiveness.
Moreover, \EMODEL-Haar performs better than \MODEL~with an improvement of 5.0\% in ADE and 4.2\% in FDE, which shows unique advantages with different transforms.

(d) \textbf{nuScenes (3D Bounding Box).}
As shown in \TABLE{tab_ade_nuScenes}, compared to PV-LSTM, the deterministic \EMODEL-DFT ($k = 1$) has achieved significant improvement of 46.8\% in ADE, 49.3\% in FDE, 0.297 in AIoU, and 0.287 in FIoU.
In addition, the deterministic Haar variation slightly outperforms the DFT variation with about 0.8\% ADE, 1.0\% AIoU, and 1.1\% FIoU.
These comparisons also indicate the superiority of \EMODEL~in forecasting 3D bounding boxes.

(e) \textbf{Human3.6M (3D Skeleton).}
We use this more challenging dataset to further evaluate the proposed models' modeling capacity.
As shown in \TABLE{tab_h36m}, \EMODEL-Haar has achieved considerable performance compared to current motion prediction methods, even though it does not consider any prior about human structures or postures.
Especially for the action ``Sitting Down'', \EMODEL-Haar outperforms \EQMOTION~for up to 12.22\% MPJPE at 400ms.
In other actions, the proposed \MODEL~and \EMODEL~also perform well, meeting or even outperforming several concurrent state-of-the-art works.
Although the proposed method does not achieve the best performance on the average of this dataset, which is roughly 5.09\% worse than \EQMOTION, these results have illustrated their effectiveness for processing high dimensional trajectories such as human skeletons.


\subsection{Quantitative Analysis and Ablation Studies}
\label{quantitative_analysis}

\textbf{Why Transform Trajectories?}
A central motivation of this manuscript is to transform trajectories so that the trajectory prediction task can be implemented in the corresponding transformed domain.
In this part, we discuss the quantitative impacts of transforms when forecasting.

\begin{figure}[tb]
    \centering
    \includegraphics[width=8.8cm]{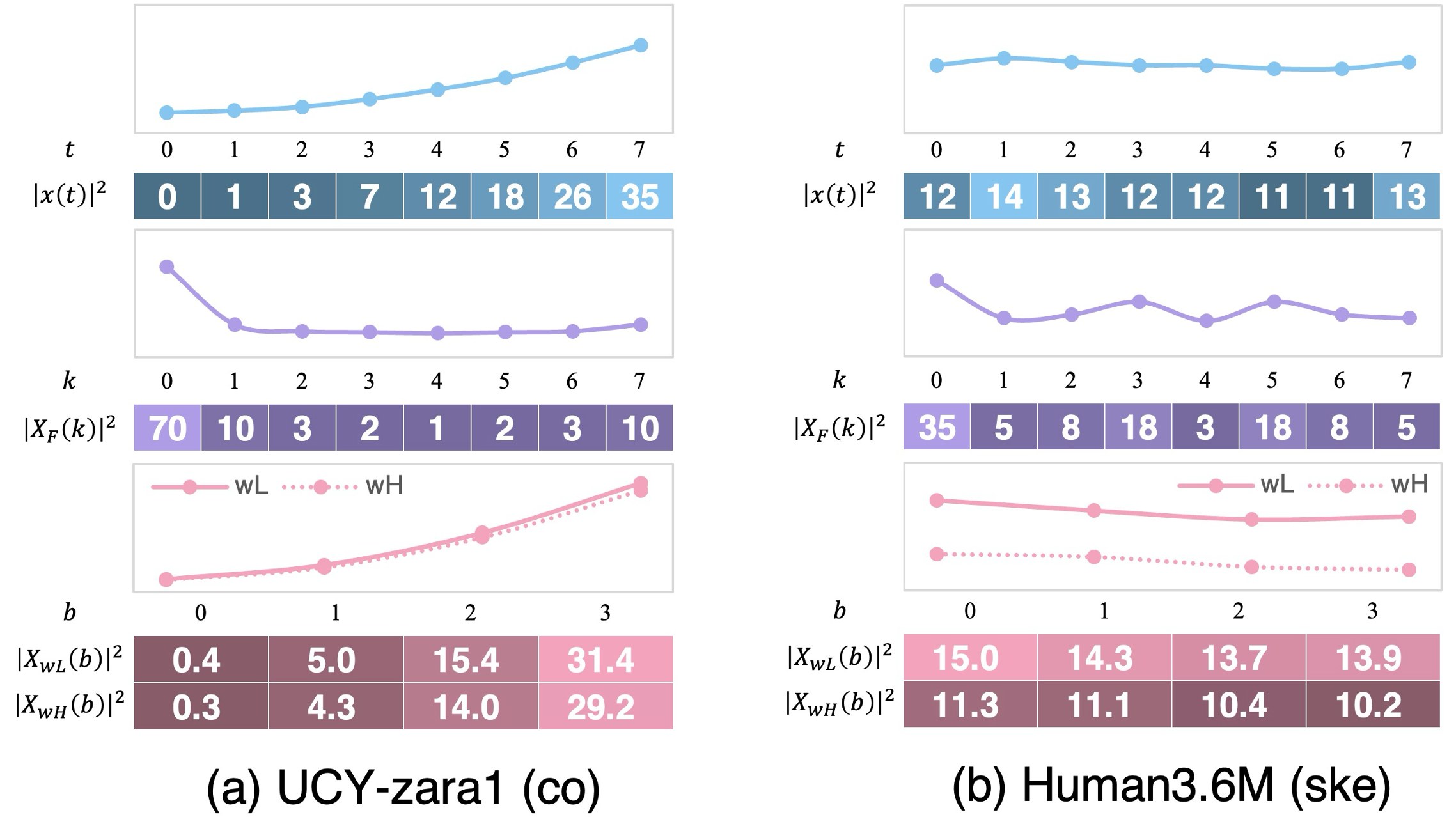}
    \caption{
        The average energy percentages of the observed trajectories on different moments and different frequency components.
    }
    \label{fig_exp_energy}
\end{figure}

\newcolumntype{P}[1]{>{\centering\arraybackslash}p{#1}}

\begin{table}[btp]
    \centering
    \caption{
        Ablation Studies (2D coordinate prediction) with \emph{best-of-20} on ETH-UCY.
        Models with ``(D)'' only predict one deterministic trajectory.
        ``T'' indicates the used transforms.
        ``S2'' indicates whether the variation uses the second stage spectrum interpolation sub-network.
    }
    \fontsize{6}{7}\selectfont{
    \label{tab_ab_ethucy}
    \begin{tabular}{
        P{0.4cm}|
        P{0.5cm}|
        P{1.3cm}|
        p{0.65cm}
        p{0.65cm}
        p{0.65cm}
        p{0.65cm}
        p{0.65cm}
    }
        \toprule
        No. & Model & (T,$N_{key}$)~~~S2 &
        eth$\downarrow$ & hotel$\downarrow$ & univ1$\downarrow$ & zara1$\downarrow$ & zara2$\downarrow$ \\

        \midrule
        CO1 & \MODELSHORT~(D) & (None, 0)~~~-&
        0.65/1.30 & 0.25/0.44 & 0.77/1.39 & 0.48/0.97 & 0.40/0.77 \\

        CO2 & \MODELSHORT~(D) & (DFT, 0)~~~-&
        0.62/1.15 & 0.22/0.37 & 0.55/1.10 & 0.45/0.92 & 0.36/0.74 \\

        CO3 & \MODELSHORT~(D) & (Haar, 0)~~~-&
        0.63/1.15 & 0.23/0.42 & 0.59/1.17 & 0.43/0.90 & 0.33/0.70 \\

        \midrule
        CO4 & \MODELSHORT & (None, 1)~~~\cm &
        0.29/0.47 & 0.12/0.18 & 0.23/0.38 & 0.25/0.37 & 0.17/0.29 \\

        CO5 & \MODELSHORT & (DFT, 1)~~~\cm &
        0.23/0.39 & 0.12/0.17 & 0.23/0.37 & 0.22/0.32 & 0.16/0.26 \\

        CO6 & \MODELSHORT & (DFT, 4)~~~\cm &
        0.24/0.38 & 0.12/0.17 & 0.22/0.36 & 0.21/0.32 & 0.16/0.26 \\

        CO7 & \MODELSHORT & (DFT, 6)~~~\cm &
        0.29/0.49 & 0.13/0.17 & 0.23/0.38 & 0.26/0.37 & 0.21/0.31 \\

        \midrule
        CO8 & \MODELSHORT~(SEP) & (DFT, 3)~~~\cm &
        0.26/0.40 & 0.11/0.17 & 0.22/0.41 & 0.21/0.39 & 0.15/0.27 \\

        CO9 & \MODELSHORT & (DFT, 3)~~~\cm &
        0.23/0.37 & 0.11/0.16 & 0.21/0.35 & 0.19/0.30 & 0.14/0.24 \\

        CO10 & \MODELSHORT & (DFT, 3)~~~\xm &
        0.24/0.37 & 0.12/0.16 & 0.23/0.35 & 0.21/0.30 & 0.15/0.25 \\

        \midrule
        CO11 & \EMODELSHORT & (DFT, 3)~~~\cm &
        0.25/0.38 & 0.10/0.16 & 0.20/0.34 & 0.19/0.30 & 0.13/0.24 \\
        
        CO12 & \EMODELSHORT & (DFT, 3)~~~\xm &
        0.27/0.38 & 0.11/0.16 & 0.21/0.34 & 0.19/0.30 & 0.14/0.24 \\

        \midrule
        CO13 & \EMODELSHORT & (Haar, 4)~~~\cm &
        0.25/0.41 & 0.10/0.15 & 0.20/0.35 & 0.18/0.31 & 0.13/0.24 \\

        CO14 & \EMODELSHORT & (Haar, 4)~~~\xm &
        0.27/0.41 & 0.10/0.16 & 0.21/0.35 & 0.19/0.32 & 0.14/0.24 \\

        \bottomrule
    \end{tabular}}
\end{table}

(a) \textbf{Energy of Spectrums.}
We first discuss transforms (DFT and Discrete Haar transform) from the energy perspective.
In mathematical analysis, Parseval's identity is a fundamental result of the summability of the Fourier series of a function.
It could also apply to Haar transform.
Simply, in one-dimension, it says that the integral of the square of the Fourier transform of a function equals the integral of the square of the function itself, \IE,
\begin{equation}
    \label{eq_energy_dft}
    \int_{-\infty}^{\infty} |X(\xi )|^2 \mathrm{d}\xi = \int_{-\infty}^{\infty} |x(t)|^2 \mathrm{d}t,
\end{equation}
where $X(\xi )$ is the Fourier transform of $x(t)$.
It states that the energy (square sum) remains constant before and after the transform, which means that the transform does not change the amount of information but \emph{redistributes} them at different time-frequency moments.
Thus, information ``buried'' in the time domain may be easily noticed in the frequency domain.

We calculate and report the energy of trajectories and their spectrums on different scenarios in \FIG{fig_exp_energy}.
Note that we shift the start points of trajectories to the origin and report normalized energy for simple comparisons.
The changes in energy over time are relatively small.
Take \FIG{fig_exp_energy} (a) as an example, the average energy on each time step is $(0, 1\%, 3\%, 7\%, 12\%, 18\%, 26\%, 35\%)$, among which the maximum gain between adjacent steps is about 9\%.
Differently, the energy has been significantly redistributed after DFT.
The energy on the fundamental frequency ($k = 0$) reaches a staggering 70\%, much greater than all adjacent time steps.
Besides, such an enormous energy difference, like the almost 60\% decrease from $k = 0$ to $k = 1$ in the Fourier spectrum, may also bring advantages when modeling agents' preferences at different frequency scales.

The Haar transform redistributes energy in a way that is different from DFT.
In detail, as shown in \FIG{fig_exp_energy}, the high-frequency portion $X_{wH}(b)$ in the Haar spectrums owns higher percentages of energy than the Fourier spectrums.
This means the Haar transform could better capture these changes when analyzing trajectories with rapidly changing characteristics than the DFT.
Another difference between the Haar transform and DFT is that the Haar transform could consider the time-frequency joint representation, which means that the energy distribution of frequency portions in Haar spectrums may change over time.
Thus, the differentiated low or high-frequency distribution in Haar spectrums may facilitate modeling the detailed agent preferences.

In conclusion, transforms do not bring additional information but redistribute the existing information in the trajectory.
In this process of looking at the trajectory from a new ``vertical'' view, the prediction network may be able to discover different information that better responds to agents' behavioral preferences with different spectral distributions.
Hence, we apply DFT and Haar transforms on trajectories in the proposed \MODEL~and \EMODEL, thus forecasting trajectories hierarchically with spectrums.

(b) \textbf{The Choose of Spectrum Keypoints.}
It can be seen from \FIG{fig_exp_energy} that most energy is concentrated in limited frequency portions, whether in Fourier or Haar spectrums.
For example, 2 to 3 of all 8 lower-frequency portions could occupy up to 90\% energy in Fourier spectrums, and 4 frequency portions could also occupy at least 50\% energy in Haar spectrums, which are different from the energy distribution in the time domain.
This phenomenon motivates us to forecast trajectories with spectrums hierarchically, \IE, first coarsely predicting the lower-frequency portions that occupy more spectral energy and then finely complementing the left with less energy.
According to \EQUA{eq_energy_dft}, we can select the same percentages of temporal keypoints as the percentages of these energy-concentrated frequency portions as an instruction, \IE, about 25\% to 50\% temporal points as keypoints.
Correspondingly, we sample the trajectory in the time domain at uniform intervals so that the sampled trajectory has a total of $N_{key}$ points, thus making sure that the lower-frequency portions could occupy more energy in the newly sampled $N_{key}$-point spectrum, which is similar to the reconstruct process in \FIG{fig_intro_hp}.
Please refer to ``Number of Spectrum Keypoints'' for detailed validations.

\begin{figure}[tb]
    \centering
    \includegraphics[width=8.8cm]{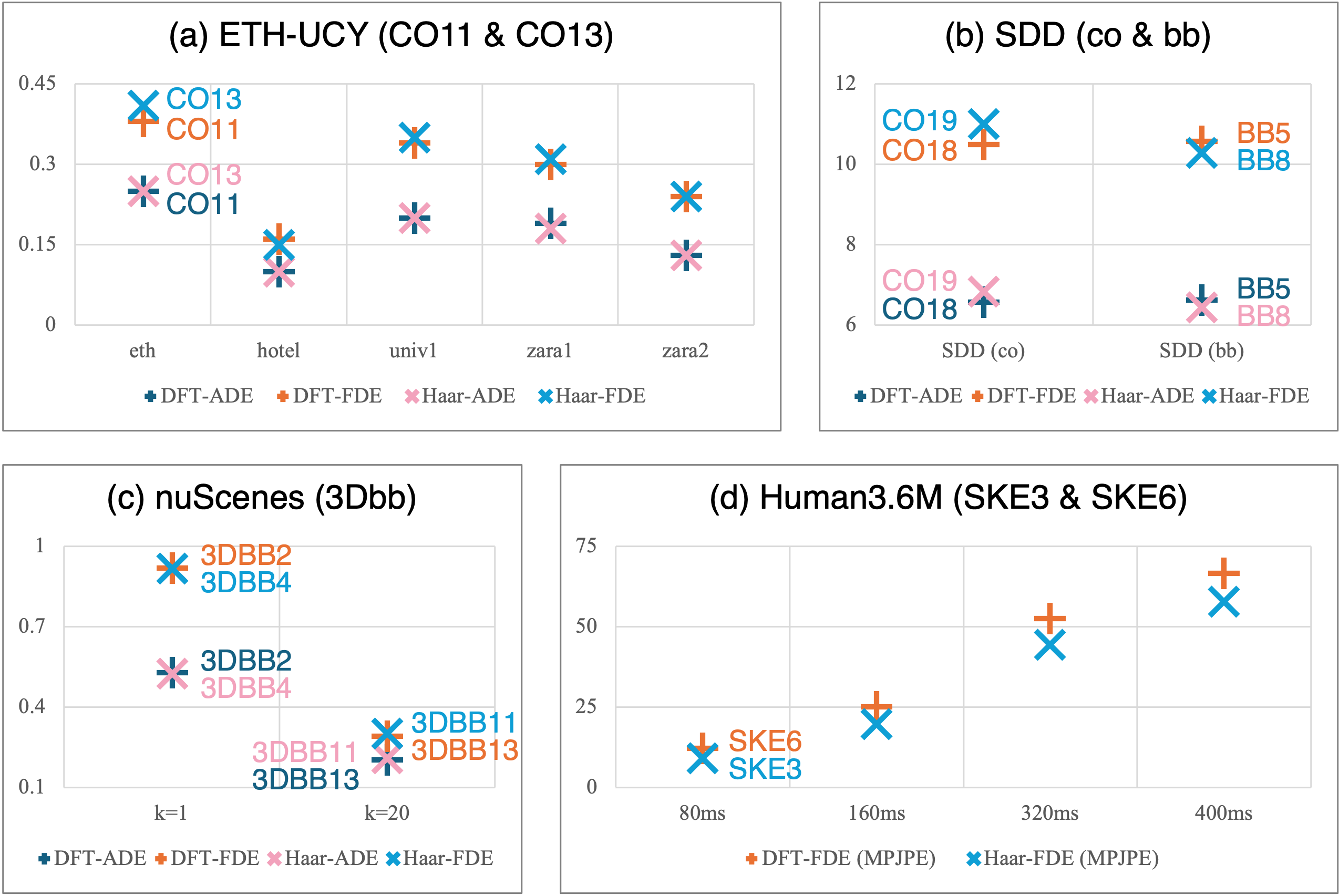}
    \caption{
        Performance comparisons of DFT (mark ``$+$'') and Haar (``$\times$'') \EMODEL~variations when forecasting different forms of trajectories.
        Lower metrics (ADE/FDE/MPJPE) indicate better prediction performances.
    }
    \label{fig_exp_transform}
\end{figure}

\begin{table}[htbp]
    \centering
    \scriptsize
    \caption{
        Ablation Studies with \emph{best-of-20} on SDD.
        2D coordinate variations are labeled with subscript ``$_{co}$'', and others are for 2D bounding boxes.
    }
    \label{tab_ab_sdd}
    \begin{tabular}{c|c|c|cccc}
        \toprule
        No. & Model & (T, $N_{key}$) & ADE $\downarrow$ & FDE $\downarrow$ & AIoU $\uparrow$ & FIoU $\uparrow$ \\

        \midrule
        CO15 & \MODELSHORT$_{co}$~(SEP) & (DFT, 3) &
        7.19 & 12.13 & \NA & \NA \\

        CO16 & \MODELSHORT$_{co}$ & (DFT, 3) & 
        7.12 & 11.39 & \NA & \NA \\

        CO17 & \MODELSHORT$_{co}$ & (Haar, 4) & 
        7.01 & 11.62 & \NA & \NA \\

        CO18 & \EMODELSHORT$_{co}$ & (DFT, 3) & 
        6.57 & 10.50 & \NA & \NA \\

        CO19 & \EMODELSHORT$_{co}$ & (Haar, 4) & 
        6.83 & 11.01 & \NA & \NA \\
        
        \midrule
        BB1 & \MODELSHORT~(SEP) & (DFT, 3) & 
        7.06 & 10.91 & 0.698 & 0.578 \\

        BB2 & \MODELSHORT & (DFT, 3) & 
        6.78 & 10.73 & 0.717 & 0.601 \\

        BB3 & \MODELSHORT & (DFT, 6) &
        7.07 & 11.37 & 0.709 & 0.587 \\

        \midrule
        BB4 & \EMODELSHORT~(SEP) & (DFT, 3) & 
        6.78 & 10.75 & 0.708 & 0.582 \\

        BB5 & \EMODELSHORT & (DFT, 3) & 
        6.62 & 10.57 & 0.725 & 0.604 \\
        
        BB6 & \EMODELSHORT & (DFT, 6) & 
        6.81 & 10.91 & 0.716 & 0.594 \\

        \midrule
        BB7 & \EMODELSHORT~(SEP) & (Haar, 4) & 
        6.99 & 11.35 & 0.699 & 0.563 \\
        
        BB8 & \EMODELSHORT & (Haar, 4) & 
        6.44 & 10.28 & 0.730 & 0.613 \\

        BB9 & \EMODELSHORT & (Haar, 6) & 
        6.90 & 10.82 & 0.713 & 0.599 \\

        \bottomrule
    \end{tabular}

\end{table}

\begin{table}[tbp]
    \centering
    \scriptsize
    \caption{
        Ablation studies on nuScenes (3D bounding box) with 2 seconds prediction horizon ($t_h = t_f = 4$).
        Models with ``(D)'' only predict one deterministic trajectory for each agent.
        Other models are validated with \emph{best-of-20}, and we set $N_{key} = 2$ for all these models.
    }
    \label{tab_ab_3dbb}
    \begin{tabular}{c|c|c|cccc}
        \toprule
        No. & Model & T & ADE $\downarrow$ & FDE $\downarrow$ & AIoU $\uparrow$ & FIoU $\uparrow$ \\

        \midrule
        3DBB1 & \MODELSHORT~(D) & DFT &
        0.534 & 0.930 & 0.597 & 0.439 \\

        3DBB2 & \EMODELSHORT~(D) & DFT &
        0.529 & 0.919 & 0.602 & 0.445 \\

        3DBB3 & \MODELSHORT~(D) & Haar &
        0.530 & 0.921 & 0.598 & 0.441 \\

        3DBB4 & \EMODELSHORT~(D) & Haar &
        0.525 & 0.918 & 0.608 & 0.450 \\
        
        \midrule
        3DBB5 & \MODELSHORT~(SEP) & Haar &
        0.210 & 0.306 & 0.761 & 0.683 \\

        3DBB6 & \MODELSHORT & Haar &
        0.213 & 0.310 & 0.762 & 0.684 \\

        3DBB7 & \MODELSHORT~(SEP) & DFT &
        0.208 & 0.302 & 0.763 & 0.686 \\
        
        3DBB8 & \MODELSHORT & DFT &
        0.207 & 0.299 & 0.764 & 0.687\\
        
        \midrule
        3DBB9 & \EMODELSHORT & None &
        0.213 & 0.308 & 0.760 & 0.681 \\

        3DBB10 & \EMODELSHORT~(SEP) & Haar &
        0.209 & 0.302 & 0.763 & 0.686 \\

        3DBB11 & \EMODELSHORT & Haar &
        0.209 & 0.301 & 0.764 & 0.688 \\

        3DBB12 & \EMODELSHORT~(SEP) & DFT &
        0.207 & 0.299 & 0.763 & 0.687 \\
        
        3DBB13 & \EMODELSHORT & DFT &
        0.203 & 0.291 & 0.766 & 0.691 \\

        \bottomrule
    \end{tabular}
\end{table}

\begin{table}[h]
    \centering
    \caption{
        Ablation studies (3D skeleton prediction) on Human3.6M.
        The metrics used are ``FDE@$x$ms'' (MPJPE) in millimeters.
    }
    \scriptsize
    \label{tab_ab_ske}
    \begin{tabular}{c|c|c|ccccc}
        \toprule
        No. & Model
        & (T, $N_{key}$) & 80 $\downarrow$ & 160 $\downarrow$ & 320 $\downarrow$ & 400 $\downarrow$ \\

        \midrule
        SKE1 & \MODELSHORT~(SEP) & (Haar, 4) & 6.02 & 16.73 & 46.58 & 63.29 \\
        SKE2 & \MODELSHORT & (Haar, 4) & 8.79 & 19.71 & 45.53 & 59.49 \\
        SKE3 & \EMODELSHORT & (Haar, 4) & 9.00 & 19.68 & 44.48 & 57.80 \\

        \midrule
        SKE4 & \MODELSHORT~(SEP) & (DFT, 4) & 7.62 & 19.62 & 51.47 & 68.92 \\
        SKE5 & \MODELSHORT & (DFT, 4) & 13.03 & 26.68 & 55.04 & 69.49 \\
        SKE6 & \EMODELSHORT & (DFT, 4) & 12.18 & 25.08 & 52.54 & 66.62 \\

        \midrule
        SKE7 & \EMODELSHORT & (None, 4) & 8.68 & 24.71 & 50.67 & 64.56 \\

        \midrule
        SKE8 & \EMODELSHORT & (Haar, 1) & 9.45 & 21.65 & 46.70 & 63.58 \\
        SKE9 & \EMODELSHORT & (Haar, 6) & 9.21 & 20.47 & 45.38 & 59.23 \\

        \bottomrule
    \end{tabular}
\end{table}

(c) \textbf{Comparisons of Fourier and Haar Transform across Trajectory Forms.}
We begin with three minimal ablation variations in \TABLE{tab_ab_ethucy}: CO1, CO2, and CO3, where all interaction-related networks have been removed, and only the Transformer backbone is used to forecast 2D deterministic trajectory directly rather than hierarchically.
Compared to CO1, the average performance of DFT variation CO2 has been improved by over 12.1\%.
Furthermore, Haar variation CO3 also achieves a comparable performance with CO2, which both greatly outperform the non-transform CO1, indicating the usefulness of these transforms in 2D cases.

Haar transform could represent the time-frequency joint character of the input sequence while Fourier transform could not.
It means that Haar variations may be suitable for modeling with complex time-frequency changes.
To verify our thought, we visualize the quantitative metrics of transform-related ablation variations across different forms of trajectories in \FIG{fig_exp_transform}.
Detailed metrics are reported in \TABLE{tab_ab_ethucy,tab_ab_sdd,tab_ab_3dbb,tab_ab_ske}.
We can see from \FIG{fig_exp_transform} that \EMODEL-DFT (CO11) and \EMODEL-Haar (CO13) perform almost the same on 2D coordinate datasets.
Specifically, the Haar variation CO19 performs even worse than the DFT variation CO18 when forecasting 2D coordinates on SDD.
The performance differences between these transforms have gradually increased as the dimensionality of the predicted trajectory increases.
Haar variations \{BB8, SKE3\} outperform DFT variations \{BB5, SKE6\} on the more challenging 2D bounding box and 3D skeleton datasets for up to 2.8\% ADE and 13.2\% MPJPE accordingly.
On the contrary, DFT variation SKE6 even performs worse than the none-transform SKE7 for about 3.2\% quantitatively at 400ms when forecasting 3D skeletons.
These results show that the Haar variation performs better at predicting datasets (scenarios) that contain more complex time-frequency changes, like the more complex 2D bounding boxes and 3D skeletons.

Notably, Haar variations do not show huge performance gains to DFT ones for 3D bounding box cases.
When generating $k=20$ trajectories, Haar variation 3DBB11 loses about 1.4\% ADE and 2.3\% FDE relative to the DFT variation 3DBB13.
We infer it is caused by the short observation period $t_h = 4$ in this case, which makes trajectories more ``smoother''.
In other words, in the relatively short observation period, the possibility of large changes in the trajectory will be significantly reduced, considering these agents' physical movement constraints.
The Haar transform has a \emph{vanishing moment} of 1, which means that the energy is distributed more in the high-frequency portions in the Haar spectrum, making it difficult to focus on the lower-frequency portions, leading to a poor representation of the smoother trajectories\footnote{
    See more discussions and analyses in the supplemental material.
}.
Nevertheless, both transforms still facilitate the prediction.
Compared to the none-transform 3DBB9, transformed variations \{3DBB11, 3DBB13\} obtain at least 1.9\% better ADE and 2.3\% better FDE.

In conclusion, DFT is more effective for simpler or smoother trajectories, but Haar transform is more effective for trajectories with higher dimensionality or fast time-frequency changing properties.
These experiments also demonstrate the different advantages of these transforms, thus supporting the choice for different scenarios.

~\\\textbf{The Existence of Dimension-wise Interactions and Verifications of Bilinear Structures across Trajectory Forms.}
We conduct the following discussions to analyze the existence of dimension-wise interactions and how the proposed bilinear structure helps model them across different trajectory forms.

(a) \textbf{2D Coordinate Cases.}
We start by comparing variations CO9 and CO8 in \TABLE{tab_ab_ethucy}.
The ``SEP'' variation CO8 considers nothing about interactions between two dimensions $x_t$ and $y_t$ in a 2D coordinate.
Especially in the zara1 dataset, CO8 performs worse than the other one for about 10.5\% ADE and 30.0\% FDE, showing a fairly non-negligible performance drop.
Results on SDD (in \TABLE{tab_ab_sdd}) also show similar trends, including only about 1\% ADE drop and a 6.1\% huge FDE drop between variations CO15 and CO16, which demonstrates the existence of dimension-wise interactions within 2D coordinates when forecasting trajectories.
Comparing variations CO11 and CO9 (\TABLE{tab_ab_ethucy}), the former (with bilinear structure) obtains 1.13\% better ADE on ETH-UCY.
Variation CO18 also outperforms variation CO16 (\TABLE{tab_ab_sdd}) with a significant improvement of 7.7\% ADE and 7.8\% FDE, which proves the usefulness of the bilinear structure in modeling dimension-wise interactions in coordinate cases.

(b) \textbf{2D Bounding Box Cases.}
Similar to the above SEP variations, in \TABLE{tab_ab_sdd}, SEP variations \{BB1, BB4, BB7\} have been constructed to forecast 4-dimensional 2D bounding boxes by forecasting two 2D corner points separately, which means that interactions between these two 2D points have not been considered.
Variation BB1 obtains 4.12\% and 1.68\% worse ADE and FDE than BB2, similarly 2.41\%/1.70\% worse between \{BB4, BB5\} and 8.54\%/10.40\% worse between \{BB7, BB8\}.
Regardless of transforms, these SEP variations show a significant performance drop, which illustrates the existence of dimension-wise interaction between two points in 2D bounding box cases, not to mention the above-validated interactions within each 2D point.
In addition, compared with BB2, DFT variation BB5 brings up to 2.4\% ADE and 1.5\% FDE improvements.
Haar variation BB8 also shows similar improvements, verifying the usefulness of the bilinear structure in 2D bounding box cases.

(c) \textbf{3D Bounding Box Cases.}
We can see from \TABLE{tab_ab_3dbb} that SEP variations \{3DBB7, 3DBB10, 3DBB12\} have lead to uo 1.5\% performance drops compared to their ``interactive'' variations \{3DBB8, 3DBB11, 3DBB13\}.
Thus, the dimension-wise interactions in 3D bounding box prediction cases can be verified through these DFT variations (we will discuss variation 3DBB5 later).
Furthermore, when generating $k=20$ trajectories, variation 3DBB13 shows better performance than variation 3DBB8, including improvements of 8.3\% in ADE and 10.4\% in FDE.
Haar variation 3DBB11 also outperforms 3DBB6 for about 1.9\% ADE and 2.6\% FDE.
From these results, the usefulness of the bilinear structure has been validated when forecasting 3D bounding boxes.

(d) \textbf{3D Skeleton Cases.}
Intuitively, the connections between different joints in the skeleton naturally become part of the dimension-wise interactions.
As shown in \TABLE{tab_ab_ske}, comparing Haar variations SKE1 and SKE2, separately forecasting each 3D point may lead to a 6.00\% performance drop (at 400ms).
The results also indicate another interesting trend, \IE, the SEP variations perform better than others over a fairly short period.
Also, the DFT variation SKE5 performs worse compared to the SEP variation SKE4.
We will discuss them later.
After adding the bilinear structure, both DFT and Haar variations (SKE6 and SKE3) are greatly improved, including 3.33\% and 8.67\% at 400ms.
Considering that skeletons have higher dimensionality ($M=51$), these results further demonstrate the existence of dimension-wise interactions in heterogeneous trajectories and verify the effectiveness of bilinear structures for modeling them.

(e) \textbf{Discussions of Modeling Capacity and Generalizability.}
We can see that there is a clear difference in the modeling capability of bilinear structures for modeling dimension-wise interactions between different trajectory forms and transforms.
For example, the SEP Haar variation 3DBB5 performs better than 3DBB6 in \TABLE{tab_ab_3dbb}, while the DFT variation SKE5 performs worse compared to the SEP variation SKE4 in \TABLE{tab_ab_ske}.
In the above analyses, we conclude that the Fourier transform is better for smoother trajectories and vice versa for the Haar transform.
Thus, we can infer that different ways to characterize each trajectory-dimension could also directly affect or limit the generalizability.

In addition, we mentioned that the SEP variations perform better than others over a fairly short period, such as the metrics under 80ms and 160ms in \TABLE{tab_ab_ske}.
Results on 2D coordinate cases in \TABLE{tab_ab_ethucy} also indicate a similar phenomenon that the impact of dimension-wise interactions on trajectories further into the future is even more critical since the changes in FDE are larger than ADE.
It can be seen from these results that there might be some ``response time'' for the dimension-wise interaction to make an effort to determine future trajectories.
In other words, dimension-wise interactions may become more critical when predicting longer-term trajectories\footnote{
    More validations are included in the supplemental material.
}.
In summary, the bilinear structure will be more effective in modeling dimensional-wise interactions in heterogeneous trajectories with more complex time-frequency properties and longer prediction terms.

~\\\textbf{Number of Spectrum Keypoints ($N_{key}$).}
This part analyzes the quantitative performance changes brought by the number of keypoints.
\TABLE{tab_ab_ethucy} reports the results of DFT variations (CO5, CO9, CO6, and CO7) under different $N_{key}$ configurations (2D coordinate).
Their results indicate that variation CO9 ($N_{key} = 3$) outperforms CO5 ($N_{key} = 1$) with 8.3\% better ADE and 6.0\% better FDE.
The performance of CO7 ($N_{key} = 6$) reduces by 5.7\% and 1.8\% compared to the base CO4.
In \TABLE{tab_ab_sdd}, 2D bounding boxes results of DFT variations (BB5 and BB6) and Haar variations (BB8 and BB9) demonstrate that the prediction network may perform worse with either a lower or higher number of keypoints, where ADE and FDE of variation BB6 are 2.9\% and 3.2\% worse than BB5, and about 7.1\% and 5.3\% ADE and FDE drops for variation BB9 compared to BB8.
The same phenomenon also occurs when forecasting 3D skeletons.
In \TABLE{tab_ab_ske}, $N_{key} = 6$ SKE9 and $N_{key} = 1$ SKE8 perform worse than SKE6 ($N_{key} = 4$) with 10.0\% and 2.5\% in FDE@400ms, respectively.
As a result, too-low or too-high $N_{key}$s may lead to metric drops, consistent with all these trajectory forms.
Similar to the quantitative analysis in \MODELCITE, we can describe these phenomena as a smaller $N_{key}$ might cause a looser planning division that makes it challenging to reflect the differences between agents' similar future choices, while a larger one may lead to a strict trends division, which could be difficult for the subsequent network to reflect agents' multiple uncertain future choices and motion preferences.


\subsection{Qualitative Analysis}
\label{sec_qual}

\begin{figure}[tb]
    \centering
    \includegraphics[width=1.0\linewidth]{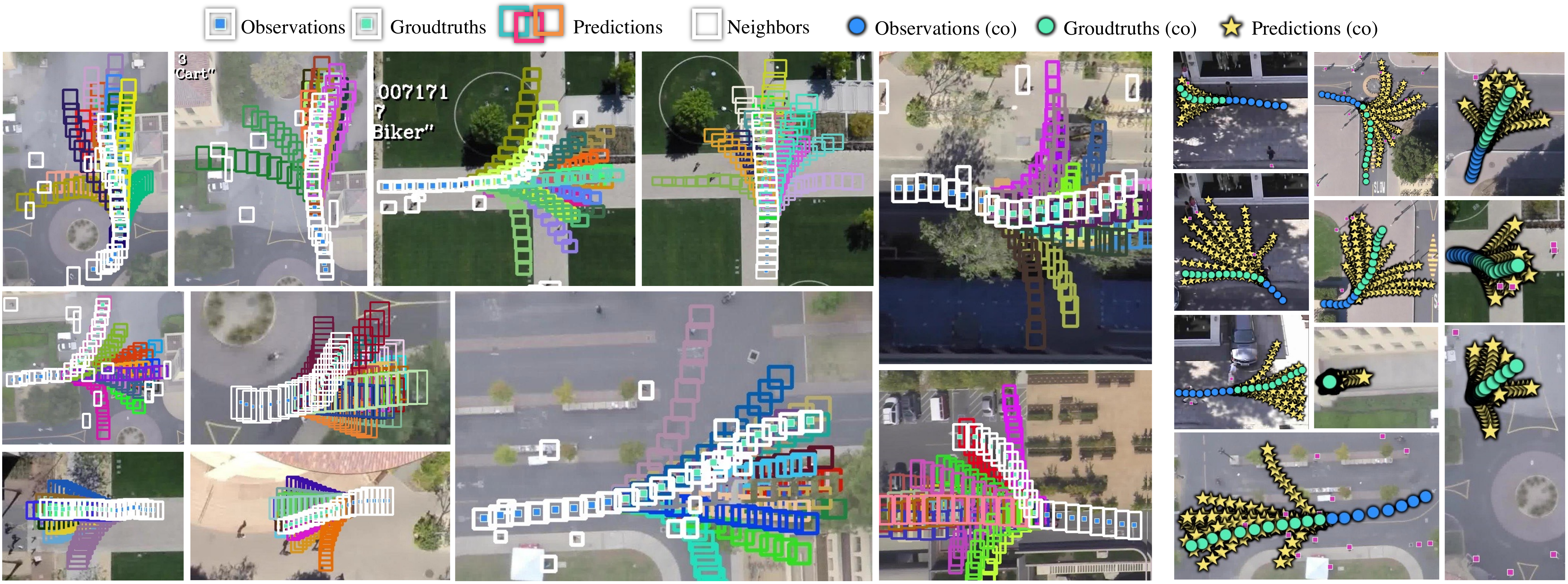}
    \caption{
        Visualized predictions given by \EMODEL-Haar with 2D bounding boxes and 2D coordinates.
        Each image contains 20 random predictions.
    }
    \label{fig_vis}
\end{figure}

\begin{figure}[tb]
    \centering
    \includegraphics[width=0.8\linewidth]{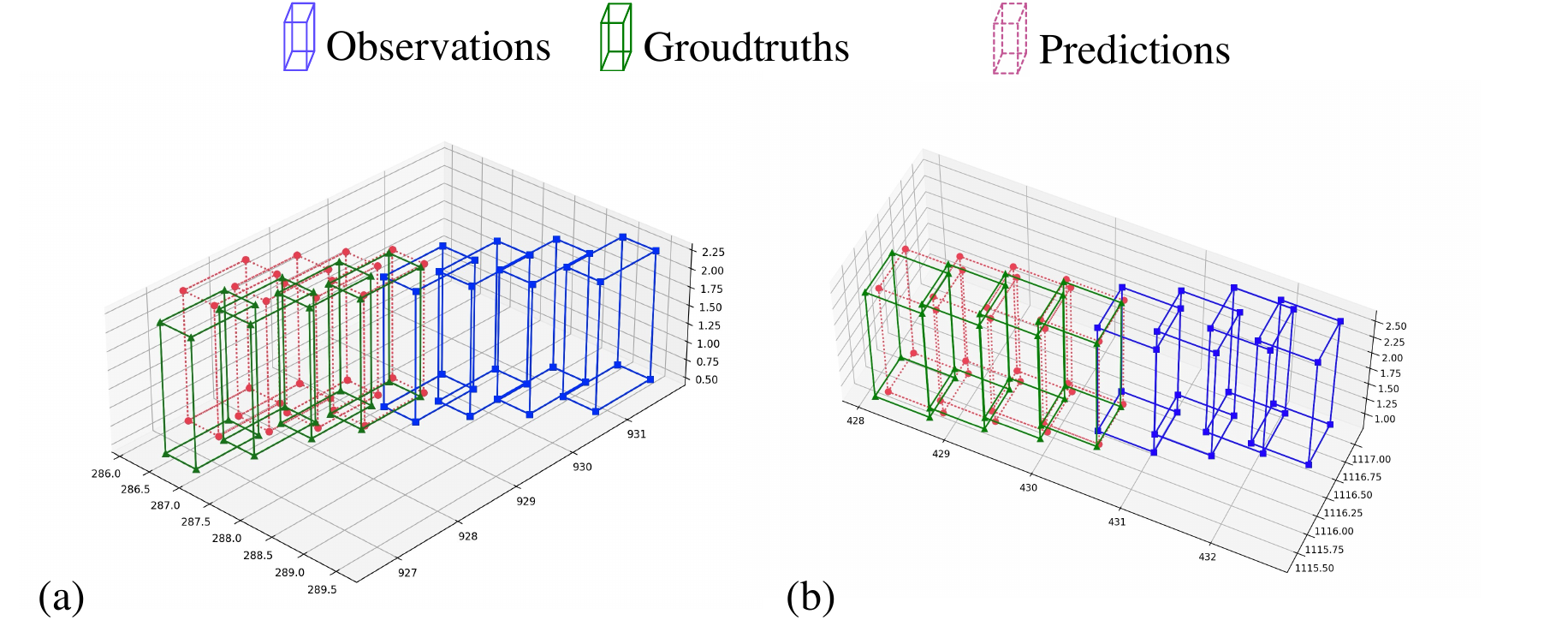}
    \caption{
        Visualized \EMODEL-Haar predictions (3D bounding boxes) on nuScenes.
        For a clear description, only 1 sampled prediction is drawn.
    }
    \label{fig_vis_3D}
\end{figure}

\begin{figure}[tbh]
    \centering
    \includegraphics[width=1.0\linewidth]{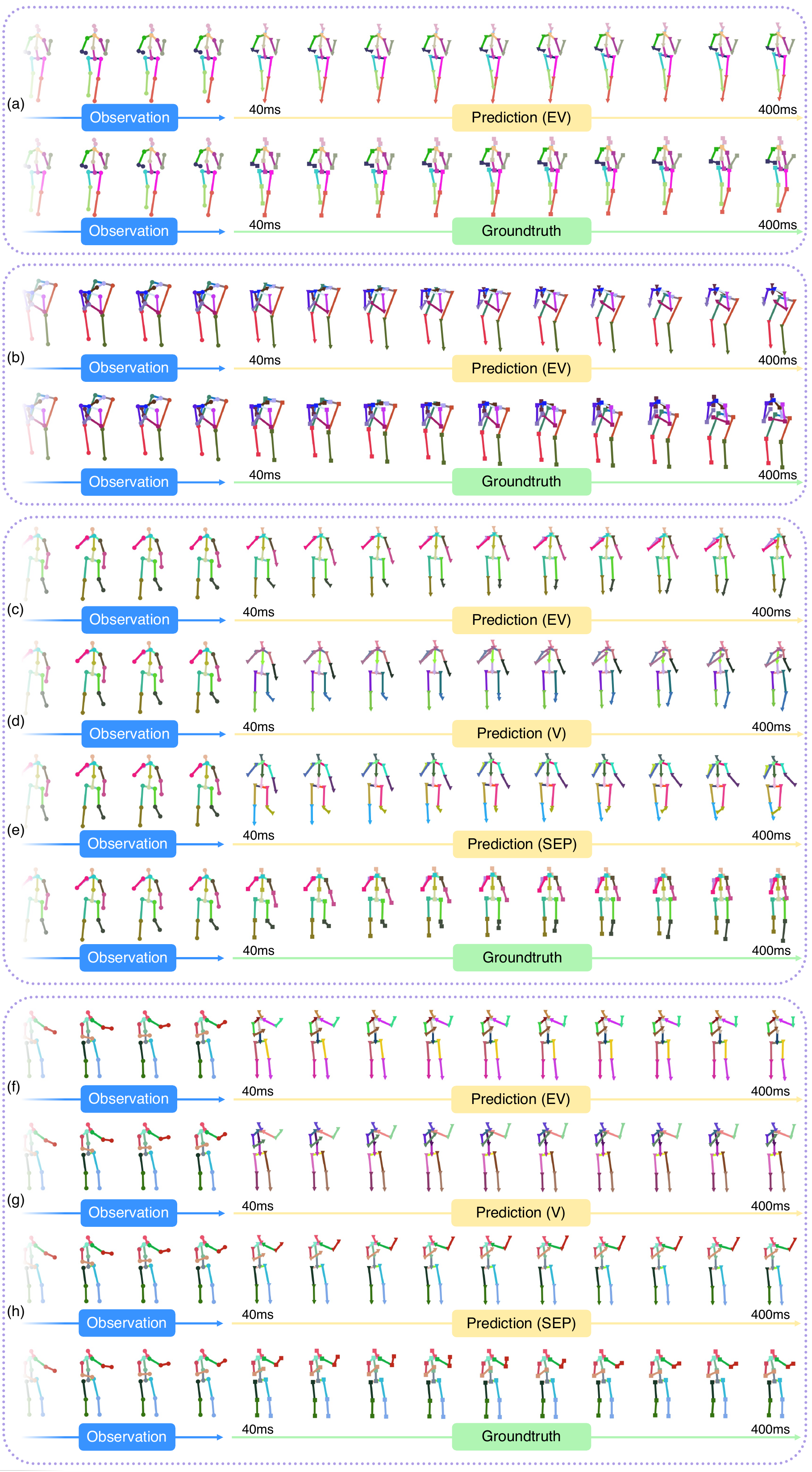}
    \caption{
        Visualized 3D skeleton predictions provided by \EMODEL-Haar, \MODEL-Haar, and \MODEL-Haar (SEP) on Human3.6M dataset.
    }
    \label{fig_skeleton}
\end{figure}

\begin{figure*}[tb]
    \centering
    \includegraphics[width=1.0\linewidth]{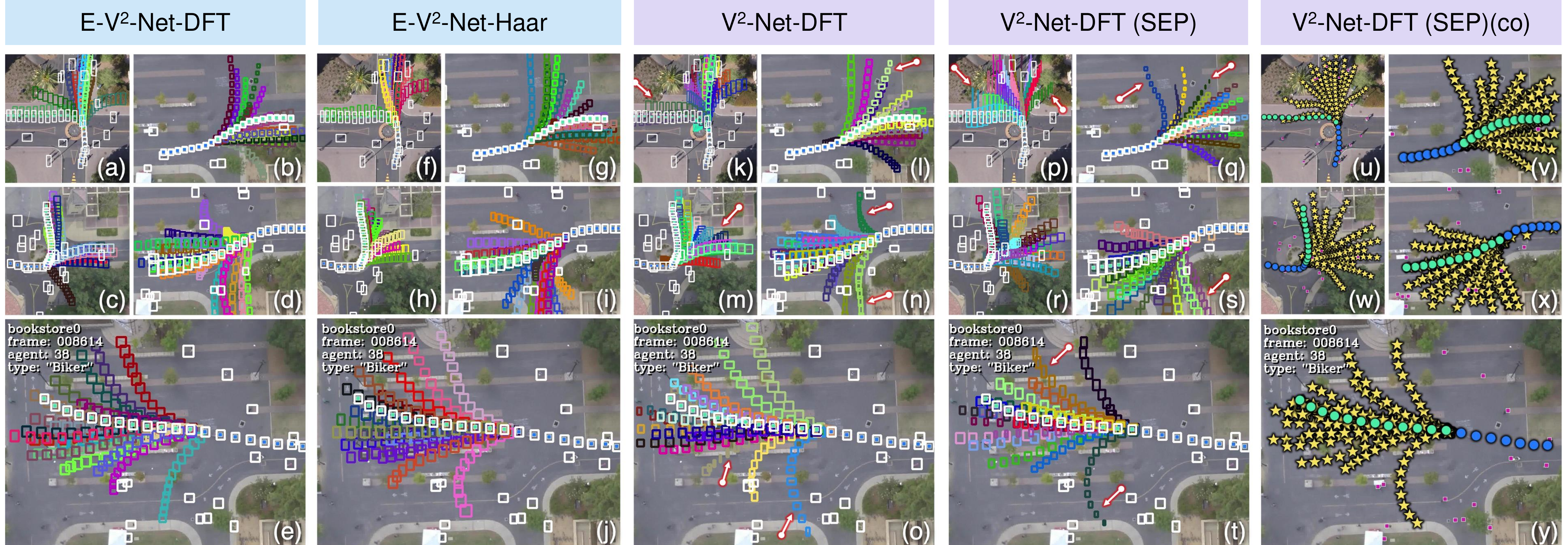}
    \caption{
        Visualized predictions provided by different \MODEL~or \EMODEL~variations on SDD (2D coordinates or bounding boxes).
    }
    \label{fig_exp_bs}
\end{figure*}

\begin{figure}[tb]
    \centering
    \includegraphics[width=1.0\linewidth]{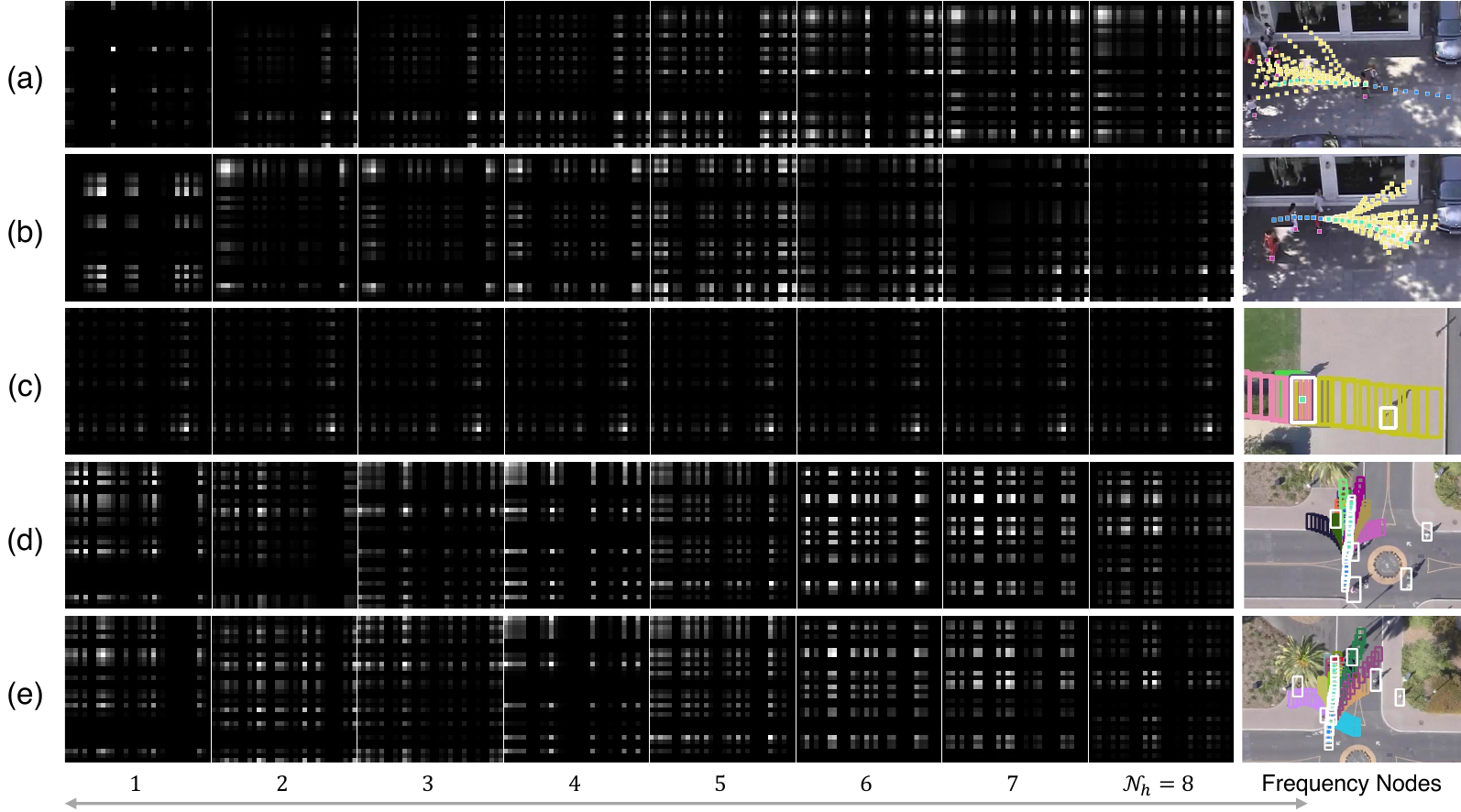}
    \caption{
        Visualized outer product matrices $\mathbf{R} = \mathbf{f}_\alpha \otimes \mathbf{f}_\alpha \in \mathbb{R}^{8 \times 64 \times 64}$ in different prediction cases.
        The horizontal axis (1 to $\mathcal{N}_h=8$) represents different frequency components.
        Bright parts indicate higher values.
    }
    \label{fig_exp_bsfeatures}
\end{figure}

\begin{figure*}[tb]
    \centering
    \includegraphics[width=1.0\linewidth]{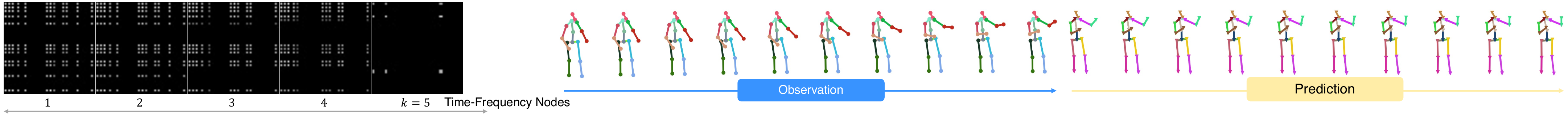}
    \caption{
        Visualized outer product matrices $\mathbf{R} \in \mathbb{R}^{5 \times 256 \times 256}$ with the corresponding visualized prediction given by \EMODEL-Haar.
    }
    \label{fig_exp_bsfeatures_ske}
\end{figure*}

\noindent\textbf{Visualization.}
As shown in \FIG{fig_vis} and \FIG{fig_vis_3D}, \EMODEL-Haar could predict heterogeneous trajectories considering multiple interactions.
For example, \EMODEL~gives a variety of different future options for bikers, carts, and cars crossing the intersection.
Besides, when agents bypass the grass, it considers the physical constraints of the environment to provide acceptable predictions.
Additionally, it also shows strong adaptability in some unique prediction scenarios.
For instance, it gives scene-friendly predictions to the biker to pass through the traffic circle: turning right or going ahead to leave the circle, and turning left to keep driving in the circle.
Notably, when predicting 2D and 3D bounding boxes, the motion trends and the box structures are considered together.
In short, \EMODEL s could provide heterogeneous predicted trajectories adapted to different scenarios.

We also visualize skeleton predictions in \FIG{fig_skeleton}.
Unlike coordinate or bounding box prediction, forecasting the skeleton is a different challenge since human actions are collaborative efforts of movements in every joint.
Interestingly, \FIG{fig_skeleton} (a) shows that the slight bend in the knee can be accurately predicted by \MODEL-Haar.
Is is also amzaing that \EMODEL-Haar can predict the postures of different joints while satisfying the physical constraints of human bodies in \FIG{fig_skeleton} (b), when predicting the action ``Sitting Down''.
The detailed body movements, such as trunk rotations, bendings and raisings of arms, and the lifting of the back, are all predicted vividly.
Notably, it does not consider any additional motion rules, which differs from most motion prediction approaches.
Therefore, these qualitative results demonstrate the superiority and competitiveness of \EMODEL~in higher-dimensional heterogeneous trajectory prediction.

~\\\textbf{Discussions of Dimension-wise Interactions in Visualized Predictions.}
Next, we further qualitatively validate the existence of dimension-wise interactions and discuss how \EMODEL~maintain these interactions when forecasting.

(a) \textbf{Separate (SEP) Variations.}
2D bounding box predictions provided by the separate variation \MODEL-DFT~(SEP) are shown in \FIG{fig_exp_bs} (p) to (t).
Predictions in subfigures (p), (q), (s), and (t) indicate that the SEP variation completely fails to establish and reflect potential dimension-wise interactive factors, such as the preservation and change of the shape of these boxes.
For example, whether for the bottom dark green prediction indicated by the arrow in \FIG{fig_exp_bs} (t), or the dark green prediction indicated in (q), the SEP variation does not hold their shapes at all, and the forecasted destination box could even almost turns into a dot.
We also visualize 3D human skeleton predictions provided by the \MODEL~(SEP) in \FIG{fig_skeleton} (e) and (h).
We can see that the predicted skeletons are almost movements of each joint individually, without being able to take into account the relationships between the joints.
As a result, the length of the predicted forearms or legs seems anomalous, and their synergistic movements appear uncoordinated.

(b) \textbf{Non-Bilinear-Structure Variations.}
Compared with the above SEP variation, \MODEL-DFT takes the 4-dimensional bounding box as the input, which means that interactions among dimensions could be partially established due to the trajectory embedding operation.
Its 2D bounding box predictions in \FIG{fig_exp_bs} (k) to (o) exhibit better shape preservation properties than the above SEP variation.
For example, compared with the above-discussed dark green predictions in \FIG{fig_exp_bs} (q) and (t), the light green prediction and the blue prediction focused in subfigures (l) and (o) show the suppression of the unnatural changes of the bounding box to a certain extent.
Nevertheless, the dark green prediction focused in subfigure (k) even gets bigger as the prediction time grows, which indicates that it still fails to model the detailed dimension-wise interactions.
As for the skeleton variation, predictions in \FIG{fig_skeleton} (d) also show that leg movements, especially the left leg, are more natural and acceptable compared to the SEP ones in \FIG{fig_skeleton} (e).

(c) \textbf{Bilinear Structure Variations.}
As shown in the first two columns of \FIG{fig_exp_bs}, \EMODEL~variations add additional bilinear structures to focus on the dimension-wise interaction.
Compared with the non-bilinear variations, two \EMODEL~variations present better predictions for 2D bounding boxes qualitatively.
Most importantly, the previously discussed examples of hard-to-maintain shapes are largely resolved.
Besides, Haar variations also better predict the motion tendency of bounding boxes.
For example, in the bottom pink prediction in \FIG{fig_exp_bs} (j), these predicted boxes become more slender when the biker turns and wider after the turn, thus conforming to the real motion pattern of the biker in the drone view, which is not reflected in the other variations, thus also indicating the qualitative improvement brought by Haar transform.
Additionally, \EMODEL~also shows better qualitative performance when forecasting skeletons compared to the non-bilinear ones.
Compared with \FIG{fig_skeleton} (g), predictions in (f) present better joint collaborations, like moving the right arm slightly up and down in front of the chest and bending the knee.
It shows the competitiveness of the bilinear structure in modeling dimension-wise interactions when forecasting high-dimensional heterogeneous trajectories.

~\\\textbf{Analyses of Outer Product Matrices.}
Next, we visualize several outer product matrices $\mathbf{R}$ to analyze how bilinear structures work in modeling dimension-wise interactions.

(a) \textbf{2D Coordinate Cases.}
As shown in \FIG{fig_exp_bsfeatures} (a) and (b), there are multiple separate bright parts in the outer product matrix.
We call these bright parts the ``high-value clusters'' for convenient analyses.
It can be seen that the network could capture potential interactions in the high-dimensional $\mathbf{f}_\alpha$, which are represented by high-value clusters in these matrices, especially these non-diagonal elements, although dimension-wise interactions are relatively easy in 2D coordinate cases compared to other heterogeneous trajectories.

(b) \textbf{2D Bounding Box Cases.}
As shown in \FIG{fig_exp_bsfeatures} (d) and (e), there are more high-value clusters in the outer product matrices which indicates there may exist more other potential dimension-wise interactions in 2D bounding boxes ($M = 4$).
Intuitively, the dimension-wise interactions in 2D bounding boxes include interactions between two corner points (top-left and bottom-right points) so that the shape of the bounding box could be maintained, leading to more high-value clusters in outer product matrices.
It also indicates that \EMODEL~could focus on the hard-to-capture slight temporal motion changes.
In \FIG{fig_exp_bsfeatures} (d) and (e), two observed bounding box trajectories have similar shapes, but their matrices $\mathbf{R}$ are obversely different in the last frequency portion.
As a result, \EMODEL~forecast these samples in completely different trends and changing properties.

(c) \textbf{3D Skeleton Cases.}
It can be seen that the observed skeletons in \FIG{fig_exp_bsfeatures_ske} basically remain static in the first four observation frames, and after that, the hands start to rise till the end.
Correspondingly, as shown in \FIG{fig_exp_bsfeatures_ske}, the outer product matrices in $k = 1$ and $k = 2$ seem the same, while some bright parts in $k = 3$ and $k = 4$ have become darker.
Interestingly, the outer product matrix in $k = 5$ is completely different from the others, which means that dimension-wise interactions at this time-frequency portion (corresponding to the last two observation frames) present different frequency properties.
This phenomenon could reflect the bilinear structure's better time-frequency localization characteristics (like the subtle and rapid changes).

~\\\textbf{Discussions of Qualitative Generalizability.}
\EMODEL~is designed to handle different trajectory forms.
However, the forecasted trajectories and their outer product matrices present different styles or distributions for different forms of trajectories.
We first take \MODEL~(SEP) as an example.
Its coordinate predictions (in \FIG{fig_exp_bs} (u) - (y)) are about acceptable, while the 2D bounding box predictions in (p) - (t) have become weird, with some of the boxes' abnormal shapes.
Its skeleton predictions could even be described as ``ridiculous'' in \FIG{fig_skeleton} (e) for it forecasts trajectories that were completely inconsistent with physical movements.
We can conclude that the dimension-wise interactions may become more ``influential'' as the dimensionality of trajectories increases.
By adding the bilinear structure with different transforms, predictions provided by the enhanced \EMODEL~models have been greatly improved in these corresponding \FIG{fig_exp_bs} (a) to (j) or \FIG{fig_skeleton} (c) and (f) cases.
However, the degrees of these improvements are also clearly limited to the dimensionalities of trajectories.
For example, the qualitative enhancements in 2D bounding box predictions are more straightforward, while there is still a more significant shortage of skeleton predictions than their groundtruths.

It can be seen that the number of high-value clusters in outer product matrices also differs with trajectory forms (dimensionalities).
The more clusters, the more complex dimension-wise interactions would be.
For example, scenario (c) is easier to handle since the agent almost stands still during observation.
Then, \FIG{fig_exp_bsfeatures} (c) have fewer high-value clusters (about only five clusters) than (a) and (b) (up to 49 clusters).
Instead, \FIG{fig_exp_bsfeatures} (d) and (e) have more clusters (up to 81), demonstrating the more complex dimension-wise interactions within bounding boxes.
The same phenomenon also occurs in 3D human skeleton prediction, and there are a considerable amount of high-value clusters (about 144 clusters) in \FIG{fig_exp_bsfeatures_ske}, which means that dimension-wise interactions have become ``more and more'' complex.

From the above analyses, although the prediction performance improvement is relatively limited in higher dimensional skeletons, it still illustrates its effectiveness in predicting heterogeneous trajectories, validating its quantitative and qualitative generalization performance.


\subsection{Parameter Amount and Inference Speed}

\begin{table}[tp]
    \centering
    \scriptsize
    \caption{
        Comparisons of inference times and parameter amount of 2D coordinate prediction on ETH-UCY ($t_h = 8, t_f = 12$).
        Results are obtained from \cite{li2021spatial} on one NVIDIA GeForce GTX 1080Ti card. 
    }
    \begin{tabular}{c|c|c|ccccccccc}
        \toprule
        Models & Time/Para. &
        Models & Time/Para. \\

        \midrule
        \SOCIALLSTM\SOCIALLSTMCITE & 1180 ms/264K & \SOCIALGAN\SOCIALGANCITE & 97 ms/46.3K \\
        \SRLSTM\SRLSTMCITE & 1179 ms/64.9K & \DAGNET\DAGNETCITE & 46 ms/2.35M \\
        \PECNET\PECNETCITE & 607 ms/2.10M & \STGCNN\STGCNNCITE & 2.0 ms/7.6K \\
        \PEEKING\PEEKINGCITE & 114 ms/360.3K & \STCNET\STCNETCITE & 1.3 ms/0.7K \\
        
        \midrule
        \EMODELSHORT-DFT & 43 ms/3.9M & \EMODELSHORT-DFT-L & 21 ms/1.9M \\
        \EMODELSHORT-Haar & 40 ms/3.9M & \EMODELSHORT-Haar-L & 19 ms/1.9M \\

        \bottomrule
    \end{tabular}
    \label{tab_parameter}
\end{table}


\begin{table}[tp]
    \centering
    \scriptsize
    \caption{
        Inference times $\tau$ and memory usages $\mu$ of \MODEL~and \EMODEL~(\LINEARNOTE) on one Apple M1 Mac mini.
        Results (as ``$\tau$ ms/$\mu$ MB'') are obtained by predicting $k=20$ trajectories under $t_h = 8,~t_f = 12$.
    }
    \begin{tabular}{c|c|cccc}
        \toprule
        No. & Model, T & co & bb & 3Dbb & ske \\

        \midrule
        INF1 & \MODELSHORT, DFT & 30 / 13.1 & 30 / 13 & 37 / 13 & 112 / 83.1 \\
        INF2 & \MODELSHORT, Haar & 26 / 13.6 & 27 / 11.3 & 27 / 11.3 & 28 / 81.9 \\
        INF3 & \EMODELSHORT, DFT & 31 / 25.9 & 34 / 25.7 & 37 / 25.8 & 112 / 147.9 \\
        INF4 & \EMODELSHORT, Haar & 27 / 20.1 & 27 / 17.8 & 28 / 17.8 & 29 / 124.1 \\

        \bottomrule
    \end{tabular}


    \label{tab_speed_ablation}
\end{table}

\textbf{Comparisons with Other Baselines.}
We report the inference speed and the number of parameters of different models in \TABLE{tab_parameter}.
All results are measured on one NVIDIA GeForce GTX 1080Ti GPU.
Models with postfix ``-L'' indicate that linear interpolation models have replaced their second-stage spectrum interpolation networks to speed up calculation (variations CO12 and CO14 in \TABLE{tab_ab_ethucy}).
Many fast calculation methods like \STCNET\STCNETCITE~have been proposed recently.
Although the time efficiency of the proposed methods is not better than theirs, we have greatly improved the quantitative prediction performance as a balance.

\textbf{Efficiency Discussions.}
In \TABLE{tab_speed_ablation}, we report the inference time under different forms of predicted trajectories on one Mac Mini with an Apple M1 chip (8GB memory) (which could be easily accessed on mobile devices such as iPads).
It can be seen that despite the higher amount of trainable parameters, models still have faster inference speeds.
Especially for Haar variations INF2 and INF4, they have almost about 400\% faster compared to DFT variations when predicting 3D skeletons\footnote{
    See more ablation efficiency discussions in supplemental material.
}.
We also report memory usages during test.
In \TABLE{tab_speed_ablation}, \EMODEL~variations allocates more memory when running tests.
Interestingly, Haar variations still exhibit better memory use properties, reducing memory by about 15\% compared to DFT variations, which also illustrates another significant advantage of the Haar variation in dealing with complex heterogeneous trajectories, demonstrating their better real-time prediction capability.

\section{Conclusion and Limitations}

This work aims to forecast heterogeneous trajectories via spectrums hierarchically.
Thus, trajectories to be analyzed and forecasted in this manuscript could have multiple forms like bounding boxes or skeletons.
Also, the dimensionality of trajectories could become larger.
Naturally, the interactions between the different dimensions of the trajectory will become more complex and more non-negligible.
One of our main focuses has moved on to the potential interactions among different trajectory-dimensions, which we name \textbf{dimension-wise interactions}.
Heterogeneous trajectories also mean that each trajectory-dimension may have more complex time-frequency properties.
Thus, the other focus lies on how to model these interactions by simultaneously considering the time-frequency response of each trajectory-dimension when forecasting.

This work further extends \MODEL\MODELCITE~in two aspects.
On the one hand, we enhance the structure and input-output settings of the original network to be applied to trajectories with any dimensionalities, and employ an additional bilinear structure to model and fuse factors ``frequency response'' and ``dimension-wise interaction'' simultaneously.
On the other hand, more transforms, such as the Haar wavelet, are introduced to cope with different trajectory forms.
Multiple experiments show that \EMODEL~outperforms most state-of-the-art methods quantitatively on the ETH-UCY benchmark, Stanford Drone Dataset, nuScenes, and Human3.6M when forecasting different forms of heterogeneous trajectories, including 2D coordinates, 2D and 3D bounding boxes, and 3D human skeletons.
Ablation discussions have also verified the usefulness of each component.

Despite the impressive performance of \MODEL~and~\EMODEL, there are still potentials for further improvement.
For example, the Haar transform is not good at handling simpler or smoother trajectories despite its better time-frequency localization characteristics.
Moreover, \MODEL~and \EMODEL~do not consider any additional priors, leading to some performance drop when encountering complex trajectory forms like human skeletons.
Therefore, the above limitations will become the focus of our future research.


%





\ifCLASSOPTIONcaptionsoff
    \newpage
\fi



\bibliographystyle{IEEEtran}
\bibliography{ref.bib}
%



%

\begin{IEEEbiographynophoto}{Beihao Xia}
    received his Ph.D. degree in Huazhong University of Science and Technology, Wuhan, China, in 2023.
    His research interests include trajectory prediction, behavior analysis, and understanding.
\end{IEEEbiographynophoto}

\begin{IEEEbiographynophoto}{Conghao Wong}
    received the master's degree from Huazhong University of Science and Technology, Wuhan, in 2022, where he is currently pursuing the Ph.D. degree.
    His research interests include computer vision and pattern recognition.
\end{IEEEbiographynophoto}

\begin{IEEEbiographynophoto}{Duanquan Xu}
    is currently an Associate Professor in Huazhong University of Science and Technology, Wuhan, China.
    He received his Ph.D. degree from Huazhong University of Science and Technology in 2008.
    His research interests include image processing, and computer vision.
\end{IEEEbiographynophoto}

\begin{IEEEbiographynophoto}{Qinmu Peng}
    is currently an Associate Professor in Huazhong University of Science and Technology, Wuhan, China.
    He received his Ph.D. degree from the Department of Computer Science at Hong Kong Baptist University in 2015.
    His research interests include medical image processing, pattern recognition, machine learning, and computer vision.
\end{IEEEbiographynophoto}

\begin{IEEEbiographynophoto}{Xinge You}
    (Senior Member, IEEE) is currently a Professor in Huazhong University of Science and Technology, Wuhan. 
    He received his Ph.D. degree from the Department of Computer Science, Hong Kong Baptist University in 2004. 
    His research have expounded in 200+ publications, such as IEEE T-PAMI, T-IP, T-NNLS, T-CYB, CVPR, ECCV, ICCV. 
    He served/serves as an Associate Editor of the \textit{IEEE Transactions on Cybernetics}, \textit{IEEE Transactions on Systems, Man, Cybernetics: Systems}. 
    His research interests include image processing, wavelet analysis, pattern recognition, machine learning, and computer vision. 
\end{IEEEbiographynophoto}

\vfill








\appendices

\section{Additional Discussions on the Haar Transform}
\label{appendix_transform}

\textbf{The Advantages of Haar Transform in Time-Frequency Analyses.}
According to Fourier expansion, a signal can be expressed as the sum of a series of sine and cosine waves.
The Fourier transform of a function $x(t)$ can be represented as
\begin{equation}
    \label{eq_fourier}
    X(k) = \int_{-\infty}^{\infty} x(t) e^{-j2\pi kt} \mathrm{d} t.
\end{equation}
However, a significant shortage of the Fourier spectrum $X(k)$ is that the time variable $t$ is not directly or indirectly included, which means that a sequence $x(t)$ only has one specific Fourier spectrum $X(k)$, no matter how the time variable $t$ changes.
In other words, it has no \emph{time resolution}, so we can not infer which frequency portions present a specific time.

As the simplest variation of wavelets, the Haar transform makes it easy to analyze the time-frequency joint representation of signals.
The main difference between wavelet transforms and the Fourier transform is that wavelets are localized in both time and frequency, whereas the standard Fourier transform is only localized in frequency.
Formally, the continuous wavelet transform of a function $x(t)$ at scale $a > 0$ and translational value $b \in \mathbb{R}$ can be expressed as
\begin{equation}
    \label{eq_wavelet}
    X_w (a, b) = \frac{1}{\sqrt{a}}
        \int_{-\infty}^{\infty}
        x(t)
        \bar{\psi} \left(\frac{t-b}{a}\right) \mathrm{d}t,
\end{equation}
where $\psi (t)$ is the continuous mother wavelet, and $\bar{\psi}(t)$ represents its complex conjugate.
In Haar wavelet, we have
\begin{equation}
    \label{eq_mother_haar}
    \psi_{\mathrm{Haar}} (t) = \left\{\begin{aligned}
        1, &\quad 0 \leq t < \frac{1}{2}, \\
        -1, &\quad \frac{1}{2} \leq t < 1, \\
        0, &\quad \mathrm{Otherwise}.
    \end{aligned}\right.
\end{equation}

(a) \textbf{Time-Frequency Joint Representation.}
Compared to the Fourier transform in \EQUA{eq_fourier}, the Haar spectrum has two variables, scale $a$ and translation $b$.
According to these variables, the mother wavelet $\psi(t)$ could generate a set of basis functions $\left\{{\psi} \left(\frac{t-b}{a}\right)\right\}$, thus decomposing signals with these series of functions (also called wavelets).
Similar to the Fourier transform, the scale $a$ represents frequencies.
Especially, the Fourier transform can be treated as a special wavelet transform with the mother function
\begin{equation}
    \label{eq_mother_fourier}
    \bar{\psi}_{\mathrm{FT}}(t) = e^{-j2\pi t}
\end{equation}
with scale $a = \frac{1}{k}$ and translation $b = 0$ in \EQUA{eq_fourier}.
Differently, the wavelet spectrum is located by both variables $a$ and $b$, where $b$ represents the ``focused time''.
In other words, we can see the particular frequency response $X_w(a, b=t_0)$ at a specific time $t_0$ from wavelet spectrums.
This one property is not available in the Fourier transform.
It means that when analyzing agents' trajectories, especially considering the uncertainty of agents' behaviors, the wavelet spectrum could better represent the changes in this spectral distribution over time, especially in more complex dynamic scenarios.
That is the first reason we choose wavelet in \MODEL~and \EMODEL~in this manuscript.

(b) \textbf{Computational Efficiency.}
One of the other reasons why we choose Haar transform from different wavelet transforms is that it has a fairly simple calculation.
Comparing \EQUA{eq_mother_haar} and \EQUA{eq_mother_fourier}, the Haar transform contains only additions and no multiplications on $x(t)$ in the computation process when computing \EQUA{eq_wavelet}, rather than the complex integrals (or sums for discrete cases) $\int x(t) e ^{-j2\pi k t} \mathrm{d} t$.
In addition, in trajectory prediction, trajectories are usually not even-symmetric sequences, meaning that there must be imaginary parts in their Fourier spectrums.
It means that additional channels are needed for neural networks to handle the corresponding imaginary parts of the spectrum (the additional channels are still needed even if they are represented as amplitudes and phases), which leads to an increase in the number of trainable parameters in the network, also reduces the inference efficiency of the whole prediction network.
On the contrary, due to the simple mother function (\EQUA{eq_mother_haar}), calculating the Haar spectrum requires only a simple addition of real numbers, thus saving this unnecessary prediction network overhead.

~\\\textbf{Limitations of Haar Transform.}
Haar transform has its limitations due to the properties of its mother function $\psi (t)$.
One main limitation is that the Haar transform focuses more on the high-frequency portions of the signal.
It means that a signal with more low-frequency portions may not be represented well by the Haar transform.
In wavelet transforms, the $n$th moment of a mother wavelet $\psi (t)$ is defined as
\begin{equation}
    \label{eq_vanishingmoment}
    M_n = \int t^n \psi (t) \mathrm{d} t.
\end{equation}
Simply, if there exists a positive integer $p$ such that $M_0 = M_1 = ... = M_{p-1} = 0$, then the mother wavelet $\psi (t)$ has $p$ \emph{Vanishing Moments}.
A mother function with more vanishing moments means that the \emph{higher}-frequency portions may take up less energy in the corresponding transformed wavelet spectrum.
This conclusion can be intuitively explained with a simple example.
We can expand the $X_w (a, b)$ (\EQUA{eq_wavelet}) into the Taylor series at $t = 0$ with order $n$ like (the translation $b$ is set to 0 for a easier computation \cite{sheng1992optical}):
\begin{align}
    X_w (a, 0) & = \frac{1}{\sqrt{a}} \int \left(
        \sum_{p=0}^{n} x^{(p)} (0) \frac{t^p}{p!} + \mathcal{O}(n + 1)
    \right) \psi \left(\frac{t}{a}\right) \mathrm{d} t \\
    & = \frac{1}{\sqrt{a}} \left(
        \sum_{p=0}^n x^{(p)} (0) \int \frac{t^p}{p!}
        \psi \left(\frac{t}{a}\right) \mathrm{d} t + 
        \mathcal{O}(n + 1)
    \right).
\end{align}
Considering the definition of $M_k$ in \EQUA{eq_vanishingmoment}, we have
\begin{align}
    X_w (a, 0) &= \frac{1}{\sqrt{a}} \left(
        \sum_{p=0}^n \frac{x^{(p)}(0)}{p!} M_n a^{p + 1} + \mathcal{O}(n + 1)
    \right).
\end{align}
Here, it can be seen that $X_w (a, 0)$ decays as fast as the first non-zero term $\frac{x^{n}(0)}{n!} M_n a^{n + 1/2}$.
The speed of convergence of the $X_w (a, 0)$ to zero with an increase of frequency variable $k = \frac{1}{a}$ is then determined by the first non-zero moment $M_n$ of the wavelet $\psi (t)$.
Thus, a wavelet $\psi (t)$ with a higher vanishing moment $p$ may lead to a higher decay rate for the higher-frequency portions (higher $k$) of the corresponding wavelet spectrum $X_w (a=\frac{1}{k}, b)$, thus meaning a higher percentage of energy distributed in the relatively lower frequency portions in the entire spectrum.

However, the Haar mother wavelet $\psi_\mathrm{Haar}$ only has $p = 1$ vanishing moment, lower than most of the other wavelets, thus leading to a slow decay with the increase of frequency variable $k = \frac{1}{a}$.
As a result, the lower-frequency portions may occupy a relatively small amount of energy in the Haar spectrum, making it more challenging to well characterize signals $x(t)$ with more lower-frequency components, \IE, the smoother signals.
In the trajectory prediction that the manuscript focused on, the trajectories of the different agents are the signals $x(t)$ above.
It means that smoother trajectories, either because of the short observation time or the smaller amplitudes of the trajectory changes, may lead to a limited ability to encode with the Haar spectrum.
Consequently, the prediction performance of the corresponding Haar transform model variations will also be limited.

~\\\textbf{Additional Experimental Validations.}
We have conducted several ablation studies to validate the above advantages and limitations of the Haar transform and its usefulness in helping forecast heterogeneous trajectories, including the quantitative validations of transforms and the validations of model efficiencies.
Here, we add more experimental discussions to further verify our thoughts. \\

(a) \textbf{Quantitative Validations of Transforms.}
We can see from the ablation results from the main manuscript that \EMODEL-DFT and \EMODEL-Haar variations perform almost the same on 2D coordinate datasets.
Specifically, the Haar variation performs even worse than the DFT variation when forecasting 2D coordinates on SDD.
However, Haar variations outperform DFT variations on the more challenging 2D bounding box and 3D skeleton datasets.
We infer from these results that the Haar variation performs better at predicting datasets (scenarios) that contain more complex time-frequency changes, like the more complex 2D bounding box changes and significant changes in human skeletons due to their pose changes.

Notably, Haar variations perform similarly to DFT ones for 3D bounding box cases.
This phenomenon is consistent with the above-mentioned limitations of the Haar transform: the trajectories in the nuScenes dataset appear smoother due to the lower length of observations ($t_h = 4$).
In other words, in the relatively short observation period, the possibility of large changes in the trajectory will be significantly reduced, considering these agents' physical movement constraints.
The Haar transform has a vanishing moment of 1, which means that the energy is distributed more in the high-frequency portions in the Haar spectrum, making it difficult to focus on the lower-frequency portions, leading to a poor representation of the smoother trajectories.

\begin{table*}[htbp]
    \centering
    \caption{
        Ablation Studies on validating the modeling capacity of different transforms under different prediction lengths (3D bounding box cases, \emph{best-of-20}).
    }
    \label{tab_appendix_vanishingmoment}
    \begin{tabular}{c|c|c|ccccccc}
        \toprule
        No. & Model & T & $t_h$ & $t_f$ & ADE $\downarrow$ & FDE $\downarrow$ & AIoU $\uparrow$ & FIoU $\uparrow$ \\
        
        \midrule
        APP-A1 & \EMODEL & DFT & 4 & 4 & 0.203 & 0.291 & 0.766 & 0.691 \\
        APP-A2 & \EMODEL & Haar & 4 & 4 & 0.209 & 0.301 & 0.764 & 0.688 \\

        \midrule
        APP-A3 & \EMODEL & DFT & 10 & 10 & 0.730 & 1.368 & 0.565 & 0.403 \\
        APP-A4 & \EMODEL & Haar & 10 & 10 & 0.647 & 1.202 & 0.575 & 0.407 \\

        \bottomrule
    \end{tabular}

\end{table*}

We have also conducted another pair of ablation experiments to validate our thoughts on how the sequence length (or the ``smoothness'') of the input sequence affects different model variations' performances.
In \TABLE{tab_appendix_vanishingmoment}, DFT and Haar variations perform almost the same when $t_h = t_f = 4$.
The Haar variation APP-A2 even performs a little worse than the DFT variation APP-A2.
However, by increasing the length of the observed and predicted trajectories to $t_h = t_f = 10$ (by keeping the sample interval $\Delta t$), the new Haar variation APP-A4 exhibits better prediction results compared to the corresponding DFT variation APP-A3, including 11.3\% better ADE and 13.1\% FDE.
This phenomenon aligns with our results obtained on forecasting 3D human skeletons under the same $t_h = t_f = 10$ condition, where the Haar variation outperforms DFT variation for about 13.2\% MPJPE at 400ms, thus validating the idea that the Haar transform above is unsuitable for dealing with ``smoother'' trajectories.

From all these quantitative results, we can see that DFT is more effective for simpler or smoother trajectories, but Haar transform will be more effective for heterogeneous trajectories with higher dimensionality or fast time-frequency changing properties.
These experiments also demonstrate the different advantages of the different transforms, thus supporting the choice of transforms for different scenarios. \\

(b) \textbf{Validations of Model Efficiencies.}
We mentioned in the above theoretical analyses that the Haar transform has better computation efficiency than DFT.
We have conducted several ablation variations of both \MODEL~and \EMODEL~to verify these properties in the main manuscript.
Here, we provide more efficiency-related experimental validations to analyze how trajectory length, dimensionality (trajectory types), and number (keypoints selection) affect the speed and memory.
Their results are reported in \TABLE{tab_app_speed,tab_app_speed_keypoints}.
It is easy to observe that the Haar variations are more efficient (\IE, less inference time and memory) than DFT variations in the same setting, \EG, comparing APP-A5 and APP-A6 or APP-A31 and APP-A32.

\begin{table*}[htbp]
    \centering
    \caption{
        Inference time and memory of \EMODEL~and \MODEL~(DFT and Haar, \LINEARNOTE) at different trajectory lengths (frames) on one Apple M1 chip.
        $b$, $L$, $t_h$, and $t_f$ denote ``batch size'', ``trajectory length'', ``the observed trajectory length'', and ``the predicted trajectory length'', respectively.
        Results are obtained by generating $k=20$ trajectories for each prediction sample.
        The reported results are shown as ``Time (ms)/Para. (MB)''.
    }
    \label{tab_app_speed}
    \begin{tabular}{c|c|c|cccccc}
        \toprule
        No. & Model & $b$, $L$, $t_h \rightarrow t_f $ & co $(M = 2)$ & bb $(M = 4)$ & 3dbb $(M = 6)$ & ske $(M =51)$ \\

        \midrule
        APP-A5 & \EMODELSHORT-DFT & \multirow{4}{*}{1, 16, $8 \rightarrow 8$} & 29 ms/8.0M & 33 ms/8.0M & 37 ms/8.0M & 111 ms/8.4M \\
        APP-A6 & \EMODELSHORT-Haar & & 26 ms/7.7M & 27 ms/7.8M & 27 ms/7.8M & 27 ms/8.0M \\
        APP-A7 & \MODELSHORT-DFT & & 27 ms/7.4M & 29 ms/7.4M & 36 ms/7.4M & 111 ms/7.9M \\
        APP-A8 & \MODELSHORT-Haar & & 25 ms/7.3M & 25 ms/7.3M & 26 ms/7.3M & 26 ms/7.6M \\

        \midrule
        APP-A9 & \EMODELSHORT-DFT & \multirow{4}{*}{1, 20, $8 \rightarrow 12$} & 29 ms/8.0M & 34 ms/8.0M & 37 ms/8.1M & 112 ms/8.6M \\
        APP-A10 & \EMODELSHORT-Haar & & 27 ms/7.7M & 27 ms/7.8M & 28 ms/7.8M & 29 ms/8.1M \\
        APP-A11 & \MODELSHORT-DFT & & 30 ms/7.4M & 30 ms/7.4M & 37 ms/7.5M & 112 ms/8.1M \\
        APP-A12 & \MODELSHORT-Haar & & 26 ms/7.3M & 27 ms/7.3M & 27 ms/7.3M & 28 ms/7.7M \\

        \midrule
        APP-A13 & \EMODELSHORT-DFT & \multirow{4}{*}{1, 24, $8 \rightarrow 16$} & 29 ms/8.0M & 36 ms/8.1M & 38 ms/8.1M & 112 ms/8.8M \\
        APP-A14 & \EMODELSHORT-Haar & & 26 ms/7.7M & 27 ms/7.8M & 27 ms/7.8M & 28 ms/8.3M \\
        APP-A15 & \MODELSHORT-DFT & & 28 ms/7.4M & 32 ms/7.4M & 38 ms/7.5M & 112 ms/8.3M \\
        APP-A16 & \MODELSHORT-Haar & & 26 ms/7.3M & 26 ms/7.3M & 27 ms/7.4M & 28 ms/7.8M \\

        \midrule
        APP-A17 & \EMODELSHORT-DFT & \multirow{4}{*}{1, 28, $8 \rightarrow 20$} & 29 ms/8.0M & 36 ms/8.1M & 38 ms/8.1M & 113 ms/9.0M \\
        APP-A18 & \EMODELSHORT-Haar & & 26 ms/7.8M & 27 ms/7.8M & 28 ms/7.8M & 28 ms/8.4M \\
        APP-A19 & \MODELSHORT-DFT & & 28 ms/7.4M & 33 ms/7.5M & 38 ms/7.5M & 112 ms/8.5M \\
        APP-A20 & \MODELSHORT-Haar & & 26 ms/7.3M & 27 ms/7.4M & 27 ms/7.4M & 28 ms/8.0M \\

        \midrule
        APP-A21 & \EMODELSHORT-DFT & \multirow{4}{*}{1, 24, $4 \rightarrow 20$} & 29 ms/7.8M & 35 ms/7.8M & 38 ms/7.9M & 112 ms/8.9M \\
        APP-A22 & \EMODELSHORT-Haar & & 26 ms/7.6M & 27 ms/7.7M & 28 ms/7.7M & 28 ms/8.3M \\
        APP-A23 & \MODELSHORT-DFT & & 27 ms/7.4M & 33 ms/7.4M & 38 ms/7.5M & 109 ms/8.5M \\
        APP-A24 & \MODELSHORT-Haar & & 26 ms/7.3M & 27 ms/7.4M & 28 ms/7.4M & 28 ms/8.0M \\

        \midrule
        APP-A25& \EMODELSHORT-DFT & \multirow{4}{*}{1, 30, $10 \rightarrow 20$} & 29 ms/8.2M & 36 ms/8.2M & 39 ms/8.3M & 113 ms/9.2M \\
        APP-A26& \EMODELSHORT-Haar & & 26 ms/7.9M & 27 ms/7.9M & 28 ms/7.9M & 28 ms/8.4M \\
        APP-A27 & \MODELSHORT-DFT & & 29 ms/7.5M & 33 ms/7.6M & 39 ms/7.6M & 112 ms/8.6M \\
        APP-A28 & \MODELSHORT-Haar & & 26 ms/7.4M & 27 ms/7.4M & 28 ms/7.4M & 29 ms/8.0M \\

        \midrule
        APP-A29 & \EMODELSHORT-DFT & \multirow{4}{*}{1, 40, $20 \rightarrow 20$} & 29 ms/8.8M & 36 ms/8.9M & 40 ms/8.9M & 113 ms/9.7M \\
        APP-A30 & \EMODELSHORT-Haar & & 26 ms/8.2M & 27 ms/8.2M & 28 ms/8.2M & 29 ms/8.7M \\
        APP-A31 & \MODELSHORT-DFT & & 29 ms/7.7M & 35 ms/7.7M & 40 ms/7.7M & 112 ms/8.8M \\
        APP-A32 & \MODELSHORT-Haar & & 26 ms/7.5M & 27 ms/7.5M & 27 ms/7.5M & 29 ms/8.1M \\

        \bottomrule
\end{tabular}
\end{table*}

\begin{table*}[htbp]
    \centering
    \caption{
        Inference time and memory of \EMODEL~and \MODEL~(DFT and Haar, \LINEARNOTE) at different keypoints when $(t_h = 8, t_f = 20)$ on one Apple M1 chip.
        $b$, $N_{key}$ denote ``batch size'', and ``the number of keypoints'', respectively.
        Results are obtained by generating $k=20$ trajectories for each prediction sample.
        The reported results are shown as ``Time (ms)/Para. (MB)''.
    }
    \label{tab_app_speed_keypoints}
    \begin{tabular}{c|c|c|c|cccc}
        \toprule
        No. & Model & $b$ & $N_{key}$ & co $(M = 2)$ & bb $(M = 4)$ & 3dbb $(M = 6)$ & ske $(M =51)$ \\

        \midrule
        APP-A33 & \EMODELSHORT-DFT & \multirow{4}{*}{1} & \multirow{4}{*}{2} & 28 ms/8.0M & 33 ms/8.0M & 35 ms/8.0M & 109 ms/8.0M \\
        APP-A34 & \EMODELSHORT-Haar & & & 27 ms/7.7M & 27 ms/7.7M & 27 ms/7.7M & 28 ms/7.9M \\
        APP-A35 & \MODELSHORT-DFT & & & 27 ms/7.4M & 31 ms/7.4M & 34 ms/7.4M & 108 ms/7.5M \\
        APP-A36 & \MODELSHORT-Haar & & & 25 ms/7.3M & 25 ms/7.3M & 26 ms/7.3M & 27 ms/7.4M \\

        \midrule
        APP-A37 & \EMODELSHORT-DFT & \multirow{4}{*}{1} & \multirow{4}{*}{4} & 29 ms/8.0M & 36 ms/8.0M & 38 ms/8.0M & 113 ms/8.2M \\
        APP-A38 & \EMODELSHORT-Haar & & & 27 ms/7.7M & 27 ms/7.7M & 28 ms/7.7M & 28 ms/7.9M \\
        APP-A39 & \MODELSHORT-DFT & & & 27 ms/7.4M & 33 ms/7.4M & 38 ms/7.4M & 112 ms/7.7M \\
        APP-A40 & \MODELSHORT-Haar & & & 26 ms/7.4M & 27 ms/7.4M & 28 ms/7.4M & 28 ms/7.5M \\

        \bottomrule
    \end{tabular}

\end{table*}

\textbf{Analyses of Inference Speed/Memory Influenced by Trajectory Length.}
The trajectory length consists of two parts: the observed trajectory length $t_h$ and the predicted trajectory length $t_f$.
\TABLE{tab_app_speed} showes the results of inference time and memory under different $(t_h, t_f)$ pairs under different $bs$ settings, respectively.
As shown in \TABLE{tab_app_speed}, when $bs = 1$, as the trajectory length increases, \EMODEL~and \MODEL~variations' inference time and memory are similar, comparing APP-A5, APP-A9, APP-A13 and APP-A17.
As the trajectory length increases, the inference time of models maintains the same level, but the memory of \EMODEL-DFT has increased by up to 12.8\%, comparing APP-A21, APP-A17, APP-A25, and APP-A29.
Moreover, comparing APP-A13 ($(t_h = 8, t_f = 16)$) and APP-A21 ($(t_h = 4, t_f = 20)$) under the same trajectory length $L = 24$, the inference time is similar but APP-A23 takes up about 1\%-3\% more memory than APP-A21 in trajectory form of 2D coordinate, 2D bounding box and 3D bounding box.
In contrast, APP-A21 takes up about 6.0\% more memory than APP-A23 in 3D human skeleton prediction.
As a result, the trajectory length has a certain impact on model efficiency, especially the longer the trajectory, the more obvious the impact.

\textbf{Analyses of Inference Speed/Memory Influenced by Trajectory Types.}
As shown in \TABLE{tab_app_speed}, as the trajectory dimensionality rises, the inference time and memory also increase.
Taking APP-A13 as an example, comparing with variation $M = 2$, the inference time and memory of DFT variations $M = 4$, $M = 6$, and $M = 51$ increase 24.1\%/1.3\%, 31.0\%/1.3\%, and 286\%/10.0\%, respectively.
Correspondingly, Haar variations APP-A14 also increase 3.8\%/0 ($M = 4$), 3.8\%/0 ($M = 6$), 7.7\%/7.8\% ($M = 51$) under the same trajectory dimensionality settings, compared with variation $M = 2$.
This phenomenon indicates that Haar variations have better computation ability when the trajectory dimensionality increases.
According to the results, trajectory types significantly impact model efficiency, especially the higher the trajectory dimension, the more significant the impact.

\textbf{Analyses of Inference Speed/Memory Influenced by the Number of Keypoints.}
As shown in \TABLE{tab_app_speed_keypoints}, comparing APP-A33 and APP-A37, the memory of models in 2D coordinate, 2D bounding box, and 3D bounding box maintain the same level, but the inference time and memory of variation $N_{key} = 4$ increase 2.5\% in 3D human skeleton than variation $N_{key} = 2$.
Moreover, the inference time of variation $N_{key} = 4$ increase 3.6\%, 9.0\%, 8.6\%, 3.7\%, respectively,than variation $N_{key} = 2$.
In short, the number of keypoints is limitedly affected by the trajectory length.

From the above efficiency comparisons, we can further verify the advantages of the Haar transform to the DFT.
In particular, the Haar variations show better computational speed and require less memory for high-dimensional heterogeneous trajectories.
Considering the quantitative performance of the Haar variations discussed above, we can conclude that the Haar transform is more effective in these high dimensional trajectory predictions relative to the DFT.
That is why we choose Haar transform as an alternative in this manuscript.

\section{Transformer Details}
\label{appendix_transformer}

We employ the Transformer \cite{vaswani2017attention} as the backbone to encode trajectory spectrums and the scene context in the two proposed sub-networks.
The Transformer used in the \EMODEL~has two main parts, the Transformer Encoder and the Transformer Decoder, both of which are made up of several attention layers.

~\\\textbf{Attention Layers.}
Multi-Head Attention operations are applied in each of the attention layers.
Following \cite{vaswani2017attention}, each layer's multi-head dot product attention with $H$ heads is calculated as:
\begin{equation}
    \mbox{Attention}(\mathbf{q}, \mathbf{k}, \mathbf{v}) = \mbox{softmax}\left(\frac{\mathbf{q}\mathbf{k}^{\top}}{\sqrt{d}}\right)\mathbf{v},
\end{equation}
\begin{equation}
    \begin{aligned}
        \mbox{MultiHead}&(\mathbf{q}, \mathbf{k}, \mathbf{v}) = \\ &\mbox{fc}\left(\mbox{concat}(\left\{ \mbox{Attention}_i(\mathbf{q}, \mathbf{k}, \mathbf{v}) \right\}_{i=1}^H)\right).
    \end{aligned}
\end{equation}
Here, $\mbox{fc}()$ denotes one fully connected layer that concatenates all heads' outputs.
Query matrix $\mathbf{q}$, key matrix $\mathbf{k}$, and value matrix $\mathbf{v}$ are the three layer inputs.
Each attention layer also contains an MLP (denoted as MLP$_a$) to extract the attention features further.
It contains two fully connected layers.
ReLU activations are applied in the first layer.
Formally, we have output feature $\mathbf{f}_o$ of this layer:
\begin{equation}
    \mathbf{f}_{o} = \mbox{ATT}(\mathbf{q}, \mathbf{k}, \mathbf{v}) = \mbox{MLP}_a(\mbox{MultiHead}(\mathbf{q}, \mathbf{k}, \mathbf{v})).
\end{equation}

~\\\textbf{Transformer Encoder.}
The transformer encoder comprises several encoder layers, and each encoder layer contains an attention layer and an encoder MLP (MLP$_e$).
Residual connections and normalization operations are applied to prevent the network from overfitting.
Let $\mathbf{h}^{(l+1)}$ denote the output of $l$-th encoder layer, and $\mathbf{h}^{(0)}$ denote the encoder's initial input.
For $l$-th encoder layer, the calculation of the layer output $\mathbf{h}^{(l+1)}$ can be written as:
\begin{equation}
    \label{eq_alpha_encoder}
    \begin{aligned}
        \mathbf{a}^{(l)}   & = \mbox{ATT}(\mathbf{h}^{(l)}, \mathbf{h}^{(l)}, \mathbf{h}^{(l)}) + \mathbf{h}^{(l)}, \\
        \mathbf{a}^{(l)}_n & = \mbox{Normalization}(\mathbf{a}^{(l)}),                   \\
        \mathbf{c}^{(l)}   & = \mbox{MLP}_e(\mathbf{a}_n^{(l)}) + \mathbf{a}_n^{(l)},             \\
        \mathbf{h}^{(l+1)} & = \mbox{Normalization}(\mathbf{c}^{(l)}).
    \end{aligned}
\end{equation}

~\\\textbf{Transformer Decoder.}
Similar to the Transformer encoder, the Transformer decoder comprises several decoder layers, and each is stacked with two different attention layers.
The first attention layer in the Transformer decoder focuses on the essential parts in the Transformer encoder's outputs $\mathbf{h}_e$ queried by the decoder's input $\mathbf{X}$.
The second layer is the same self-attention layer as in the encoder.
Similar to \EQUA{eq_alpha_encoder}, we have the decoder layer's output feature $\mathbf{h}^{(l+1)}$:
\begin{equation}
    \label{eq_alpha_decoder}
    \begin{aligned}
        \mathbf{a}^{(l)}      & = \mbox{ATT}(\mathbf{h}^{(l)}, \mathbf{h}^{(l)}, \mathbf{h}^{(l)}) + \mathbf{h}^{(l)}, \\
        \mathbf{a}^{(l)}_n    & = \mbox{Normalization}(\mathbf{a}^{(l)}),                   \\
        \mathbf{a}_2^{(l)}    & = \mbox{ATT}(\mathbf{h}_e, \mathbf{h}^{(l)}, \mathbf{h}^{(l)}) + \mathbf{h}^{(l)},     \\
        \mathbf{a}_{2n}^{(l)} & = \mbox{Normalization}(\mathbf{a}_2^{(l)})                  \\
        \mathbf{c}^{(l)}      & = \mbox{MLP}_d(\mathbf{a}_{2n}^{(l)}) + \mathbf{a}_{2n}^{(l)},       \\
        \mathbf{h}^{(l+1)}    & = \mbox{Normalization}(\mathbf{c}^{(l)}).
    \end{aligned}
\end{equation}

~\\\textbf{Positional Encoding.}
Before feeding agents representations or trajectory spectrums into the Transformer, we add the positional coding to inform the relative position of each timestep or frequency portion in the sequential inputs.
The position coding $\mathbf{f}_e^t$ at step $t~(1 \leq t \leq t_h)$ is obtained by:
\begin{equation}
    \begin{aligned}
        \mathbf{f}_e^t                  & = \left({f_e^t}_0, ..., {f_e^t}_i, ..., {f_e^t}_{d-1}\right) \in \mathbb{R}^{d}, \\
        \mbox{where}~{f_e^t}_i & = \left\{\begin{aligned}
             & \sin \left(t / 10000^{d/i}\right),     & i \mbox{ is even}; \\
             & \cos \left(t / 10000^{d/(i-1)}\right), & i \mbox{ is odd}.
        \end{aligned}\right.
    \end{aligned}
\end{equation}
Then, we have the positional coding matrix $\mathbf{f}_e$ that describes $t_h$ steps of sequences:
\begin{equation}
    \mathbf{f}_e = (\mathbf{f}_e^1, \mathbf{f}_e^2, ..., \mathbf{f}_e^{t_h})^{\top} \in \mathbb{R}^{{t_h}\times d}.
\end{equation}
The final Transformer input $\mathbf{X}_T$ is the addition of the original sequential input $\mathbf{X}$ and the positional coding matrix $\mathbf{f}_e$.
Formally,
\begin{equation}
    \mathbf{X}_T = \mathbf{X} + \mathbf{f}_e \in \mathbb{R}^{t_h \times d}.
\end{equation}

~\\\textbf{Layer Configurations.}
We employ $L = 4$ layers of encoder-decoder structure with $H = 8$ attention heads in each Transformer-based sub-networks.
The MLP$_e$ and the MLP$_d$ have the same shape.
Both of them consist of two fully connected layers.
The first layer has 512 output units with the ReLU activation, and the second layer has 128 but does not use any activations.
The output dimensions of fully connected layers used in multi-head attention layers are set to $d$ = 128.

~\\\textbf{Difference between Dimension-wise Interactions and Positional Encoding.}
Transformers focus more on modeling the relations of different moments (different steps) in the sequence.
They use positional encoding to locate the position of different time steps in the temporal sequence and then adopt the attention mechanism to obtain the similarity of information in different time steps.
It means that the positions of some moments will be paid with more attention among all moments in the temporal sequence in the Transformers.

\begin{figure}[h]
    \centering
    \includegraphics[width=0.8\linewidth]{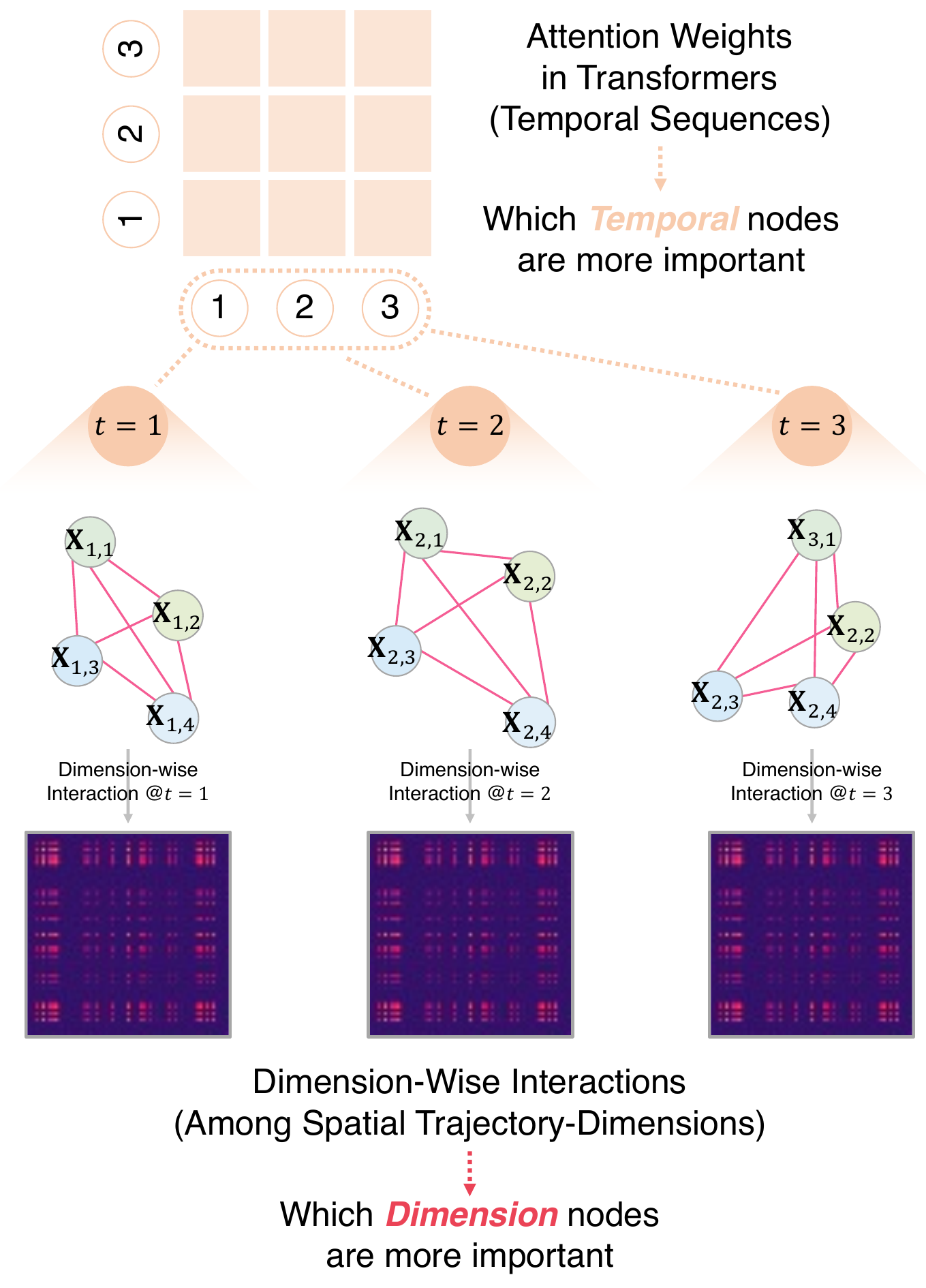}
    \caption{
        Different roles of the Transformer and the dimension-wise interaction, taking a time-sequence with three times steps $t \in \{1, 2, 3\}$ and four trajectory dimensions ($M=4$) as an example.
    }
    \label{fig_app_dwi_bs}
\end{figure}

However, the dimensional-wise interaction this manuscript studied refers to the interaction within different trajectory-dimensions at a certain time (frequency) step.
We take a simple example to explain the difference between the Transformer and the dimension-wise interaction in \FIG{fig_app_dwi_bs} for easier understanding.
Suppose there are three time steps $t \in \{1, 2, 3\}$, and four trajectory dimensions ($M=4$), \IE, we have the trajectory matrix $\mathbf{X} \in \mathbb{R}^{3 \times 4}$.
Thus, the four trajectory-dimensions focused in this manuscript are $\{ \mathbf{X}_{:,1}, \mathbf{X}_{:,2}, \mathbf{X}_{:,3}, \mathbf{X}_{:,4}\}$, where each trajectory-dimension $\mathbf{X}_{:,m} \in \mathbb{R}^{3}, m \in \{1, 2, 3, 4\}$.

When modeling the trajectory $\mathbf{X}$ in Transformers, it will be applied the positional encoding to locate the relative positions of different temporal moments $t$ (the corresponding trajectory is $\mathbf{X}_t \in \mathbb{R}^{4}$), and then the attention mechanism will be applied to obtain the similarity of information in these three time steps ${\mathbf{X}_1, \mathbf{X}_2, \mathbf{X}_3}$.
Simply, it can be treated as a matrix with three rows and three columns to indicate the relations between these steps.
Then, the most attentive steps (like $\mathbf{X}_{t_0}$) will be specifically taken into account when encoding the trajectory $\mathbf{X}$.

However, the bilinear structure aims to model the dimension-wise interactions among the four trajectory-dimensions $\{\mathbf{X}_{:,1}, \mathbf{X}_{:,2}, \mathbf{X}_{:,3}, \mathbf{X}_{:,4}\}$.
Correspondingly, we aim at building connections to describe the relations, for example it can be simply represented by a matrix with four rows and four columns.
Finally, it aims at considering the relations between different trajectory-dimensions, for example, which dimensionality $\mathbf{X}_{:,m_0}$ would affect the forecasted $\hat{\mathbf{X}}_{:,m}$ the most.

Thus, as shown in \FIG{fig_app_dwi_bs}, the most significant difference between the dimension-wise interaction and the positional encoding in Transformers is that they can consider different ``aspects'' of the input sequence.
Transformer focuses on the relations between different time steps, and the bilinear structure focuses on the interactions between all trajectory-dimensions.
In the above example (trajectory $\mathbf{X} \in \mathbb{R}^{3 \times 4}$), dimension-wise interactions concern more from the last tensor-dimension $\{\mathbf{X}_{:,1}, \mathbf{X}_{:,2}, \mathbf{X}_{:,3}, \mathbf{X}_{:,4}\}$ while the other one focuses from the first (temporal) tensor-dimension $\{\mathbf{X}_1, \mathbf{X}_2, \mathbf{X}_3\}$.
There are no such structures in Transformers that could directly describe the relations among trajectory-dimensions $\{\mathbf{X}_{:,1}, \mathbf{X}_{:,2}, \mathbf{X}_{:,3}, \mathbf{X}_{:,4}\}$.
Thus, we use the bilinear structure to represent the above relations, \IE, dimension-wise interactions, upon the Transformer backbone to further capture their time-frequency properties simultaneously.

\section{Linear Least Squares Trajectory Prediction}
\label{appendix_lls}

The linear least squares trajectory prediction method aims to minimize the mean square error between the predicted and agents' groundtruth trajectories.
When predicting, we perform a separate least squares operation for each dimension of the $M$-dimensional observed trajectory $\mathbf{X}$.
Simply, we want to find the $\mathbf{x}_m = (b_m, w_m)^{\top} \in \mathbb{R}^2~(1 \leq m \leq M)$, such that
\begin{equation}
    \begin{aligned}
        \hat{\mathbf{Y}} &= (\hat{\mathbf{Y}}_1, \hat{\mathbf{Y}}_2, ..., \hat{\mathbf{Y}}_m, ..., \hat{\mathbf{Y}}_M) \in \mathbb{R}^{t_f \times M}, \\
        \mbox{where}~&\hat{\mathbf{Y}}_m = \mathbf{A}_f\mathbf{x}_m = \left(\begin{matrix}
            1 & t_h + 1 \\
            1 & t_h + 2 \\
            ... & ... \\
            1 & t_h + t_f 
        \end{matrix}\right)\left(\begin{matrix}
            b_m \\
            w_m 
        \end{matrix}\right).
    \end{aligned}
\end{equation}

For one of agents' observed $M$-dimensional trajectory $\mathbf{X} \in \mathbb{R}^{t_h \times M}$, we have the trajectory slice on the $m$-th dimension
\begin{equation}
    \mathbf{X}_m = ({r_m}_1, {r_m}_2, ..., {r_m}_{t_h})^{\top}.
\end{equation}
Suppose we have a coefficient matrix $\mathbf{A}_h$, where
\begin{equation}
    \mathbf{A}_h = \left(\begin{matrix}
        1 & 1 & 1 & ... & 1 \\
        1 & 2 & 3 & ... & t_h 
    \end{matrix}\right)^{\top}.
\end{equation}
We aim to find a $\mathbf{x}_m \in \mathbb{R}^{2}$, such that the mean square $\Vert \mathbf{A}_h \mathbf{x}_m - \mathbf{X}_m \Vert_2^2$ could reach its minimum value.
Under this condition, we have
\begin{equation}
    \mathbf{x}_m = (\mathbf{A}_h^{\top} \mathbf{A}_h)^{-1} \mathbf{A}_h^{\top} \mathbf{X}_m.
\end{equation}
Then, we have the predicted $m$-th dimension trajectory
\begin{equation}
    \hat{\mathbf{Y}}_m = \mathbf{A}_f \mathbf{x}_m.
\end{equation}

The final $M$-dimensional predicted trajectory $\hat{\mathbf{Y}}$ is obtained by stacking all results.
Formally,
\begin{equation}
    \hat{\mathbf{Y}} = \mathbf{A}_f (\mathbf{x}_1, \mathbf{x}_2, ..., \mathbf{x}_M).
\end{equation}

\section{2D DFT v.s. Bilinear Structure}
\label{appendix_2ddft}

\begin{figure}[tb]
    \centering
    \includegraphics[width=1.0\linewidth]{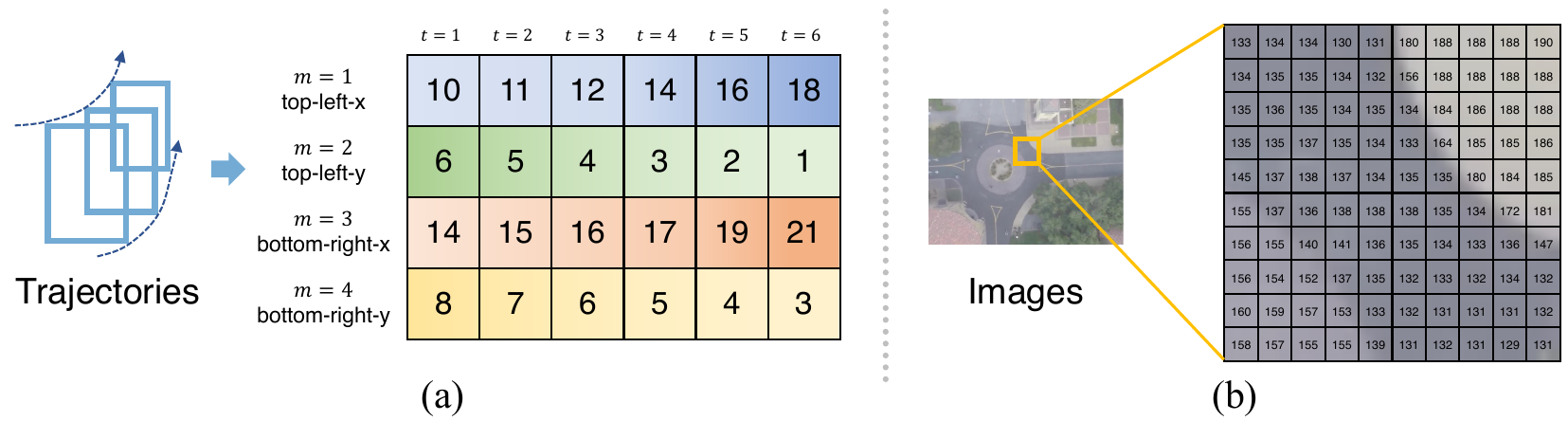}
    \caption{
        Matrices views of a trajectory (2D bounding box) and an image.
    }
    \label{fig_appendix_dft2d}
\end{figure}

We apply different transforms on \emph{each dimension} of the trajectory to obtain the corresponding trajectory spectrums either in \MODEL~or the enhanced \EMODEL.
Moreover, considering that one of our main contributions is to establish connections between trajectories (or spectrums) of different dimensions, a more natural idea might be to apply some 2D transform directly to these trajectories.
However, it appears to be less effective from both theoretical analyses and experimental results.
In this section, we will discuss the discrepancy between the 2D transform and the proposed bilinear structure in describing the two factors, including the frequency response of the trajectory and the dimension-wise interactions, from different perspectives, taking DFT as an example.

~\\\textbf{DFT on Different Directions in Trajectories.}
The 2D DFT can be decomposed into two consecutive 1D DFTs performed in different directions of the target 2D matrix.
The $M$-dimensional trajectory $\mathbf{X} \in \mathbb{R}^{N \times M}$ is also a 2D matrix similar to 2D grayscale images.
Although the 2D DFT and its variations have achieved impressive results in tasks related to image processing, they might not be directly applied to trajectories.
We will analyze this problem specifically by focusing on the different directions of the transforms in the trajectory.

\FIG{fig_appendix_dft2d} shows an $M=4$ 2D bounding box trajectory and an image with the matrix view.
As shown in \FIG{fig_appendix_dft2d} (b), whether the image is sliced horizontally or vertically, the resulting vector could reflect the change in grayscale values in a particular direction.
Therefore, when performing the 2D transform, the first 1D transform will extract the frequency response in a specific direction, while the second 1D transform will fuse it with the frequency response in the vertical direction.

In contrast, different slice directions of the trajectory may lead to different meanings.
If the trajectories are sliced according to the time dimension, then four 1D time series will be obtained as shown in \FIG{fig_appendix_dft2d} (a).
Applying 1D transforms to these four sequences, we can obtain four trajectory spectrums that could describe agents' frequency responses and thus describe their motions from the global plannings and interaction details at different scales.
However, if the trajectory is sliced from the dimensional direction, then $N$ (6 in the figure) 4-dimensional vectors will be obtained.
These vectors contain information about agents' locations and postures at a particular moment.
In addition, the focused dimension-wise interactions are also contained in these vectors.
However, it should be noted that we are more interested in the relationships between the data in these vectors, \IE, the ``edges'' between the different data.
If a 1D transform is applied to these vectors, the resulting spectrum may hardly have a clear physical meaning, because the temporal or spatial adjacencies of these points are not reflected in these 4-dimensional vectors.

For example, suppose we want to apply the 1D DFT on the 4-dimensional (2D bounding box) vector $\mathbf{x} = (x_l, y_l, x_r, y_r)^{\top}$.
Simply, we have:
\begin{equation}
    \begin{aligned}
        \mathbf{\mathcal{X}} = \mbox{DFT}[\mathbf{x}] &= 
            \begin{pmatrix}
                1 & 1 & 1 & 1 \\
                1 & -j & -1 & j \\
                1 & -1 & 1 & -1 \\
                1 & j & -1 & -j \\
            \end{pmatrix}
            \begin{pmatrix}
                x_l \\ y_l \\ x_r \\ y_r \\
            \end{pmatrix} \\
        &= \begin{pmatrix}
                x_l + y_l + x_r + y_r \\
                x_l - x_r - j(y_l - y_r) \\
                x_l - y_l + x_r - y_r \\
                x_l - x_r + j(y_l - y_r) \\
            \end{pmatrix}.
    \end{aligned}
\end{equation}
Accordingly, we have its fundamental frequency portion $\mathbf{\mathcal{X}}[0] = x_l + y_l + x_r + y_r$ and the high-frequency portion $\mathbf{\mathcal{X}}[2] = x_l - y_l + x_r - y_r$.
However, since the four positions $\{x_l, y_l, x_r, y_r\}$ do not have specific time-dependent or space-dependent like time-sequences and images, these frequency components may hardly reflect the specific frequency response.
For example, the fundamental frequencies can represent their average value for a time series, yet the values obtained by directly summing the 4 position coordinates of the 2 points of the 2D bounding box would be uninterpretable.
In other words, each element in this 4-dimensional vector is relatively independent, and their connection relationships are more like a \emph{graph} rather than a sequence where an order is required.

\begin{figure}[tb]
    \centering
    \includegraphics[width=1.0\linewidth]{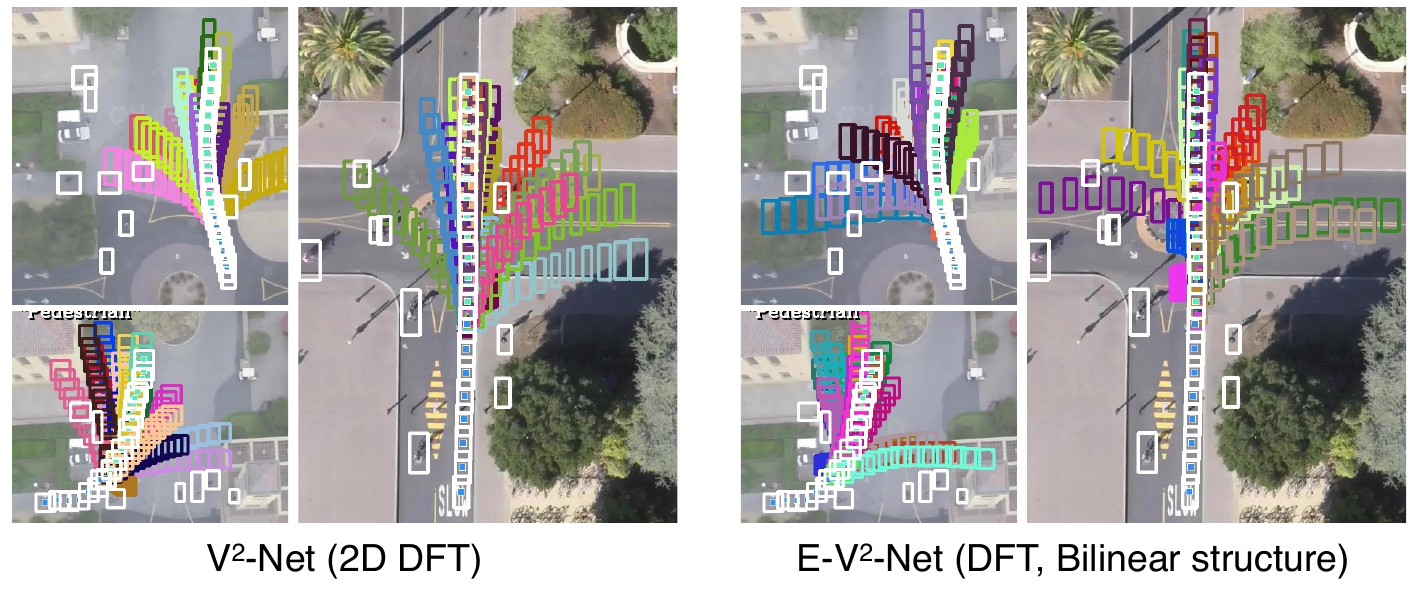}
    \caption{
        Visualized comparisons of 2D DFT bilinear structure.
    }
    \label{fig_appendix_dft2dvis}
\end{figure}

\begin{table*}[tbp]
    \centering
    \caption{
        Validation of 2D DFT and bilinear structures with \emph{best-of-20} on SDD (2D bounding box) the nuScenes (3D bounding box).
    }
    \label{tab_ab_appendix}
    \begin{tabular}{c|c|cccc|c|cccc}
        \toprule
        No. & Model & Type & T & $N_{key}$ & BS & Dataset & ADE $\downarrow$ & FDE $\downarrow$ & AIoU $\uparrow$ & FIoU $\uparrow$ \\

        \midrule
        APP-B1 & \MODEL & bb & DFT & 3 & \xm & \multirow{3}{*}{\makecell[c]{SDD\\(2D bounding box)}} &
        6.78 & 10.73 & 0.717 & 0.601 \\

        APP-B2 & \MODEL & bb & 2D DFT & 3 & \xm & &
        6.74 & 10.84 & 0.723 & 0.602 \\

        APP-B3 & \EMODEL & bb & DFT & 3 & \cm & &
        6.62 & 10.57 & 0.725 & 0.604 \\

        \midrule
        APP-B4 & \MODEL & 3dbb & DFT & 2 & \xm & \multirow{3}{*}{\makecell[c]{nuScenes\\(3D bounding box)}} &
        0.229 & 0.335 & 0.747 & 0.666 \\

        APP-B5 & \MODEL & 3dbb & 2D DFT & 2 & \xm & &
        0.234 & 0.341 & 0.739 & 0.656 \\

        APP-B6 & \EMODEL & 3dbb & DFT & 2 & \cm & &
        0.210 & 0.300 & 0.762 & 0.688 \\

        \bottomrule
    \end{tabular}
\end{table*}

~\\\textbf{Quantitative Analyses.}
To verify our thoughts, we perform ablation experiments on SDD and nuScenes to compare the effects of 2D DFT and the bilinear structure quantitatively.
As shown in \TABLE{tab_ab_appendix}, the results of APP-B2 and APP-B3 (or APP-B5 and APP-B6) show that 2D DFT does not improve quantitative trajectory prediction performance as effectively as bilinear structures.
On the contrary, in the more complex 3D bounding boxes ($M = 6$) prediction, using 2D DFT instead degrades the prediction performance compared to 1D DFT when no bilinear structures are used.
These experimental results validate our thoughts of not using 2D transforms but bilinear structures.

~\\\textbf{Qualitative Analyses.}
We visualize the prediction results of different models under the effect of 2D DFT and bilinear structure qualitatively.
As shown in \FIG{fig_appendix_dft2dvis}, the \MODEL~(2D DFT) performs not as well as \EMODEL~in both the prediction of agent motions and the interactions within the bounding box.
In detail, predictions given by \MODEL~(2D DFT) capture fewer path possibilities in the top left prediction scenario.
In addition, some predicted trajectories are with less smoothness and naturalness.
For example, the prediction in color \textbf{\color[HTML]{DF6091}{\#DF6091}} to the left of the bottom left prediction scene gives a turn with a large angle to observation, which could not be physical-acceptable in the actual scenario.
In contrast, predictions given by \EMODEL~have not shown similar results in this scenario.
On the other hand, as shown in the traffic circle prediction scenario on the right, the shape of the bounding box is not well maintained in \MODEL's predictions, such as the prediction in color \textbf{\color[HTML]{93C3CA}{\#93C3CA}} to turn right to across the street.


\section{Additional Validations of Social Interations}
\label{appendix_generalization}

When forecasting heterogeneous trajectories, interactions are more complex, including social interactions and dimension-wise interactions.
To further validate the generalizability, we conduct a series of experiments to explore the contributions of social and dimension-wise interactions to the predicted trajectories.

When forecasting agent-$i$'s position $\hat{p}^m_{t+1}$ at the next time step $t+1$, the prediction process can be represented as:
\begin{equation}
    \hat{p}^m_{t+1} = \mathrm{Net} \left(p^m_t, I_{soc}, I_{dwi} \right).
\end{equation}
Here, $\mathbf{p}^i_t = \left(p^1_t, p^2_t, ..., p^M_t \right)^\top \in \mathbb{R}^M$ denotes agent-$i$'s positions during the observion period.
$I_{soc} = I\left(\mathbf{p}^i_t, \{\mathbf{p}^j_t\}_{j \neq i} \right)$ indicates the social interaction.
$I_{dwi} = I\left(p^1_t, p^2_t, ..., p^m_t, ..., p^M_t \right)$ indicates the dimension-wise interaction.
Thus, when forecasting trajectories, two extra factors, the social interaction and the dimension-wise interaction, are considered in the proposed methods simultaneously.

To analyze the respective contributions of the social interaction and the dimension-wise interaction to the prediction performance, we take the \EMODEL-DFT as an example to conduct a series of experiments on heterogeneous trajectories (2D coordinates and 2D and 3D bounding boxes).
Specifically, the corresponding contributions of dimension-wise interactions $I_{dwi}$ (donate as $C_{dwi}$) and contributions of social interactions $I_{soc}$ (donate as $C_{soc}$) to \EMODEL-DFT's prediction performance are obtained by calculating the ratio of the sum of squares of dimension-wise interactions features $\mathbf{f}_{dwi}$ and social interactions features $\mathbf{f}_c$.
Formally,
\begin{align}
    C_{soc} & = \frac{\mathbf{f}_c^\top\mathbf{f}_c}{\mathbf{f}_c^\top\mathbf{f}_c + \mathbf{f}_{dwi}^\top\mathbf{f}_{dwi}}, \\
    C_{dwi} & = 1- C_{soc}.
\end{align}
Notably, this is a simple metric to measure the contributions of social interactions and dimension-wise interactions to the prediction network, and it is only used for rough analyses.

We conduct a series of corresponding quantitative experiments to analyze the factor that the prediction network depends more on when forecasting heterogeneous trajectories, like trajectory dimensionality or other factors.
The human skeleton dataset Human3.6M contains nothing about social interactions among subjects.
Thus, we only conduct experiments to verify the contributions of these factors on 2D coordinate, 2D bounding box, and 3D bounding box datasets.
Their results are shown in \TABLE{tab_app_dwi_soc}.
We can see that as the dimension of a frame of trajectory vector $M$ increases, the contribution of dimension-wise interactions increases.
When forecasting 3D bounding boxes ($M = 6$), dimension-wise interactions contribute up to 83\%.
In other words, with the increase of trajectory dimensionality, the dimension-wise interaction may play a more significant role than social interactions when forecasting heterogeneous trajectories.
Especially comparing APP-C1 and APP-C2, which are both tested on SDD, it shows that as trajectory dimensionality $M$ increases, the dimension-wise interaction contributes more, although there are no social interaction differences and data differences (they are both SDD).

\begin{table}[htbp]
    \centering
    \caption{
        Quantitative results of contributions of the social interaction and the dimension-wise interaction to the prediction performance in \EMODEL~on heterogeneous trajectories.
        $M$, $I_{dwi}$, and $I_{soc}$ denote the dimension of a frame of trajectory vector, dimension-wise interaction, and social interactions, respectively.
    }
    \label{tab_app_dwi_soc}
    \begin{tabular}{c|c|c|c|cc}
        \toprule
        No. & Dataset & Type & $M$ & $I_{dwi}$ (\%) & $I_{soc}$ (\%) \\
        \midrule
        APP-C1 & SDD (co) & co & 2 & 61 & 39 \\
        APP-C2 & SDD (bb) & bb & 4 & 67 & 33 \\
        APP-C3 & nuScenes & 3dbb & 6 & 83 & 17 \\

        \bottomrule
    \end{tabular}
\end{table}

\end{document}